\documentclass[12pt,a4paper]{article}
 
\usepackage{parskip}
\usepackage{setspace}
\usepackage[pdftex]{graphicx}

\usepackage[cmex10]{amsmath}
\usepackage{amsfonts}
\usepackage{amssymb}
\usepackage{amsthm}
\newtheorem{mydef}{Definition}

\usepackage{algorithm}
\usepackage{algpseudocode}
\algnewcommand{\IfThenElse}[3]{% \IfThenElse{<if>}{<then>}{<else>}
	\State \algorithmicif\ #1\ \algorithmicthen\ #2\ \algorithmicelse\ #3}
\algrenewcommand{\Return}{\State\algorithmicreturn~}

\usepackage{xfrac}

\usepackage{anysize}
\marginsize{2cm}{2cm}{2cm}{2cm}

\usepackage{subfigure}
\usepackage[font=footnotesize]{caption}

\usepackage{multirow}
\usepackage{rotating}
\usepackage{makecell}
\usepackage{booktabs}
\usepackage[table]{xcolor}
\usepackage{color}

\usepackage[comma,authoryear]{natbib}

\usepackage[hidelinks=true]{hyperref}

\usepackage{enumerate}

\usepackage{lineno}
\usepackage{authblk}
\title{Heuristic Design of Fuzzy Inference Systems: A Review of Three Decades of Research}%: A review of the state-of-the-art
\author[1,2]{Varun~Ojha\thanks{Please cite as:\\Ojha, V, Abraham A, Snasel V. Heuristic Design of Fuzzy Inference Systems: A Review of Three Decades of Research, \textit{Engineering Applications of Artificial Intelligence (85), pp 845-864\\\href{https://doi.org/10.1016/j.engappai.2019.08.010}{doi.org/10.1016/j.engappai.2019.08.010}}}}
\author[3,4]{Ajith~Abraham}
\author[5]{V\'aclav~Sn\'a\v{s}el}
\affil[1]{ETH Z\"urich, Switzerland}
\affil[2]{University of Reading, United Kingdom}
\affil[3]{University of Pretoria, South Africa}
\affil[4]{Machine Intelligence Research Labs (MIR Labs), WA, United States}
\affil[5]{Technical University of Ostrava, Czech Republic}
\date{}
\begin{document}
% paper title
% make the title area
\maketitle
% As a general rule, do not put math, special symbols or citations in the abstract or keywords.
\begin{abstract}
This paper provides an in-depth review of the optimal design of type-1 and type-2 fuzzy inference systems (FIS) using five well known computational frameworks: genetic-fuzzy systems (GFS), neuro-fuzzy systems (NFS), hierarchical fuzzy systems (HFS), evolving fuzzy systems (EFS), and multiobjective fuzzy systems (MFS), which is in view that some of them are linked to each other. The heuristic design of GFS uses evolutionary algorithms for optimizing both Mamdani-type and Takagi-Sugano-Kang-type fuzzy systems; whereas, the NFS combines the FIS with neural network learning method to improve the approximation ability. HFS combines two or more low-dimensional fuzzy logic units in a hierarchical design to overcome the curse of dimensionality. EFS solves the data streaming issues by refining (evolving) the system incrementally, and MFS solves the multi-objective trade-offs like the simultaneous maximization of both interpretability and accuracy. The overall synthesis of these dimensions explores the FIS's potential challenges and opportunities; the complex relations among the dimension; and the FIS's potential to combining one or more computational frameworks adding another dimension: deep fuzzy systems.~\\

\end{abstract}
%%%%%%%%%%%%%%%%%%%%%%%%%%%%%%%%%%%%%%%%%%%%%%%%%%%%%%%%%%%%%%%%%%%%%%%%%%%%%%%%%%%%%%%%%%%%%%%%%%%%%%%%

%\newpage
%\setcounter{secnumdepth}{4}
%\setcounter{tocdepth}{4}
%\tableofcontents
%\clearpage

\section{Introduction}
\label{sec_fuzz_ref_intro} %
Research in fuzzy inference systems (FIS) initiated by~\cite{zadeh1988fuzzy} has drawn the attention of many {\color{black}disciplines} over the past three decades. The success of FIS is evident from its applicability and relevance in numerous research areas: control systems~\citep{lee1990fuzzy,wang1996approach}, engineering~\citep{precup2011survey}, medicine~\citep{jain2017fuzzy}, chemistry~\citep{komiyama2017chemistry}, computational biology~\citep{jin2008fuzzy}, finance and business~\citep{bojadziev2007fuzzy}, computer networks~\citep{elhag2015combination,gomez2002evolving}, fault detection and diagnosis~\citep{lemos2013adaptive}, pattern recognition~\citep{melin2011face}. These are just a few among numerous FIS's successful applications~\citep{liao2005expert,castillo2014review}, which is mainly attributable to FIS's ability to manage uncertainty and computing for noisy and imprecise data~\citep{zadeh1992fuzzy}.

The enormous amount of research and innovations in multiple dimensions of FIS propelled its success. These research dimensions realize the concept of: genetic-fuzzy systems (GFS), neuro-fuzzy systems (NFS), hierarchical fuzzy systems (HFS), evolving fuzzy systems (EFS), and multiobjective fuzzy systems (MFS) which are fundamentally relied on two basic {\color{black}fuzzy rule types}: Mamdani type~\citep{mamdani1974application}, and Takagi--Sugano--Kang (TSK) type~\citep{takagi1985fuzzy}. Both rule types have ``IF X is A THEN Y is B'' rule structure, i.e., the rules are in {\color{black}the} \textit{antecedent} and \textit{consequent} form. However, {\color{black}the rule types} Mamdani-type and TSK-type differ in their respective consequent. For the consequent, Mamdani-type takes an output action (a class), and TSK-type takes a polynomial function. Thus, they differ in their approximation ability. The {\color{black}Mamdani-type} has a better interpretation ability, and the {\color{black}TSK-type} has a better approximation accuracy. {\color{black}For antecedent, both types take a similar form that is a \textit{rule induction} process take place for input space partition to form antecedent part of a rule. Therefore, the \textit{rule types}, the \textit{rule induction} process, and the \textit{interpretability-accuracy} trade-off govern} the FIS's dimensions.   

In GFS, researchers investigate mechanisms to {\color{black}encode and optimize the FIS's components. The encoding takes place in the form of genetic vectors and genetic population and the optimization take place in the form of FIS's structure and parameters optimization.} \cite{herrera2008genetic,cordon2004ten}, and \cite{castillo2012optimization} summarized research in GFS {\color{black}with taxonomy to explain both encoding and structure optimization using a genetic algorithm (GA).} NFS research investigates network structure formation and parameter optimization~\citep{jang1993anfis} and {\color{black}answers} the variations in network formation methods and the variations in parameter optimization techniques. \cite{buckley1995neural,andrews1995survey}, and \cite{karaboga2018adaptive} offer {\color{black}summaries} of such variations. \cite{torra2002review} and \cite{wang2006survey} reviewed research in HFS which summarizes the variations in {\color{black}HFS} design types and {\color{black}HFS} parameter optimization techniques. {\color{black}The EFS research} enables incremental learning ability into {\color{black}FISs}~\citep{kasabov1998evolving,angelov2008evolving}, and the MFS research {\color{black}enables FISs} to deal with multiple objectives simultaneous~\citep{ishibuchi2007multiobjective,fazzolari2013review}.

This review paper offers a synthesized view of each dimension: GFS, NFS, HFS, EFS, and MFS. The synthesis recognizes these dimensions being linked to each other where the concept of one dimension applies to another. For example, NFS and EFS models can be optimized by GA. Hence, GFS entails its concepts to NFS and EFS. The complexity and concept arises from {\color{black}the synthesis offer a potential to investigate} \textit{deep fuzzy systems} (DFS), which may take advantage of GFS, HFS, and NFS simultaneously in a hybrid manner where NFS will offer solutions to network structure formation, HFS may offer solutions to resolving hierarchical arrangement of multiple layers, and GFS may offer solutions to parameter optimization. Moreover, EFS and MFS also play a role in DFS is {\color{black}if the goal will be to construct a system for the data stream and to optimize a system} for interpretability-accuracy trade-off.

This review walks through each dimension: GFS, NFS, HFS, EFS, and MFS, including a discussion on the standard FIS. First, the {\color{black}rule structure, rule types, and FISs types are discussed in Sec.~\ref{sec_fis}.} A discussion on the FIS's designs describing how various FIS's paradigms emerged through the interaction of FIS with neural networks (NN) and evolutionary algorithms (EA) is given in Sec.~\ref{sec_fs_farmework}. {\color{black}Sec.~\ref{sec_gfs} discusses the GFS paradigm which emerged through FIS and EA combinations}. Sec.~\ref{sec_nfs} describes the NFS paradigm including reference to self-adaptive and online system notions (Sec.~\ref{sec_nfs_notions}); basic layers (Sec.~\ref{sec_nfs_layer}); and feedforward and feedback architectures (Sec.~\ref{sec_nfs_arch}). {\color{black}They are followed} by the discussions on the HFS's properties and the HFS's implementations (Sec.~\ref{sec_hfs}). Sec.~\ref{sec_efs} summarized the EFS which offers an incremental leaning view in FIS. Sec.~\ref{sec_mfs} offered the discussions on MFS which covers the Pareto-based multiobjective optimization and the FIS's multiple objective trade-offs implementations. Followed by the challenges and the future scope in Sec.~\ref{sec_fuzzy_rev_challenges}, and conclusions in Sec.~\ref{sec_fuzzy_rev_con}.

\section{Fuzzy inference systems}
\label{sec_fis}
\begin{figure}
    \centering
    \includegraphics[width=0.6\textwidth]{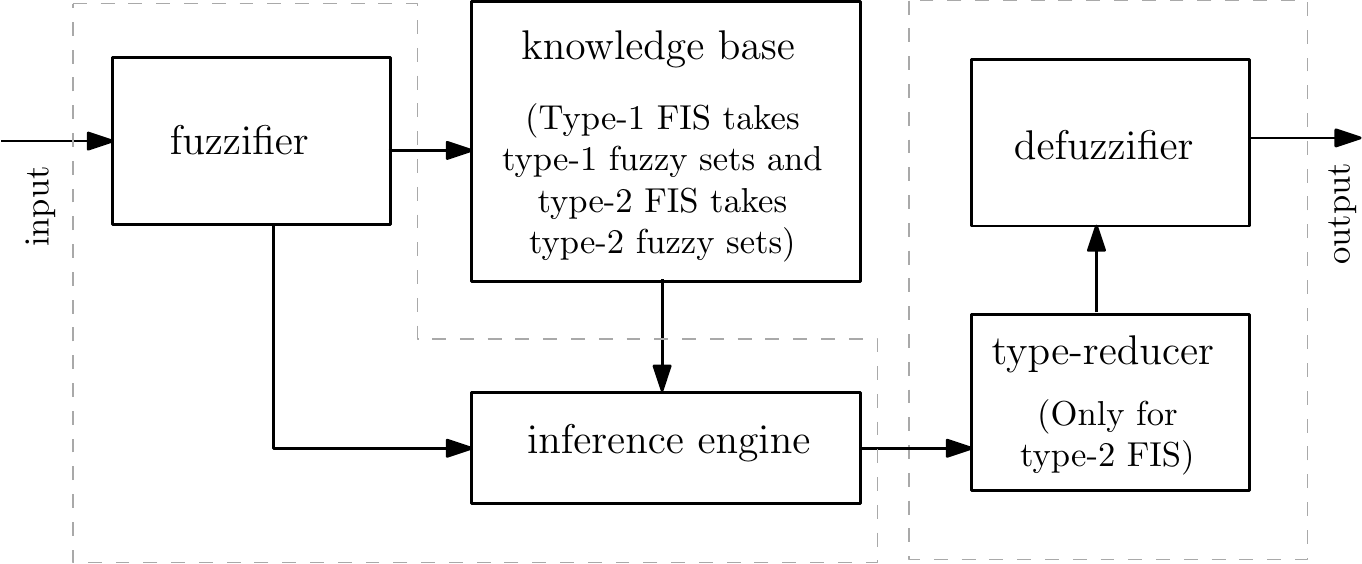}
    \caption{Typical fuzzy inference system.}
    \label{fig_FIS}
\end{figure}

A standard FIS (Fig.~\ref{fig_FIS}) is composed of the following components: 
\begin{enumerate}[(1)]
    \item {\color{black}a \textbf{fuzzifier} unit that fuzzifies the input data;} 
    
    \item a \textbf{knowledge base} (KB) unit, which contains fuzzy rules of the form IF-THEN, i.e., 
    \begin{flushleft}
        IF a set of conditions (antecedent) is satisfied~\\
        THEN a set of conditions (consequent) can be inferred
    \end{flushleft}

    \item an \textbf{inference engine} {\color{black}module that computes the rules firing strengths to infer knowledge from KB; and} 

    \item a \textbf{defuzzifier} unit that translates inferred knowledge into a rule action (crisp output). 
\end{enumerate}

The KB of the FIS is composed of a database (DB) and a \textit{rule-base} (RB). The DB assigns {\color{black}fuzzy sets (FS)} to the input variables and the FSs transforms the input variables to fuzzy membership values. For \textit{rule induction}, RB constructs a set of rules fetching FSs from the DB.

{\color{black}In a FIS, an input can be a numeric variable or a linguistic variable. Moreover, an input variable can be singleton [Fig.~\ref{fig_fis_inputs}(a)] and non-singleton [Fig.~\ref{fig_fis_inputs}(b)]. Accordingly, a FIS is \textbf{singleton FIS} if it uses singleton inputs, i.e., FIS uses crisp and precise single value measurement as the input variables, which is the most common practice. However, real-world problems, especially in engineering, measurements are noisy, imprecise, and uncertain. Thus, FIS that uses non-singleton input is a \textbf{non-singleton FIS}. Thus, in principle, a non-singleton FIS differs with a singleton FIS in input fuzzification process where a ``fuzzifier'' transform a non-singleton input and a singleton input to a fuzzy membership value.
\begin{figure}
    \centering
    \subfigure[]
    {  
        \includegraphics[width=0.4\textwidth]{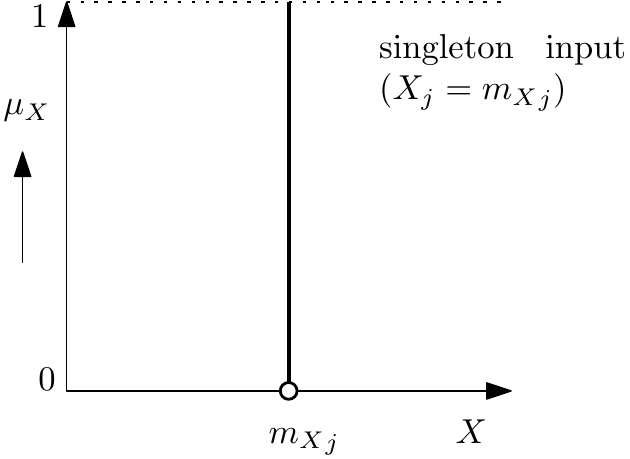}%signleton	
    }\quad~\quad~\quad
    \subfigure[]
    {   
        \includegraphics[width=0.4\textwidth]{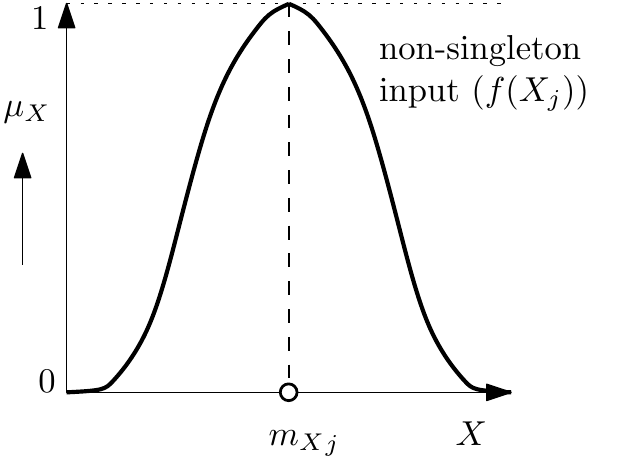}%non-singleton
    }
    \caption{\color{black}Input examples: (a) singleton input $ \mu_{X_j}(X_j)$ and (b) non-singleton input $ \mu_{X_j}(X_j) = f(X_j) $.}
    \label{fig_fis_inputs}
\end{figure}

A fuzzifier maps a singleton input (crisp input) $ X_j \in \textbf{X}$, $ \textbf{X} = (X_1, X_2, \ldots, X_P) $ for $ {m_X}_j $ (a value in $ X_j $) [Fig.~\ref{fig_fis_inputs}(a)] to the following membership function for the input fuzzyfication:
\begin{equation}
\label{eq_fuzzifier_sigleton}
\mu_{X_j}(X_j) = \left\lbrace
\begin{array}{ll}
1, & X_j = {m_X}_j\\
0, & X_i \ne {m_X}_j \quad \forall X_j \in \textbf{X}
\end{array}
\right.
\end{equation}

For non-singleton inputs, a fuzzifier maps input $ X_j $ (that is considered as noisy, imprecise, and uncertain) onto a Gaussian function (typical choice for numeric variables) as:
\begin{equation}
\label{eq_fuzzifier_non_sigleton}
\mu_{X_j}(X_j) = f(X_j) = \exp\left[-\frac{1}{2}\left(\frac{X_j-{m_X}_j}{\sigma_X}\right)^2\right]
\end{equation}
where $ {m_X}_j $ is input (considered as mean, a value along line $ X_j $) and $ \sigma_X \ge 0$ is the standard deviation (std.) that defines the spread of the function $ \mu_{X_j} $. The value of the fuzzy set at $ {m_X}_j $ is $ \mu_{X_j}({m_X}_j) = 1$ and $  \mu_{X_j}(X_j) $ decreases from unity as $ X_j $ moves away from $ {m_X}_j $ \citep{mouzouris1997nonsingleton}. In general, for a singleton or non-singleton input $ X_j $, inference engine output $ \mu_{AX_j} $ is a combination of fuzzified input $ \mu_{X_j}(X_j) $ with an antecedent FS $ \mu_{A_j} (X_j) $ as per:
\begin{equation}
\label{eq_antecedent}
\mu_{AX_j}(\bar{X_j}) = \sup\left\lbrace \mu_{X_j}(X_j) \star \mu_{A_j}(X_j) \right\rbrace
\end{equation}
where $ \star $ is \textit{t-norm} operation that can be minimum or product and $ \bar{X_j} $ indicate supremum of Eq.~\eqref{eq_antecedent}. Fig.~\ref{fig_fis_guss_product} is an example product operation in Eq.~\eqref{eq_antecedent}. Fig.~\ref{fig_fis_guss_product} evaluates the product of input FS $ \mu_X $ and the antecedent fuzzy set $ \mu_A $ that result in $ \mu_{AX} = 0.04$ for where $ {m_X}+j = 2.0 $, $ \sigma_X = 2.0$, $ m_A = 0.0$, and $ \sigma_A = 1.5$. The product $ \mu_{AX}(\bar{X}) $ gives a maximum value at $ \bar{X} = 0.72 $ (in Fig.~\ref{fig_fis_guss_product}) which is calculated as:
\begin{equation}
\label{eq_product}
\bar{X} = \frac{m_A \sigma_X^2 + m_X \sigma_A^2}{\sigma_A^2 + \sigma_X^2}
\end{equation}
\begin{figure}
    \centering
    \includegraphics[width=0.4\textwidth]{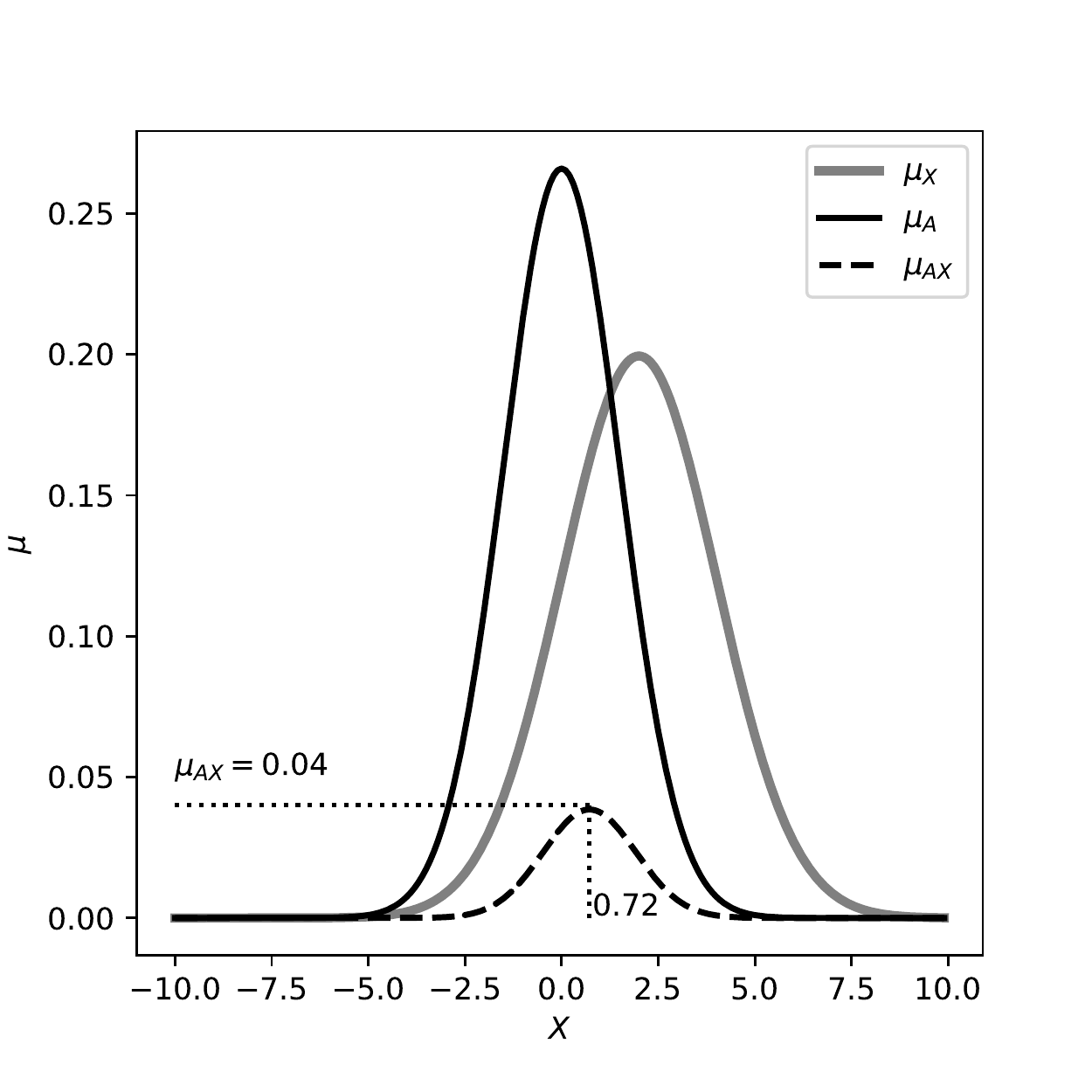}%signleton	
    \caption{\color{black} Product (as a \textit{t-norm} operation) $ \mu_{AX} $ of input FS $ \mu_X $ and the antecedent fuzzy set $ \mu_A $ as per Eq.~\eqref{eq_antecedent}.}
    \label{fig_fis_guss_product}
\end{figure}
}%blue color ends here

The design of RB {\color{black}further distinguishes} the type of FISs: a \textbf{Mamdani-type FIS}~\citep{mamdani1974application} or a \textbf{Takagi-Sugano-Kang (TSK)-type FIS}~\citep{takagi1985fuzzy}. {\color{black}A TSK-type FIS differs with a Mamdani-type FIS only in the implementation of fuzzy rule's consequent part. In  Mamdani-type FIS rule's consequent part is an FS, whereas in TSK-type FIS rule's consequent part is a polynomial function.}

The DB contains FSs that {\color{black}are} either a type-1 fuzzy set (T1FS) or a type-2 fuzzy set (T2FS). The basic form of a fuzzy membership function (MF) is coined as a T1FS; whereas, T2FS allows an MF to be fuzzy itself by extending membership value into an additional membership dimension. Hence, the fuzzy set (FS) types {\color{black}also differentiate FIS types}: \textbf{type-1 FIS} (T1FIS) and the \textbf{type-2 FIS} (T2FIS).
 
{\color{black} For simplicity, this paper is singleton FIS centric and refers non-singleton FIS to appropriate research. As well as, since Mamdani-type FIS differs with TSK-type FIS only in its consequent part, this paper focuses on TSK-type FIS.} 

\subsection{Type-1 fuzzy inference systems}
\label{sec_fls1}
A TSK-type FIS is governed by the ``IF--THEN'' rule of the form~\citep{takagi1985fuzzy}:
\begin{equation}
\label{eq_type1_rules}
\mathbf{r}^i: \text{IF } X^i_1 \text{ is } {A}^i_1 \text{ and } \cdots \text{ and } X^i_{p^i} \text{is } {A}^i_{p^i} \text{ THEN } Y^i \text{ is } B^i,
\end{equation}
where $ \mathbf{r}^i $ is the $ i^{th} $ rule in the FIS's RB. The $ i^{th} $ rule has $ A^i $ as the T1FS, and $ B^i $ as a function of inputs $X^i_1, X^i_2, \ldots, X^i_{p^i} $ that returns a crisp output $ Y^i $. At the $ i^{th} $ rule, $ p^i \le P$ inputs are selected from $ P $ inputs. Note that $ p^i $ varies from rule-to-rule, and thus, the input dimension at a rule $ i $ is denoted as $ p^i $. {\color{black}That is,  the subset of inputs to a rule has $ p^i \le  P $ elements, which leads to a \textit{incomplete rule} because all available inputs may not be present to rule premises (antecedent part). Otherwise, a \textit{complete rule} has all available inputs at its  premises.} The function $ B^i $, for TSK-type, is commonly expressed as: 
\begin{equation}
\label{eq_type1_consequent}
B^i = c^i_0 + \sum\limits_{j=1}^{p^i} c^i_jX^i_j,
\end{equation}
where $ X^i_j$ is the inputs and $ c^i_j$ for $ j$ = $ 0 $  to  $p^i$ is the \textit{free parameters} at the consequent part of a rule. For Mamdani-type, $ B^i $ may be expressed as a ``class.'' The basic building blocks of a FIS is shown in Fig.~\ref{fig_FIS} whose defuzzified crisp output is computed as follows. At first, the inference engine fires the RB's rules, each rule has a firing strength $ F^i $ as:
\begin{equation}
\label{eq_type1_firing_strength}
F^i =  \prod\limits_{j = 1}^{p^i} \mu_{A^i_j} (X^i_j),
\end{equation}
where {\color{black} $ \mu_{A^i_j} \in [0,1] $ is the membership value} of $ j^{th} $ T1FS MF (e.g., Fig.~\ref{fig_fis_MF}a) at the $ i^{th} $ rule. {\color{black}Assuming firing strength $ F^i $ has to be computed for a non-singleton input $ \mu_{X^i_j}(X^i_j) $, then firing strength $ F^i $ will replace $ \mu_{A^i_j}(X^i_j) $ in Eq.~\eqref{eq_type1_firing_strength} by $ \mu_{AX^i_j}(X^i_j) $ as per Eq.\ref{eq_antecedent}. A detail generalization definition of firing strength computation is given by~\cite{mouzouris1997nonsingleton}.}
    
The defuzzified output $ \hat{Y} $ of T1FIS{\color{black}, as an example,} is computed as: 
\begin{equation}
\label{eq_type1_out}
\hat{Y} = \frac{\sum_{i=1}^M B^i F^i}{\sum_{i=1}^M F^i},
\end{equation}
where $ M $ is the {\color{black} total rules} in the RB. 
%
%An example of T1FS  $ A $ is illustrated in~Fig.~\ref{fig_fis_MF}a, which has the following form:
%\begin{equation}
%\label{eq_type1_MF}
%\mu_A(x) = \frac{1}{1 + \left(\frac{x-m}{\sigma} \right)}
%\end{equation}
%where $ m $ and $ \sigma $ are the center and the width respectively of MF $ \mu_A(x) $.

\subsection{Type-2 fuzzy inference systems}
\label{sec_fls2}
\begin{figure}
    \centering
	\subfigure[]
	{  
		\includegraphics[width=0.4\textwidth]{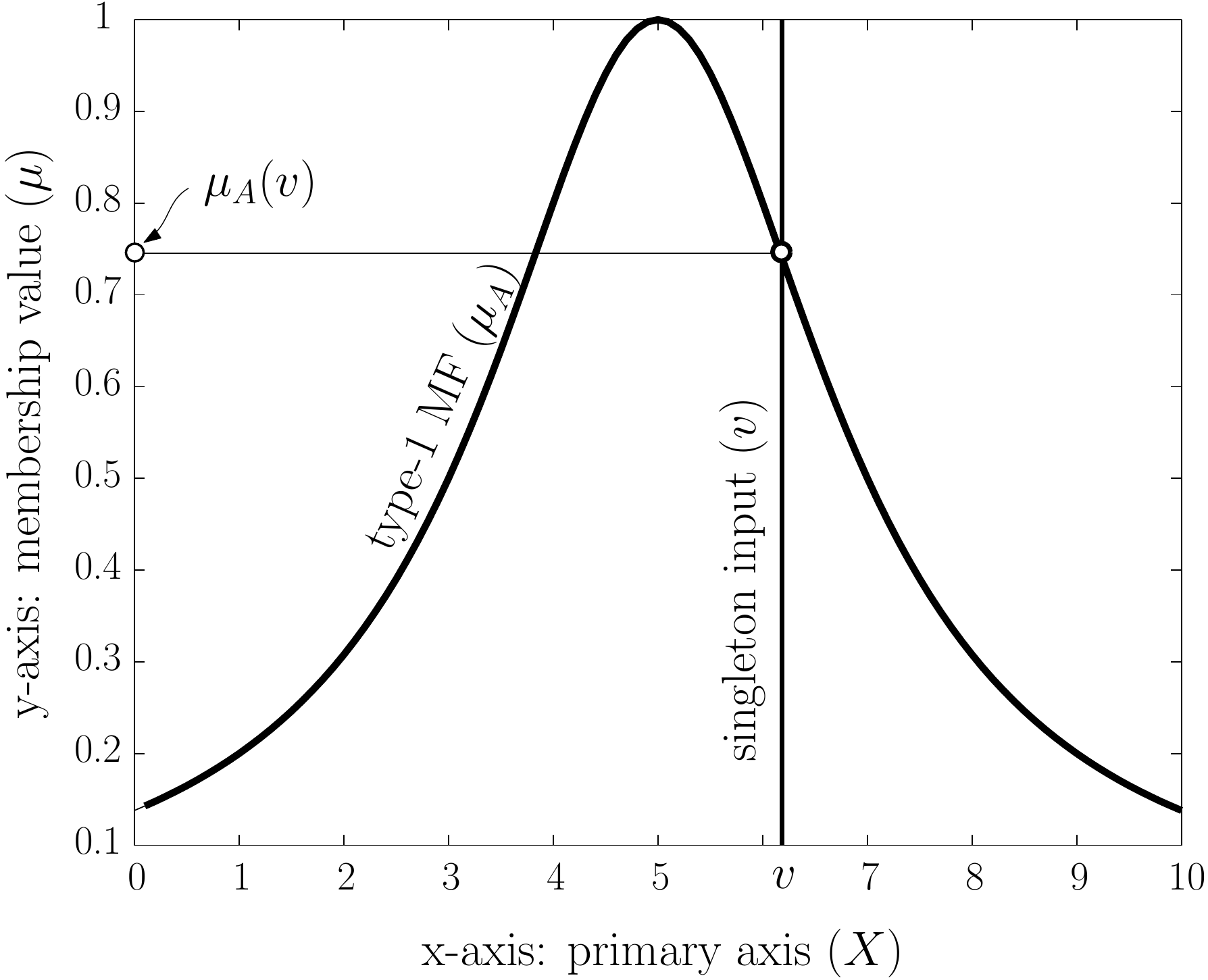}%T1GMF	
	}\quad~\quad
	\subfigure[]
	{   
		\includegraphics[width=0.4\textwidth]{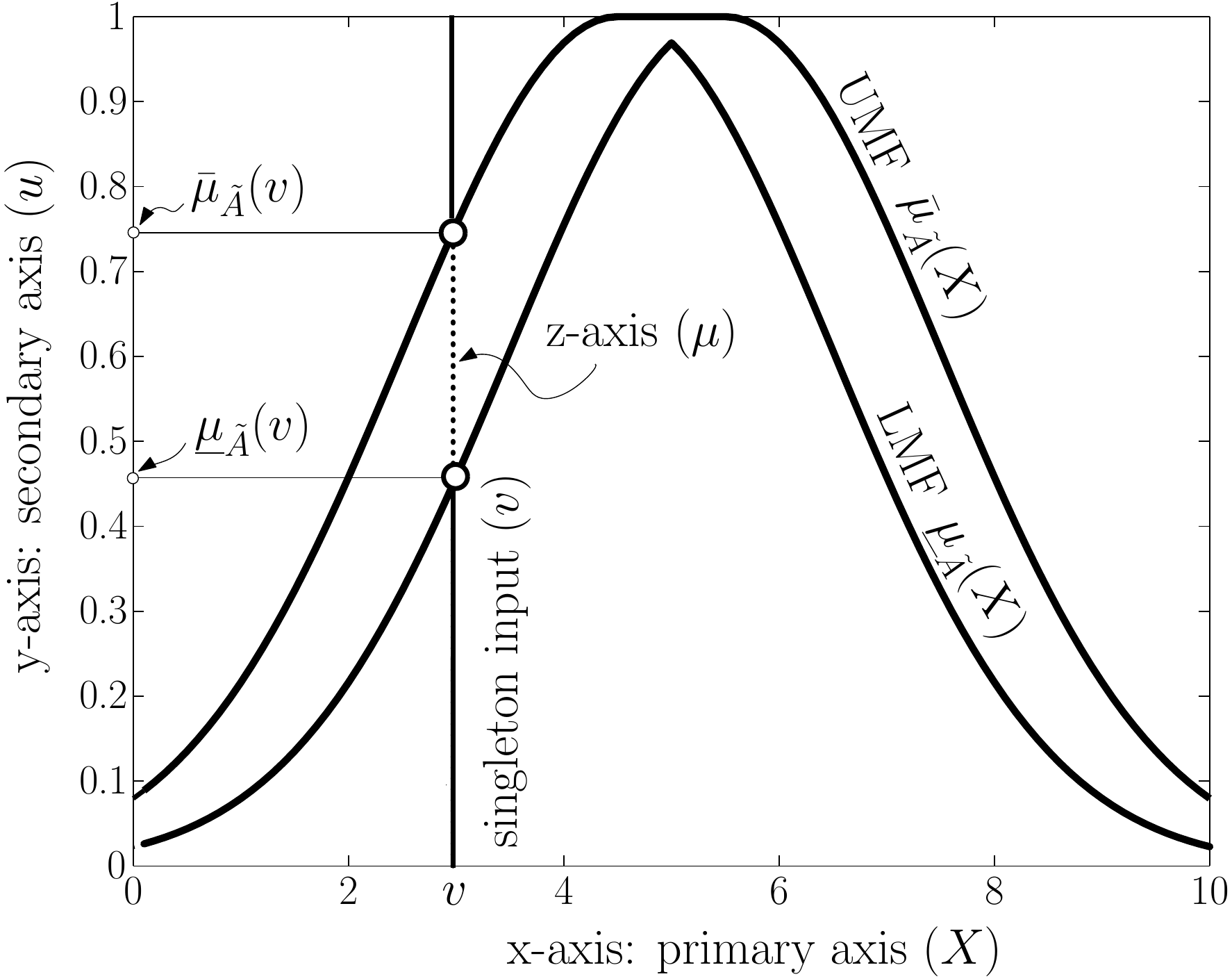}%T2GMF
	}
	\caption{Fuzzy MF examples: (a) Type-1 MF $ \mu_A(X) = 1/[1 + \left(\left( X-m \right)/\sigma\right)^2] $ with center $ m = 5.0 $ and width $ \sigma = 2.0 $. (b)Type-2 MF with fixed $ \sigma = 2.0 $ and with means $ m_1 = 4.5 $ and $ m_2 = 5.5 $. UMF $ \bar{\mu}_{\tilde{A}} (X) $ as per Eq.~\eqref{eq_upMF} is in solid line and  LMF $ \underline{\mu}_{\tilde{A}} (X) $  as per Eq.~\eqref{eq_loMF} is in dotted line.}
    \label{fig_fis_MF}
\end{figure}

A T2FS $\tilde{A}$ is characterized by a 3D MF~\citep{mendel2013km}: The x-axis is the primary variable, the y-axis is secondary variable (primary MF denoted by $ u $), and the z-axis is the MF value (secondary MF denoted by $ \mu $). Hence, for {\color{black}a singleton} input $ X $, a T2FS $\tilde{A}$ is defined as:
\begin{equation}
\label{eq_type2MF}
\tilde{A} = \left\lbrace \left( \left(X,u\right), \mu_{\tilde{A}} \left(X,u\right)  \right) | \,\forall X \in \mathbf{X}, \forall u \in [0,1] \right\rbrace.
\end{equation}
The MF value $ \mu $ has a 2D support, called ``footprint of uncertainty'' of $ \tilde{A} $, which is bounded a lower membership function (LMF) $ \underline{\mu}_{\tilde{A}}(X) $ and an upper membership function (UMF) $ \bar{\mu}_{\tilde{A}} (X)$. A T2FS bounded by an LMF and a UMF is an \textit{interval type-2 fuzzy set} (IT2FS), e.g., a Gaussian function [Eq.~\eqref{eq_GaussT2MF}] with uncertain mean $m \in [m_1, m_2] $ and {\color{black}std.} $ \sigma \ge 0 $ is an IT2FS (e.g., Fig.~\ref{fig_fis_MF}b): 
\begin{equation}
\label{eq_GaussT2MF}
\mu_{\tilde{A}} (X,m,\sigma) = \exp \left( -\frac{1}{2} \left(\frac{X - m}{\sigma}\right)^2\right), \quad m \in [m_1,m_2].
\end{equation} 
An LMF [Eq.~\eqref{eq_loMF}] {\color{black} $ \underline{\mu}_{\tilde{A}} (X) \in [0,1] $} and a UMF [Eq.~\eqref{eq_upMF}] {\color{black} $ \bar{\mu}_{\tilde{A}} (X) \in [0,1]$}  of an IT2FS can be defined as~\citep{karnik1999type}:
\begin{equation}
\label{eq_loMF}
\underline{\mu}_{\tilde{A}} (X) = \left\lbrace
\begin{array}{ll}
\mu_{\tilde{A}} (X,m_2,\sigma), & X \le (m_1 + m_2)/2\\
\mu_{\tilde{A}} (X,m_1,\sigma), & X > (m_1 + m_2)/2
\end{array}
\right.
\end{equation}
%and the UMF can be defined as~\citep{karnik1999type}:
\begin{equation}
\label{eq_upMF}
\bar{\mu}_{\tilde{A}} (X) = \left\lbrace
\begin{array}{ll}
\mu_{\tilde{A}} (X,m_1,\sigma), & X < m_1\\
1, &  m_1 \le x \le m_2 \\
\mu_{\tilde{A}} (X,m_2,\sigma),  & X > m_2
\end{array}
\right.
\end{equation}

\noindent
In Fig.~\ref{fig_fis_MF}b, a point $ v $ along the x-axis of 3D-IT2FS cuts the UMF and LMF along the y-axis, and the value $ \mu $ of the type-2 MF is taken along the z-axis [dotted line, which a MF in the third dimension in Fig.~\ref{fig_fis_MF}b between $ \bar{\mu}_{\tilde{A}} (X = v) $ and $ \underline{\mu}_{\tilde{A}} (X= v) $]. Considering the IT2FS MF, the $ i^{th} $ IF--THEN rule of TSK-type T2FIS, for inputs $ \mathbf{X} = ( X_1,X_2,\ldots,X_{p^i} )$, takes the form:
\begin{equation}
\label{eq_type2_rules}
\mathbf{r}^i: \text{IF } X^i_1 \text{ is } \tilde{A}^i_1 \text{ and } \cdots \text{ and } X^i_{p^i} \text{ is } \tilde{A}^i_{p^i} \text{ THEN } Y^i \text{ is } \tilde{B}^i,
\end{equation}
where $ \tilde{A}^i $ is a T2FS, $ \tilde{B}^i $ is  a function of $ \mathbf{X} $ that returns a pair $[\underline{b}^i, \bar{b}^i]$ called left and right weights of the consequent part of a rule. In TSK, $ \tilde{B}^i $ is usually written as:
\begin{equation}
\label{eq_type2_consequent}
\tilde{B}^i  = [c^i_0-s^i_0,{c}^i_0+s^i_0]  + \sum\limits_{j=1}^{p^i} [c^i_j-s^i_j,{c}^i_j+s^i_j]X^i_j,
\end{equation}
where $ X^i_j$ is the input and $ c^i_j $ for $ j$ = $ 0 $  to  $p^i$ is a rule's consequent part's parameter and $ s^i_j $ for $ j$ = $ 0 $  to  $p^i$ is its deviation factor. The firing strength $ F^i = [\underline{f}^i, \bar{f}^i]$ of IT2FS is computed as:
\begin{equation}
\label{eq_fire}
\underline{f}^i = \prod\limits_j^{p^i} \underline{\mu}_{\tilde{A}_j^i} \mbox{ and }  \bar{f}^i = \prod\limits_j^{p^i} \bar{\mu}_{\tilde{A}_j^i}.
\end{equation}
At this stage, the inference engine fires the rule and the \textit{type-reducer}, e.g., center of set $ Y_{cos} $ as per Eq.~\eqref{eq_cos} reduces the T2FS to T1FS~\citep{karnik1999type,wu2009enhanced}:
\begin{equation}
\label{eq_cos}
Y_{cos} = [Y_l, Y_r] %= \underset{f^i \in F^i, \,\, b^i \in \tilde{B}^i}{\bigcup} \frac{\sum_{i = 1}^M f^i b^i}{\sum_{i = 1}^M f^i},
\end{equation}    
where $ Y_l $ and $ Y_r $ are left and right {\color{black}ends} of the interval. For the ascending order of $ \underline{b}^i $ and $ \bar{b}^i$, $ y_l $ and $ y_r $ are computed as:
\begin{equation}
\label{eq_yl}
Y_l = \frac{\sum_{i = 1}^L \bar{f}^i \underline{b}^i  + \sum_{i = L+1}^{M}\underline{f}^i \underline{b}^i}{\sum_{i = 1}^L\bar{f}^i + \sum_{i = L+1}^{M}\underline{f}^i} \mbox{ and }  Y_r = \frac{\sum_{i = 1}^R \underline{f}^i \bar{b}^i  + \sum_{i = R+1}^{M}\bar{f}^i \bar{b}^i}{\sum_{i = 1}^R\underline{f}^i + \sum_{i = R+1}^{M}\bar{f}^i},
\end{equation}
%\end{equation}
where $ L $ and $ R $ are the switch points determined as per 
$ \underline{b}^L \le Y_l \le \underline{b}^{L+1} \text{ and } \bar{b}^R \le Y_r \le \bar{b}^{R+1},$
respectively. Subsequently, defuzzified crisp output $ \hat{Y} = (Y_l + Y_r)/2 $ is computed. 

{\color{black}For a \textit{non-singleton interval type-2 FIS}, lower and upper intervals of non-singleton inputs are created. Additionally, similar to the non-singleton input fuzzification $ \mu_{AX} $ in the case of non-singleton type-1 FIS using input FS $ \mu_X $ and antecedent FS $ \mu_A $ shown in Eq.~\eqref{eq_antecedent}, for non-singleton type-2 FIS, both lower $ \underline{\mu}_{\tilde{AX}} $ and upper $ \bar{\mu}_{\tilde{AX}} $ intervals products are calculated using lower and upper input FSs $ \underline{\mu}_X $ and $ \bar{\mu}_X $ and lower and upper antecedent FSs $ \underline{\mu}_{\tilde{A}} $ and $ \bar{\mu}_{\tilde{A}} $. \cite{sahab2011adaptive} describe the computation of non-singleton type-2 FIS in detail.} 

\subsection{Heuristic designs of fuzzy systems}
\label{sec_fs_farmework}
\begin{figure}
	\centering
	\includegraphics[scale=0.8]{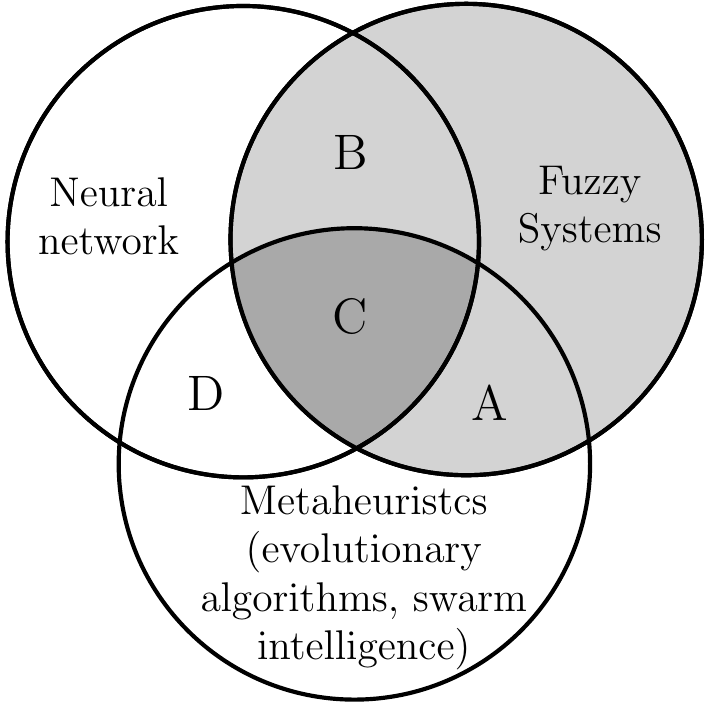}
	\caption{Spectrum of fuzzy inference system paradigms.}
	\label{fig_fuzzy_paradigm}
\end{figure}

The FIS types: Type-1 (Sec.~\ref{sec_fls1}) and Type-2 (Sec.~\ref{sec_fls2}) follow a similar design procedure and differ only in the type of FSs  being used. The heuristic design of FIS can be viewed from its hybridization with neural networks (NN), evolutionary algorithms (EA), and metaheuristics (MH) (Fig.~\ref{fig_fuzzy_paradigm}). And, such a confluence offers~\citep{herrera2008genetic}: 
\begin{enumerate}[\hspace{2em}(1)]
    \setlength{\itemsep}{0pt}
    \setlength{\parskip}{0pt}
    \item genetic fuzzy systems (A);
    \item neuro-fuzzy systems (B); 
    \item hybrid neuro-genetic fuzzy systems (C); and
    \item heuristics design of NNs (D).
\end{enumerate}
This paper discusses areas A, B, and C of Fig.~\ref{fig_fuzzy_paradigm}, area D in Fig.~\ref{fig_fuzzy_paradigm} is discussed in detail by~\cite{ojha2017metaheuristic}. The heuristic design installs learning capabilities into FIS which come from the optimization of its components. The FIS optimization/learning in a supervised environment is common practice.

%\subsubsection{Learning/optimization of fuzzy systems}
Typically, in \textbf{supervised learning}, a FIS is trained/optimized by supplying training data ($ \mathbf{X}, \mathbf{Y} $) of $ N $ input--output pairs, i.e., $ \mathbf{X} = (X_1, X_2, \ldots, X_P) $ and $ \mathbf{Y} = (Y_1, Y_2, \ldots, X_Q) $. Each input variable~$ X_j = \langle x_{j1},x_{j2},\ldots,x_{jN} \rangle^T$ is an $ N $--dimensional vector, and it has a corresponding $ N $--dimensional desired output vector $Y_j = \langle y_{j1}, y_{j2},\ldots, y_{jN} \rangle^T$. For the training data ($ \mathbf{X}, \mathbf{Y} $), a FIS model $ f(\mathbf{X},R)$ produces output $ \hat{\mathbf{Y}} = (\hat{Y}_1, \hat{Y}_2, \ldots,\hat{Y}_Q) $, where $ f: \mathbf{X} \times \mathbf{Y} \rightarrow \hat{\mathbf{Y}} $, $R = \{ \mathbf{r}_1, \mathbf{r}_2,\ldots, \mathbf{r}_M \}$ is a set of fuzzy $ M $ rules, and $ \hat{Y}_j = \langle \hat{y}_{j1}, \hat{y}_{j2},\ldots, \hat{y}_{jN} \rangle^T$  is an $ N $--dimensional model's output, which is compared with the desired output $ Y_j $, $ \forall \, j = 1,2,\ldots, Q $ and $\forall \, k =1,2,\ldots, N $, by using some error/distance/cost function $ c_f $ over model $ f(\mathbf{X},R)$. 

The cost function $ c_f $ can be a \textit{mean squared error} function or can be an \textit{accuracy} measure, depending on the desired outputs being continuous (regression) or discrete (classification)~\citep{caruana2004data}. Learning of FIS is therefore rely on reducing a cost function $ c_f $ by employing strategies for designing and optimizing a FIS model $ f(\mathbf{X},R)$, where model design may be refereed to how the FIS's components interact with each other and optimization may be referred to: RB design, RB parameter learning, and rule selection. In summary, FIS design, optimization, learning, and modeling is viewed as:
\begin{enumerate}[\hspace{2em}(1)]
    \setlength{\itemsep}{0pt}
    \setlength{\parskip}{0pt}
    \item the selection of FSs via fuzzy partitioning of input-space;
    \item the design of FIS's rules via an arrangement of rule and inputs;
    \item the optimization of the rule's parameters; and
    \item the inference from the designed FIS.
\end{enumerate}
%\begin{equation}
%\label{eq_mse}
%c_f(\mathbf{y}_i,\hat{\mathbf{y}_i)  = \frac{1}{N}\sum\limits_{i = 1}^{N} \sum\limits_{j = 1}^{q} \left(y_{ij} - \hat{y}_{ij} \right)^2, 
%\end{equation} 
%where $ y_{ij} $ are the desired outputs and $ \hat{y}_{ij} $ are the model's outputs, and their differences are summed over $N$ data pairs. 

\begin{figure}
	\centering
	\includegraphics[width=0.7\textwidth]{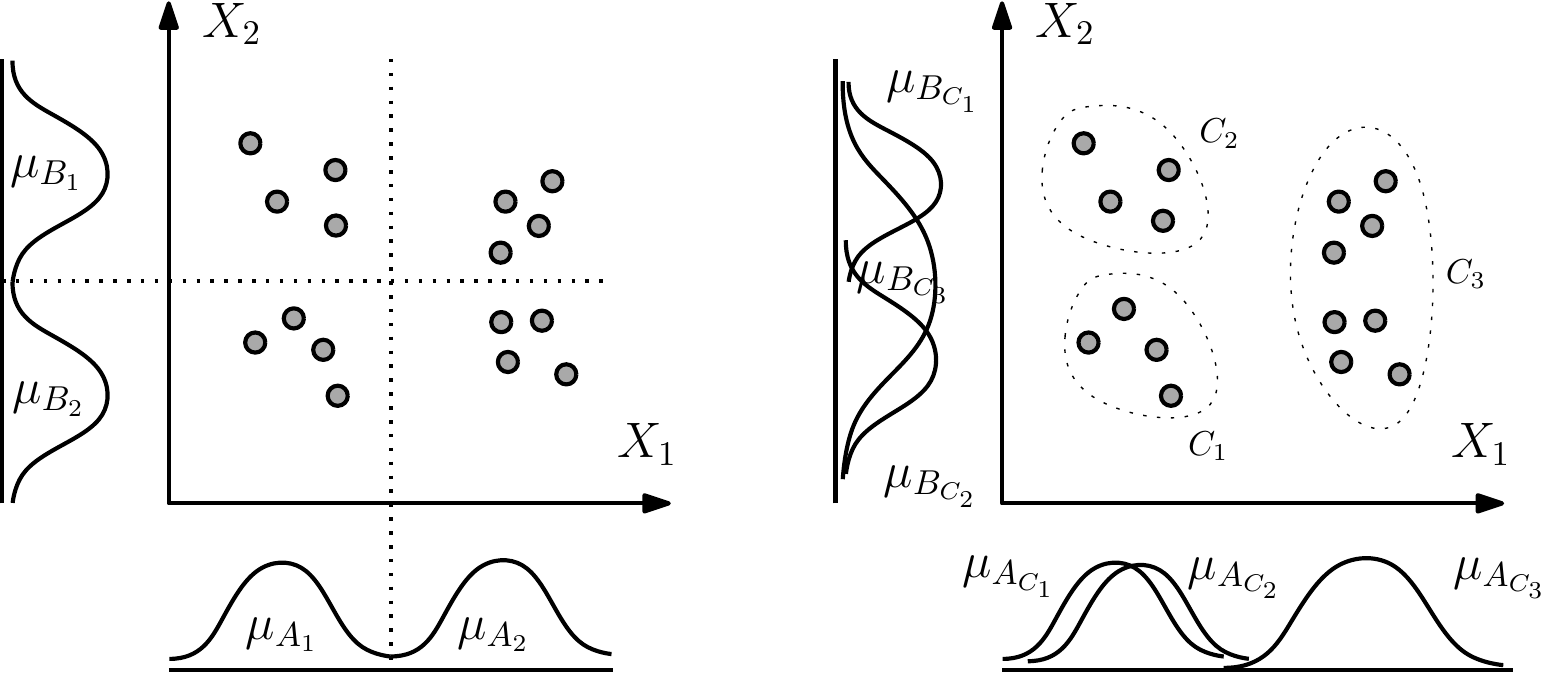}
	\caption{Input-space partitioning: grid partitioning (left) and clustering partitioning (right). Two-dimensional input-space (inputs $ X_1 $ and $ X_2 $) is partitioned by assigning MF $ \mu_{A_j} $ to input $ X_1 $ and MF $ \mu_{B_j} $ to input $ X_2 $. In the case of grid partitioning (left), $ j = 1, 2 $; and in the case of clustering based partitioning (right), $ j = C_1, C_2, C_3 $.}
	\label{fig_input_space}
\end{figure}

%\subsubsection{Fuzzy partitioning} 
Often a Gaussian function, a triangular function, or a trapezoidal function are selected as the MF of an FS~\citep{zadeh1999fuzzy}. The \textbf{input-space partition} corresponding to the MF assignments is one of the most crucial aspects of FIS design. For example, a two-dimension input-space in Fig.~\ref{fig_input_space} having inputs $ X_1 $ and $ X_2 $ are partitioned using a grid-partitioning method~\citep{jin2000fuzzy,jang1993anfis} or a clustering-based partitioning method~\citep{juang1998online,kasabov2002denfis}. Fig.~\ref{fig_input_space} is an example of inputs space partitioning for numerical variables. An example of partitioning for linguistic terms is explained by~\cite{cord2001genetic}. \cite{mao2005adaptive} presented an example of  input-space partitioning using a binary tree, where the root of the tree takes whole input $ \mathbf{X} $ and partition it into two children nodes $ X_l \in \mathbf{X} $ and $ X_r \in \mathbf{X} $. The partitioned subsets  $ \{X_l,X_r\} \subset \mathbf{X} $ are assessed for a defined cost function $ c_f $. If the cost $ c_f $ is lower than a defined threshold $ \epsilon_{err} $ than the input-space partitioning stops, else continues. 

\begin{figure}
    \centering
    \includegraphics[width=0.7\textwidth]{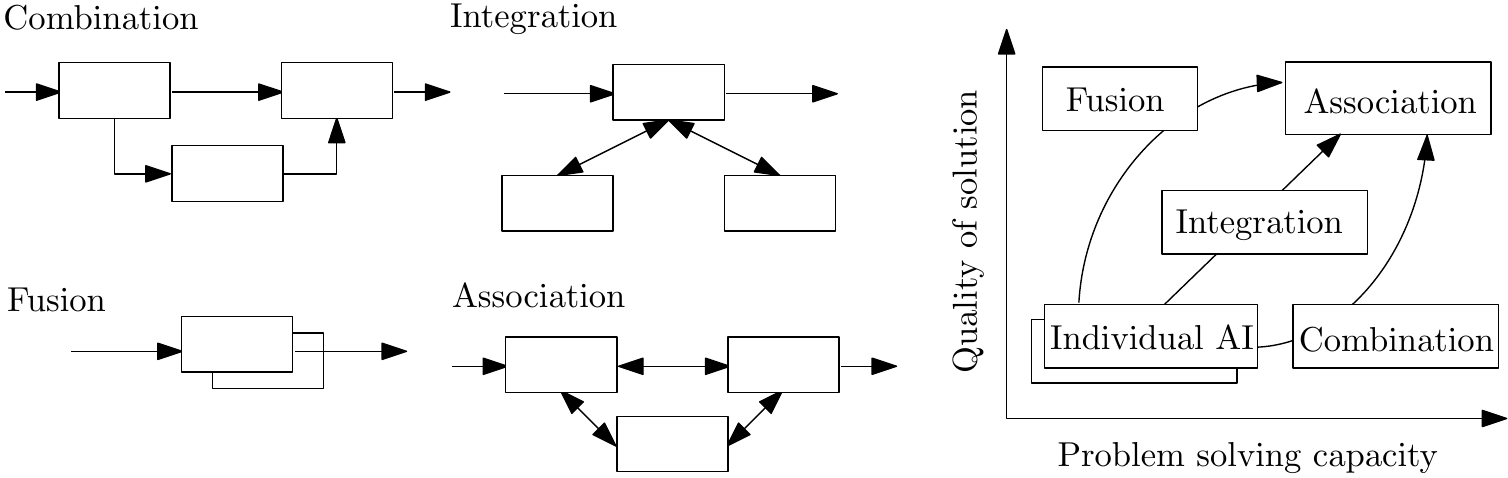}
    \caption{Synergetic artificial intelligence~\citep{funabashi1995fuzzy}}
    \label{fig_synergetic_ai}
\end{figure}

After the input-space partition, FIS is designed via an arrangement of rules and optimization of rule's parameters for inference from FIS. As per Fig.~\ref{fig_fuzzy_paradigm}, FIS design can be performed by combining the FIS concept with GA and NN. Such synergy between two or more methods improves the system's approximation capabilities~\citep{funabashi1995fuzzy}. In this respect, let us revisit four different \textbf{synergetic models} (Fig.~\ref{fig_fuzzy_paradigm}) which indicate four ways of hybridizing artificial intelligence (AI) techniques. The fuzzy system modeling combined with EA, MH, and NN falls in within the synergetic model: (1) \textit{combination}, when the produced rules are optimized by an EA algorithm or an MH algorithm, and (2) \textit{fusion}, when EA or an NN are used to design FIS, i.e., to construct RB.

\section{Genetic fuzzy systems}
\label{sec_gfs}
EA~\citep{back1996evolutionary} and MH~\citep{talbi2009metaheuristics} have been effective in FIS optimization~\citep{cordon2004ten,herrera2008genetic,sahin2012hybrid}. EA and MH are applied to design, optimize, and learn the fuzzy rules, and this gives the notions of evolutionary/genetic fuzzy systems (GFS). The basic needs of GFS are: 
\begin{enumerate}[\hspace{2em}(1)]
    \setlength{\itemsep}{0pt}
    \setlength{\parskip}{0pt}
	\item defining a population structure;
	\item encoding FIS's elements as the individuals in the population;
	\item defining genetic/meta-heuristic operators; and
	\item defining fitness functions relevant to the problem.
\end{enumerate}

\subsection{Encoding of genetic fuzzy systems}
\label{sec_gfs_encoding}
The questions \textit{how to define a population structure} and \textit{how to encode elements of a FIS} as the individuals (called \textit{chromosome}) of the population opens a diverse implementation of GFS. A FIS has the following elements: input-output variables; rule's premises FSs;  rule's consequent FSs and rule's parameters; and the rule set. These elements are combined (encoded to create a vector) in a varied manner that offers diversity in answering the mentioned questions.
  
Lets $ R $ be an RB, a set of $ M $ rules $ \mathbf{r}_i \in  R,  ~\forall i = 1,2,\ldots, M$, then Fig.~\ref{fig_gfs_population} represent two basic genetic population structures: $ \mathcal{S}_a $ and $\mathcal{S}_b $.

\begin{figure}[h!]
	\centering
	\begin{minipage}[t]{0.2\textwidth}
		\begin{equation*}
		\mathcal{S}_a = 
		\begin{pmatrix}
		\mathbf{r}_{1} \\
		\mathbf{r}_{2} \\
		\vdots  \\
		\mathbf{r}_{M} 
		\end{pmatrix}
		\end{equation*}
	\end{minipage}
	\begin{minipage}[t]{0.4\textwidth}
		\begin{equation*}
		\mathcal{S}_b = 
		\begin{pmatrix}
		\mathbf{r}_{11} & \mathbf{r}_{12} & \dots  & \mathbf{r}_{1M} \\
		\mathbf{r}_{21} & \mathbf{r}_{22} & \dots  & \mathbf{r}_{2M} \\
		\vdots & \vdots & \ddots & \vdots \\
		\mathbf{r}_{K1} & \mathbf{r}_{K2} & \dots  & \mathbf{r}_{KM}
		\end{pmatrix}
		\end{equation*}		
	\end{minipage}
	\caption{Population structures: $ \mathcal{S}_a$ and $\mathcal{S}_b $ where $ M $ is total rules in a RB and $ K $ is the population size in $ \mathcal{S}_b$.}
	\label{fig_gfs_population}
\end{figure}

A rule $ \mathbf{r}_i \in R $ that has $ p^i $ FSs, $ A_i $ for T1FS and $ \tilde{A}_i $ for T2FS, for $ i = 1 \text{ to }  p^i $, the $ i^{th} $ rule parameter vector $ \mathbf{r}_i $ may be encoded as~\citep{herrera1995tuning,ishibuchi1997comparison,ojha2016metaheuristic}:
\begin{equation}
\label{eq_michigan_vector}
\mathbf{r}_i = \left\lbrace 
\begin{array}{ll}
\langle A^i_1, A^i_2,\ldots, A^i_{p^i}, c^i_0,c^i_1,\ldots,c^i_{p^i} \rangle& \mbox{for T1FS} \\
\langle \tilde{A}^i_1, \tilde{A}^i_2,\ldots, \tilde{A}^i_{p^i}, c_0^i,s_0^i, c_1^i,s_1^i,\ldots,c_p^i,s_p^i \rangle & \mbox{for T2FS}\\
\end{array}		
\right.
\end{equation} 
where $ A^i $ has two parameters $ m_i $ and $ \sigma_i $ represent center and width of T1FS; and $ \tilde{A}^i $ has three parameters $ m_i $, $ \lambda $, and $ \sigma_i $ represent center, deviation factor, {\color{black}and} width respectively. The variable $ c^i_j$ for $ j = 0 \text{ to } p^i $ are the type-1 rule's consequent weights (parameters) and variable $ c_j^i$ and $s_j^i$ for $ j = 0 \text{ to } p^i $ are the type-2 rule's consequent weights and weights deviations respectively. 

For linguistic fuzzy terms, FS $ A^i $ will take a single integer $ t_i \in \{0,1,2,\ldots\} $ (e.g., the integers 0, 1, and 2, respectively may indicate a linguistic term ``very small,'' ``small,'' and ``large''). For a Mamdani-type rule,~\cite{thrift1991fuzzy} and \cite{kim1995designing} proposed decision matrix [a rule table as per Eq.~\eqref{fig_rule_table}] for fuzzy rules. Such a decision table can be encoded as a genetic vector for the FIS learning~\citep{hadavandi2010integration}.    

\begin{figure}
    \begin{equation*}
    \begin{array}{cc|cccc}
    & & & X_2  & & \\ 
    & & A_1 & A_2  & \ldots & A_{k} \\
    \hline
    & A_1  & B_1  & B_2  & \ldots & B_2   \\
    X_1 & A_2   & B_3  & B_1  & \ldots & B_3  \\
    & \vdots&\vdots&\vdots& \ddots & B_3  \\
    & A_{k} & B_2  & B_3  & \ldots & B_2  \\
    \end{array}
    \end{equation*}
    \caption{Fuzzy decision table for rule contraction (e.g., IF$ X_1 $ is $ A_1 $ AND $ X_2 $ is $ A_2 $ THEN $ Y $ is $ B_2 $) and genetic encoding consisting two input variables $ X_1 $ and $ X_2 $ and an output $ Y $. The decision table has FSs $ A_i, i = 1,2,\ldots $ at the premises part of the rule and at the consequent part of the rule $ B_j, j =1, 2, \ldots $ indicate output fuzzy set in the case of Mamdani-type rule and linear equation [see Eq.~\eqref{eq_type1_consequent} and Eq.~\eqref{eq_type2_consequent}].}
    \label{fig_rule_table}
\end{figure}

Considering genetic fuzzy populations  $ \mathcal{S}_a $ in Fig.~\ref{fig_gfs_population}, the \textit{Michigan approach}~\citep{michigan1982} suggests encoding of a rule $  \mathbf{r}_i $ parameters as a chromosome, $ C_i = \mathbf{r}_i $ in population $ S_a $, i.e., $\mathcal{S}_a = (\mathbf{r}_1,\mathbf{r}_2,\ldots,\mathbf{r}_M)$ of $ M $ rules. Hence, optimization fuzzy system is the reduction of cost function $ c_f (\mathcal{S}_a) $ over entire population. In Michigan approach, the optimization of population is met through mutation and crossover of rules, discarding and adding new rules into the population~\citep{ishibuchi1997comparison}.   

Second genetic fuzzy population  $ \mathcal{S}_b $ in Fig.~\ref{fig_gfs_population} has each chromosome $ C_i $ representing a RB: 
\begin{equation}\label{eq_pittsburgh_chrome}
 C_i = R_i = \{\mathbf{r}_{i1}, \mathbf{r}_{i2}, \dots, \mathbf{r}_{iM}\},
\end{equation} 
a set of $ M $ rules/chromosomes for $ i=1,2,\ldots,K$. Thus, the population $ \mathcal{S}_b = ( R_1, R_2, \ldots, R_K) $ for $ i = 1 \text{ to } K $ enables both ``rule optimization'' and ``rule selection'' opportunities. The rule selection using population $ \mathcal{S}_b $ is known as the \textit{Pittsburgh approach}~\citep{pittsburgh1980} that suggests encoding of fuzzy rule set into a single chromosome, a vectored representation of RB. Pittsburgh approach suggest selecting a subset of $ m $ rules from a set (sometime randomly generated) of $ M $ rules, $ m < M $. In {\color{black}the} Pittsburgh approach, the optimization of {\color{black}the} population is met through mutation and crossover of the RB and {\color{black}by} enabling and disabling the rules in an RB. Hence, the optimization of FIS is the reduction of the cost function $ c_f(C_i = R_i) $ of the chromosomes within the population $ \mathcal{S}_b $~\citep{ishibuchi1997comparison}.

Relaying on the population structure $ \mathcal{S}_a $ and $ \mathcal{S}_b $, numerous literature offers GFS with varied FIS's elements encoding methods: \cite{lee1993integrating} created a \textit{composite chromosome} combining tuple of MF components and rule consequent parameters. Similar composite encoding was performed by~\cite{papadakis2002ga} for TSK-type rules. \cite{wu2006genetic} puts MF's parameter of a type-2 fuzzy rule on a genetic vector. Using the population structure $ S_b $, \cite{ishibuchi1995selecting} created rules as per Eq.~\eqref{eq_pittsburgh_chrome}, where each rule $ \mathbf{r}_{ij}, i=1 \text{ to } K, j=1 \text{ to } M $ takes one of three status: $ 1 $ if $ \mathbf{r}_{ij} ~\in R_i $, $ -1 $ if $ \mathbf{r}_{ij} ~\notin R_i $, and $ 0 $ if $ \mathbf{r}_{ij} $ was created  as a dummy rule.    

\cite{hoffmann1997evolutionary} presented a concept of \textit{messy encoding} by assigning an integer value to FIS's elements while encoding them as a chromosome. For example, the rule IF$ X_1 $ is $ A_2 $ AND $ X_2 $ is $ A_3 $ THEN $ Y_3 $ is $ A_1 $ were encoded as per $ \left\langle  1~2~2~3~3~1 \right\rangle $ where input variables were $ X_1 \rightarrow 1$, $ X_2 \rightarrow 2$, $ X \rightarrow 3$ and FSs were $ A_1 \rightarrow 1$, $ A_2 \rightarrow 2$, $ A_3 \rightarrow 3$. \cite{hoffmann1997evolutionary} argued that such an encoding is benefited from GA since the sequence is messed up by GA operations, and thus creates a diverse rule. \cite{melin2012genetic} amid for obtaining the best rule by assigning a status to TIFIS $ \rightarrow $ 0 and TIFIS $ \rightarrow $ 1, Mamdani-type rule $ \rightarrow $ 0, and TSK-type rule $ \rightarrow $ 1 apart from assigning an integer value to a FS.   

\subsection{Training of genetic fuzzy systems}
\label{sec_gfs_train}
GFS training depended on FIS's encoding, and the GFS training should answer the questions: 
\begin{enumerate}[\hspace{2em}(1)]
    \setlength{\itemsep}{0pt}
    \setlength{\parskip}{0pt}
    \item Which EA/MH algorithms be used?
    \item Whether only a few elements of FIS training is sufficient?
    \item How should EA/MH operators be defined for the encoded GFS?
\end{enumerate} 
The answer to the first question relies on how an individual chromosome was encoded, as well as; it is a matter of choice from the range of optimization  algorithms~\citep{back1996evolutionary,talbi2009metaheuristics}. The answer to second questions was investigative by~\cite{carse1996evolving} with  four GFS learning schemes: (1) learning MF parameters for fix rules; (2) learning rules by keeping MF parameters fix; (3) learning both MF parameters and rules in stages (one after another); and (4) learning both MF parameters and rules simultaneously. \cite{carse1996evolving} concluded that learning in both MF and rule is necessary for solving {\color{black}a} complex system, and GFS benefits from the cooperation of rules. However, it was left for an empirical evaluation to determine the best performance of stage-wise or simultaneous learning. The answer to the third question is subjective to population definition (Fig.~\ref{fig_gfs_population}) and encoding mechanisms (Section~\ref{sec_gfs_encoding}) since a chromosome (solution vector) can be coded in three ways: a \textit{binary-valued} vector, an \textit{integer-valued} vector, and a \textit{real-valued} vector. Accordingly, an EA/MH optimization as per Algorithm~\ref{algo_gen_optimization} is employed, and the algorithm's operators are chosen and designed. 
%\begin{minipage}[t]{\textwidth}
%    \null 
\begin{algorithm}[h!]
    \caption{General optimization procedure}
    \label{algo_gen_optimization}
    \begin{algorithmic}[H]
        \Procedure{Optimize }{$ \mathcal{S}$ for $\mathcal{S}_a \mbox{ or } \mathcal{S}_b$}
        \State For $ n := 0 $, set population $ \mathcal{S}^* := \mathcal{S}^n $; 
        
        $ \mathcal{S}^n \in \mathbb{R}^{K\times L} $ or $ \mathcal{S}^n \in \mathbb{N}^{K\times L} $ or $ \mathcal{S}^n \in \mathbb{Z}^{K\times L}_2$ %$\mathbb{Z}_2 = \{0,1\} $
        
        $ K \rightarrow $ individuals (chromosomes) and $ L \rightarrow$ parameters (genes)
        \State $ c_f^n := $ \textsc{Evaluate} ($ \mathcal{S}^n$)
        
        \Repeat
        
        \State $\mathcal{S}^{n+1}$ := \textsc{Operator} $(\mathcal{S}^n)$
        \State $ c_f^{n+1} := $ \textsc{Evaluate} ($ \mathcal{S}^{n+1}$)
        \If{$(c_f^{n+1} < c_f^n)$ }% If ...
        { $\mathcal{S}^* := \mathcal{S}^{n+1}$} 
        \EndIf %{ do nothing}% ...else...
         \State $ n := n + 1$
         
        \Until { cost $ c_f^n \le c_{f_{min}} $ or iteration $ n \ge n_{max} $}
        
        \Return $\mathcal{S}^*$
        \EndProcedure 
        
        \Procedure{Evaluate }{$ \mathcal{S}$}
        %\For{$i$ = $0$ to $K$} 
        \If{$ \mathcal{S}$ is $\mathcal{S}_a $}
         \State {compute cost $ c_f $ over $\mathcal{S}$ }
        \Else 
        %\If {$ \mathcal{S}$ is $\mathcal{S}_b $} 
        \State {compute cost $ c_f $ over $C_i $, i.e., for each chromosome $ C_i $ in $ \mathcal{S} $}
        %\EndIf
        \EndIf
        
        %\noindent
        \Return $c_f$
        \EndProcedure 
        
        \Procedure{Operator }{$ \mathcal{S}$}
        \If{EA}
        \State Apply \textit{Selection}, \textit{Crossover}, \textit{Mutation}, and \textit{Elitism} on $ \mathcal{S} $
        \Else { for MH}
        \State Apply MH Operator(s) on $ \mathcal{S} $
        \EndIf
        
        \Return $\mathcal{S}$
        \EndProcedure 
    \end{algorithmic}
\end{algorithm}
%\end{minipage}%
%\begin{minipage}[t]{0.5\textwidth}
%    \null
%    \begin{algorithm}[H]
%        \caption{Metaheuristics (MH)}
%        \begin{algorithmic}
%            \Procedure{MH}{$\mathcal{S} \in \{\mathcal{S}_a, \mathcal{S}_b\}$}
%            \State Initialize $\mathcal{S}^* = S^0$             
%            \Repeat
%            
%            \State $W_{n+1} := MH_{Operator}(W_n)$ %\Comment New solution/set of solution generated using Metaheuristic operator
%            \State $\mathrm{W}^*= fittest(\xi(W_{n+1}),\xi(W_n))$ %\Comment $\mathrm{w}^*  \rightarrow$ Optimum solution selection
%            
%            \Until {\textit{ Stopping criteria satisfied}}
%            
%            \Return $\mathrm{W}^*$
%            \EndProcedure 
%        \end{algorithmic}
%    \end{algorithm}
%\end{minipage}

The binary-values vector and the integer-valued vector optimization is both a \textit{combinatorial} and a \textit{continuous} optimization problem, both of which follow the general procedure as per Algorithm~\ref{algo_gen_optimization}. It is a combinatorial optimization when the binary vector and integer vector encoding domain is discreet. That is, the encoding (assignment) of each FIS's element takes either 0 or 1~\citep{ishibuchi1995selecting}, or takes an integer number~\citep{hoffmann1997evolutionary,tsang2007genetic}, and FIS's fitness depends on finding the best combination of FIS's elements. Hence, a global search algorithm like genetic algorithm (GA)~\citep{goldberg1988genetic}, discrete particle swarm optimization (PSO)~\citep{kennedy1997discrete}, or discrete Ant algorithms~\citep{dorigo1999ant} can be employed to optimize binary vector and integer-valued vector. The FIS optimization is a continuous optimization problem when the domain is continuous and FIS optimization is finding the best performing real-valued vector representing the rules parameters~\citep{herrera1995tuning}. Hence, GA~\citep{wright1991genetic}, PSO~\citep{kennedy2011particle}, ACO~\citep{socha2008ant}, or a search algorithm~\citep{yang2010nature} can be used for the real-valued vector optimization as per Algorithm~\ref{algo_gen_optimization}.

The optimization in a binary or an integer vector invites \textit{crossover operator} like single-point crossover, two-point crossover, and composite crossover; and the \textit{mutation operator} like bit flip, random bit resetting,~\citep{goldberg1988genetic}. Whereas, real vector invites crossover operators like uniform crossover, arithmetic crossover~\citep{goldberg1991real,eshelman1993real}. \cite{ishibuchi1999hybrid} exploited both approaches Pittsburgh and Michigan simultaneously, where for the Pittsburgh approaches they designed mutation operator as the Michigan approach for rule generation.

{\color{black} Typically, as an example, for a one-point crossover and for two selected chromosomes $  C_{p1} $ and $ C_{p2} $ (also called parents), two new chromosomes $ C_{o1}  $ and $ C_{o2} $ (also called offspring) are produced by swapping elements of the parent chromosomes (a chromosome is vector few elements) as follows:
\begin{equation}
\label{eq_crossover}
\begin{array}{llcll}
C_{p1} = \{\mathrm{r}_{11}, \mathrm{r}_{12}, \overset{point}{\downarrow} \mathrm{r}_{13}, \mathrm{r}_{14}\} &  \mbox{parent 1} & 
\Rightarrow 
& C_{o1} = \{\mathrm{r}_{11}, \mathrm{r}_{12}, \mathbf{r}_{\mathbf{23}}, \mathbf{r}_{\mathbf{24}}\} & \mbox{offspring 1} \\

C_{p2}  = \{\mathrm{r}_{21}, \mathrm{r}_{22}, \overset{point}{\downarrow} \mathrm{r}_{23}, \mathrm{r}_{24}\} & \mbox{parent 2} & 
\Rightarrow 
& C_{o2}  = \{\mathrm{r}_{21}, \mathrm{r}_{22}, \mathbf{r}_{\mathbf{13}}, \mathbf{r}_{\mathbf{14}}\} & \mbox{offspring 2}
\end{array}
\end{equation} 

Similarly, as an example, for a one-point mutation, one a chromosome $  C_{p1} $ is selected and a new chromosome $ C_{o1}  $ is produced by replacing the an element $ \mathbf{r}_{1j}  $ of the chromosome $  C_{p1} $ by a new element $ \mathbf{r}_{\mathbf{new}} $ or a random element (e.g., flipping 0 to 1 in binary chromosome, replacing a integer by another integer, and replacing a real-value by another random real-value) as follows:
\begin{equation}
\label{eq_mutation}
\begin{array}{llcll}
C_{p1} = \{\mathrm{r}_{11}, \overset{\overset{point}{\downarrow}}{\mathrm{r}_{12}}, \mathrm{r}_{13}, \mathrm{r}_{14}\} &  \mbox{parent 1} & \Rightarrow & C_{o1} = \{\mathrm{r}_{11}, \mathbf{r}_{\mathbf{new}}, \mathrm{r}_{13}, \mathrm{r}_{14}\} & \mbox{offspring 1} 
\end{array}
\end{equation} 
}

The real-valued vector encoding of FIS's elements allows a varied FSs to lie on the same genetic vector. Hence, it is necessary to ensure that each gene (dimension) corresponding to a FIS's element takes a value within a defined interval. For example, in Eq.~\eqref{eq_michigan_vector}, the variables $ m^i_1$ and $ \sigma^i_j $ are MF'S parameter, and they need a defined interval like $ m^i_j \in [m_{left}, m_{right}]$ and $ \sigma^i_j \in [\sigma_{left},\sigma_{right}]$ to control the MF's shape. \cite{cordon1997three} defined \textit{interval of performance} for assuring a boundary for each dimension in the vector.

\cite{martinez2010fuzzy} employed PSO for finding optimal MF parameter of an encoded GFS. \cite{shahzad2009hybrid} combined PSO and GA in a hybrid approach where PSO and GA start with similar populations of rules and swap the best solution iteratively among PSO and GA populations to make communication between  both optimizers. \cite{martinez2015hybrid} extended \cite{shahzad2009hybrid} hybrid PSO and GA approach to optimize T2FIS, and \cite{valdez2011improved} {\color{black}proposed a} hybrid approach of PSO-based FIS and GA-based FIS where {\color{black} depending upon their errors, the two rule types were activated and deactivated during the FIS optimization}. An \textit{empirical evaluation} of bio-inspired algorithms summarized by \cite{castillo2012comparative} suggests that ACO outperformed PSO and GA as GFS optimization. Examples of MH-based GFS implementations are chemical optimization\citep{melin2013optimal}, harmony search~\citep{pandiarajan2016fuzzy}, artificial bee colony optimization~\citep{habbi2015self}, bacteria foraging optimization~\citep{verma2017optimal}.

\subsection{Other forms of genetic fuzzy systems}
\label{sec_gfs_other}
Similar to Michigan approach, also in \textit{iterative rule learning} scheme \citep{venturini1993sia,gonzalez1997multi,ahn2007iterative} and \textit{cooperative-competitive rule learning}~\citep{greene1993competition,whitehead1996cooperative}, each rule of an RB are encoded into separate genotypes, and the population of such genotype leads to the formation of RB iteratively. Iterative learning scheme starts with an empty set and adds rules one-by-one to the set by finding an optimum rule from a genetic selection process. For this purpose, the genetic operators such as mutation and crossover are applied over one or two rule(s) to make offspring rule(s), and the quality of the generated rule(s) is(are) evaluated using a predefined rule quality measure. Therefore, iteratively selecting rules according to rule quality measure criteria for forms an optimum RB in an iterative manner~\citep{venturini1993sia}. 

The cooperative-competitive rule learning is also an RB learning method that determines an optimum RB from competition and cooperation of rules from a genetic/meta-heuristic population. GFS is also implemented as the reinforcement learning system. \cite{juang2000genetic} proposed symbiotic evolutionary learning of fuzzy reinforcement learning system which uses a cooperative coevolutionary GA for the evolution of fuzzy rules from a population of rules. A reinforcement T2FIS optimization was performed by ACO in \citep{juang2009reinforcement}. Aiming cooperation among FIS's components, \cite{delgado2004coevolutionary} {\color{black}split} the genetic population into four separate populations: RB, individual rules, FSs, and FISs. They proposed a coevolutionary GFS  
relying on a hierarchical collaborative approach where each population, cooperatively shared application domain fitness as well as the population's individuals.  

A \textit{fuzzy tree system}, e.g., TSK rule in~\citep{mao2005adaptive,chien2002learning}, allows the rules to be implemented as a \textit{binary tree} and an \textit{expression tree} and the rules tree structures to be optimized by genetic programming (GP)~\citep{koza1994genetic}. ~\cite{hoffmann2001genetic} implemented TSK rule as a local linear incremental model tree, where the algorithm incrementally built the tree while partitioning the input-space using a binary tree formation. On the other hand, the \textit{expression tree} approach for fuzzy rule implementation and optimization  using rules tree population was performed in~\citep{sanchez2001combining,cordon2002new}. Their approach also included a mapping of rule-tree parameters (leaf node) onto a vector for its optimization using simulated annealing~\citep{aarts1988simulated}.

\section{Neuro-fuzzy systems}
\label{sec_nfs}
\index{neuro-fuzzy system}
Since {\color{black}the} early 90s~\citep{jang1991fuzzy,jang1993anfis,buckley1994fuzzy,andrews1995survey,karaboga2018adaptive}, neuro-fuzzy systems (NFS) that represent a fusion of both FIS and NN has been forefront among FIS's research dimensions, especially attributed to its data-driven learning ability which does not require {\color{black}prior} knowledge of the problem. However, NN needs sufficient training pattern to learn, and a trained NN model does not explain how to interpret its computational behavior, i.e., NN's computational behavior is a ``black box,'' which does not explain how the output was obtained  for the input data. On the other hand, FIS requires prior knowledge of the problem and do not have learning ability, but it tells how to interpret its computational behavior, i.e., it explains how the output was obtained for the input data. 

The shortcomings of both NN and FIS can be eliminated by combining them while making an NFS~\citep{feuring1999stability,ishibuchi2001numerical}. Usually, for the rule extraction from NFS, two types of combinations are practiced~\citep{andrews1995survey}: \textit{cooperative NFS} and \textit{hybrid NFS}. The cooperative NFS is the simplest approach closer to combination and association synergetic AI (Fig.~\ref{fig_synergetic_ai}). In cooperative NFS, NN and FIS work independently, and NN determines FIS's parameters from the training data~\citep{sahin2012hybrid}. Subsequently, FIS performs the required interpretation of the data. Hybrid NFS is closer to \textbf{fusion synergetic} AI (Fig.~\ref{fig_synergetic_ai}), in which, both NN and FIS are fused to create a model. Working in synergy improve the learning ability of NFS  since both NN and FIS are independently capable of approximate to any degree of accuracy~\citep{buckley1999equivalence,li2000equivalence}.

NFS are trained in two fundamental manners: supervised learning (See section~\ref{sec_fs_farmework}) and reinforcement learning~\citep{lin1994reinforcement,moriarty1996efficient}. This paper scope includes supervised learning extensively; whereas, the reinforcement learning for NFS is available in \citep{berenji1992learning} through {\color{black}the} implementation of generalized approximate reasoning based intelligence-control and in~\citep{nauck1993fuzzy} through model named NEFCON.

\subsection{Notions of neuro-fuzzy systems}
\label{sec_nfs_notions}
\paragraph{Self-adaptive/Self-organizing/Self-constructing system}
In NFS's context, the \textit{adaptive systems} or the \textit{self-adaptive systems} may refer to the automatic tuning and adjustment of MF's parameters~\citep{jang1993anfis,wang2002self}. Whereas, a system is non-adaptive if human expert determines the MFs and their parameters. Similarly, \textit{self-organizing systems}~\citep{juang1998online,wang1999self} and \textit{self-constructing systems}~\citep{lin2001self} refer to the creation of fuzzy rules and the adaptation of MF's parameters without the intervention of human experts. The implementation of a self-organizing NFS and a self-constructing NFS holds the key to formation appropriate RB~\citep{juang1998online,lin2001self}. 

There are two leaning aspects of self-adaptive NFS: \textit{structural learning} and \textit{parameter learning}~\citep{lin1995neural}. An NFS, therefore, will be self-adaptive if it performs either of these two learning aspects or both of them during learning. In addition to the learning without human intervention, adaptive systems like self-adaptive systems and self-construction systems when strictly refer to online training and incremental learning for every piece of new training data, then the system may be referred to as an evolving fuzzy system (EFS) (see Sec.~\ref{sec_efs}). 

\paragraph{Online learning system/Dynamic learning system} 
Online learning refers to \textit{sample-by-sample} learning. A learning system is an \textit{online learning system} that adapts its structure and parameters each time it sees a training sample rather than seeing the entire training samples set (batch) at once~\citep{jang1993anfis}. Similarly, {\color{black}a} \textit{dynamic learning system} and {\color{black}a} \textit{dynamically changing system} adapts its structure and parameters on receiving new training sample~\citep{wu2000dynamic,wu2001fast}. In a sense, systems that grow their structures by adding MFs nodes and rule nodes are also referred to as the \textit{dynamically growing systems} and the \textit{dynamic evolving systems}~\citep{kasabov2002denfis,kasabov2001line}. FIS's research dimension EFS encompass online and dynamic learning systems (see Sec.~\ref{sec_efs}). 

Another viewpoint refers {\color{black}to} dynamic learning systems as the recurrent fuzzy systems. In other words, the systems which accommodate temporal dependency and whose next (one step ahead) adaptation (learning) is a function of {\color{black}the} model's previous output~\citep{jang1992self,juang1999recurrent}. In FIS research, these jargons are used with diverging context. 

\subsection{Layers of neuro-fuzzy systems}
\label{sec_nfs_layer}
{\color{black}An NFS architecture typically is composed of a maximum of seven layers as shown in Fig.~\ref{fig_nfs_gen_layer} whose layers that can be customized in various forms for both type-1 and type-2 FISs. The type-1 and type-2 FISs only differ in the type of FSs they used. Hence, the variations in type-1 and type-2 NFS architecture depends on the  FS type used at the MF layer $ L_M $ and the methods used at nodes to performs the computation for type-1 and type-2 FSs. Moreover, the type-reduction that requires for type-2 FIS can be implemented at one of the layer indicated available in the consequent part.} 

The implementation of NFS architecture categorized into two types of layers: the layers implementing the \textit{antecedent} part and the layers implementing the \textit{consequent} part of a rule. The number of layers in the design of NFS may vary depending upon how the antecedent and consequent part were implemented. Regardless of a layer mention in Fig.~\ref{fig_nfs_gen_layer} explicitly appear or not in an NFS architecture, the functionality of that layer is accommodated in the either of adjacent layers to that layer. {\color{black}Let us} discuss the functionality of the typical NFS layers:

\begin{figure}
	\centering
	\includegraphics[width=\textwidth]{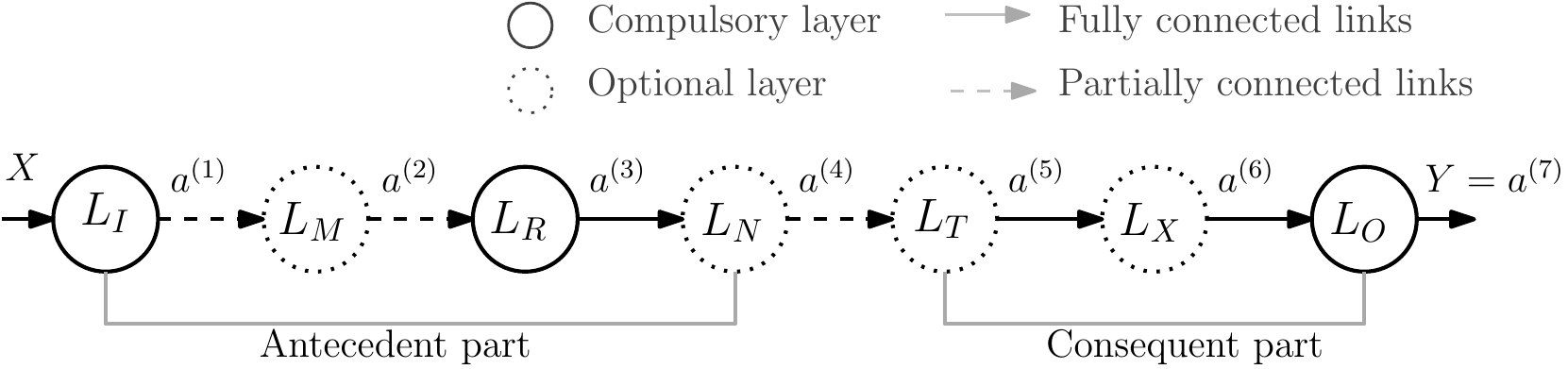}
	\caption{Neuro-fuzzy system architecture (NFS) with {\color{black}a maximum of} seven layers. An NFS receives a vector $ X $ as an input and subsequently propagated it through various layers producing outputs $ a^{(1)} $ to $ a^{(7)} $. The symbols $ L_I $, $ L_M $, $ L_R $, $ L_N $, $ L_T $, $ L_X $, and $ L_O $ stand for the NFS layers input, membership function, rule, normalization, term, extra (additional), and output  respectively.}
	\label{fig_nfs_gen_layer}
\end{figure}

\paragraph*{Input layer ($ L_I $):} A node at the input layer holds $ X \in \mathbf{X} $, and primarily has a function $ f(X) = X $, i.e., the raw input is fed to the next layer without any manipulation. To the best of our literature knowledge, all models agree to the transfer of inputs to the next layer without any modification. Hence, $ a^{(1)}_i = f(X_i) ; 1 \le i \le P$ represents the output a node $ i $ of the input layer, where $ P $ is the dimension of the input-space. However, models agree to either fuzzify inputs at the membership function layer ($ L_M $) or fuzzify inputs by employing a fuzzy weight to the link connecting input layer ($ L_I $) directly to rule layer ($ L_R $). 

The connections/links between $ L_I $ and $ L_M $ is therefore, not fully connected. Rather, each input is connected to its partitioned FSs. Or in the absence of layer $ L_M $ connection between $ L_I $ and $ L_R $ are not filly connected. Such partially connections between $ L_I $ and $ L_M $ or between $ L_I $ and $ L_R $ play {\color{black}an} important role in obtaining diverse rules. 

\paragraph*{Membership function layer ($ L_M $):} A node at the MF layer $ L_M $, also called \textit{fuzzifier} layer, holds $ \mu $, and primarily has a function $ f(X) = \mu(a^{(1)}) = \mu(X) $, i.e., a MF $ \mu(.) $ is applied on input $ X $. MFs are often problem specific. An MF can be a Gaussian function, a triangular function, or a trapezoidal function. MF layer $ L_M $ often refereed as the fuzzification layer that performs fuzzification of the inputs. MF layer is also responsible for the partitioning of the input-space (Fig.~\ref{fig_input_space}). The mapping of inputs to MF layer also helps to overcome the \textit{curse of dimensionality}~\citep{brown1995high}.

Additionally, whether an MF layer $ L_M $ is a separate layer or it acts as a fuzzy weight between the layers $ L_I $ and $ L_R $, the MF layer's operation remains the same. The input to an MF layer is  $ a^{(1)}_i  = X_i$ that has been partitioned into $ p^i $ FSs with $ a^{(2)}_{ij}  = \mu_{ij} (a^{(1)}_j); 1 \le j \le P$ and $1 \le j \le p^i$. Traditionally, inputs partition $ p^i $ is kept fixed. However, automatically determining the input-space partition by using clustering based method gives flexibility to NFS's structural adaptation, and such an act is often refereed as structural learning. It also reflects the notions of the self-constructing system~\citep{lin2001self}. Examples of clustering for input-space partition are: 
K-nearest neighbor~\citep{wang1999self}; mapping constrained agglomerative~\citep{wang2002self}; evolving clustering~\citep{kasabov2001evolving}; and evolving self-organizing map~\citep{deng2003line}.

\paragraph*{Rule layer ($ L_R $):} A node at the rule layer holds a function $ \prod(.) $, and primarily performs $ a_j^{(3)} = \prod_{j=1}^{p^i} (a_j^{(2)}) = \prod_{j=1}^{d} \mu_j(X) $, i.e., a rule layer node typically computes \textit{T-norm} of the previous layer's inputs $ a_j^{(2)} $. Thus, a node at rule layer represents the antecedent (premises) part of a rule that takes $ d $ inputs $ a^{(2)}_j ; 1 \le j \le d$, where $ d \le p^i $ is FS fed to a rule node. 

The inputs $ d $ to a rule node may or may not be equal to the total number of partitions $ p^i $ of an input $ a^{(1)}_i $. It also indicates that connections between layer $ L_M $ and layer $ L_R $, which are often partly connected, govern the diversity of the rules being formed. It also gives flexibility for a structural adaptation (structural learning) in {\color{black}the} fuzzy system being realized. For example, an algorithm may starts with no rule, and {\color{black}it only recruits} a rule node if it is necessary during its online leaning~\citep{tung2011safin,juang1998online}. The output of a rule layer node or in other words antecedent part of a rule is denoted as $ a^{(3)}_k = \prod_{j=1}^{d} a^{(2)}_{kj}; 1 \le k \le M $. Hence, the output of $ M $ rules ($ M $ nodes at the rule layer) can be denoted as $ a^{(3)}_k $.  

\paragraph*{Normalization layer ($ L_N $):} Normalization layer $ L_N $ computes the firing strength of the rules, which is $ a^{(4)}_i = a^{(3)}_i / \sum_{k=1}^{M} a^{(3)}_k; i = 1, 2, \ldots, M$. Therefore, the number of nodes at the layer $ L_N $ is thus equal to the number of nodes at the layer $ L_R $ and the connection between $ L_R $ and $ L_N $ is fully connected. %The weights $ w_{ik} $ between $ L_R $ and $ L_N $ are set to 1 due to normalize.

\paragraph*{Term/Consequent layer ($ L_T $):} The nodes as term layer $ L_T $ computes consequent part of a rule. Thus, the number of nodes at term layer $ L_T $ are the same as the number of nodes at the layer $ L_R $ and layer $ L_N $. Each node at this later has a function $ \varphi(.) $ and the definition of $ \varphi(.) $ depends on the FIS's type implemented, e.g., Mamdani or TSK. In other words, what type of function implemented at the nodes of layer $ L_T $~\citep{horikawa1992fuzzy}. Assuming that nodes at the layer $ L_T $ are constant, then the output $ a^{(5)}_k = a^{(4)}_k c_k $, where $ c $ is a constant. Another type of consequent/term implementation of TSK (first-order liner equation) node, where $ a^{(5)}_k = a^{(4)}_k \left( \sum_{i = 1}^{n} x_i c_{ki} +c_{k0} \right) $.

\paragraph*{Additional layer ($ L_X $):} Additional layer $ L_x $ is infrequent in NFS architecture design, which performs specific operation $ \psi(a^{(5)} $ producing the output $ a^{(6)} $. The definition of $ \psi(.) $ in ~\citep{park2002fuzzy} is a polynomial neural network. Whether the additional layer $ L_X $ is present ($ a^{(6)} = \psi(a^{(5)} $) or absent ($  a^{(6)} = a^{(5)}$), the input to {\color{black}the} output layer $ L_O $ is $ a^{(6)} $.

\paragraph*{Output layer ($ L_O $):} For a single output problem, output layer $ L_O $ holds a single node that usually is the summation of incoming inputs to the node, i.e., $ a^{(7)} = \sum_{k = 1}^{R} a^{(6)}_k$. Therefore, the output node act as a \textit{defuzzifier}. Hence, the operation at {\color{black}the} output layer with a function $ \theta(.) $ applied on $ a^{(6)} $ is to obtain NFS's output $ Y = a^{(7)} = \theta( a^{(6)})$.

\subsection{Architectures of neuro-fuzzy systems}
\label{sec_nfs_arch}
\subsubsection{Feedforward designs} 
\label{sec_nfs_arch_ff}
Feedforward NFS architecture have forward connections from one layer to another and have at least three layers: input, rule{\color{black},} and output. Therefore, the simplest NFS architecture is IRO, i.e., \textbf{I}nput, \textbf{R}ule, and \textbf{O}utput layer architecture. 

\paragraph*{IRO architecture:} 
\cite{masuoka1990neurofuzzy} represented IRO NFS architecture as a combination of the input-variable-membership net, the {\color{black}rule-net}, and the output-variable-membership net. {\color{black}Moreover}, the fuzzy rules are directly translated into NNs where the nodes at layer $ L_I $ realize rule's antecedent MFs, the node at layer $ L_R $ represent fuzzy operation (e.g., AND), and the nodes at layer $ L_O $ realize the rule'consequent part. This type of representation can easily be translated back and forth between fuzzy rules and NNs. However, the expert intervention will be required in the NFS construction.  

\begin{figure}
	\centering
	\subfigure[]
	{
		\includegraphics[width=0.3\columnwidth]{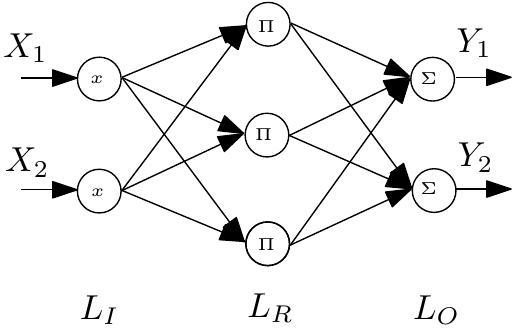}
		\label{fig_nfs_arch_iro}%	
	}
	\subfigure[]
	{
		\includegraphics[width=0.3\columnwidth]{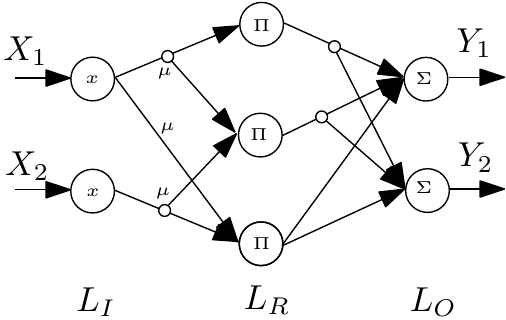}
		\label{fig_nfs_arch_iro1}
	}
	\subfigure[]
	{
		\includegraphics[width=0.3\columnwidth]{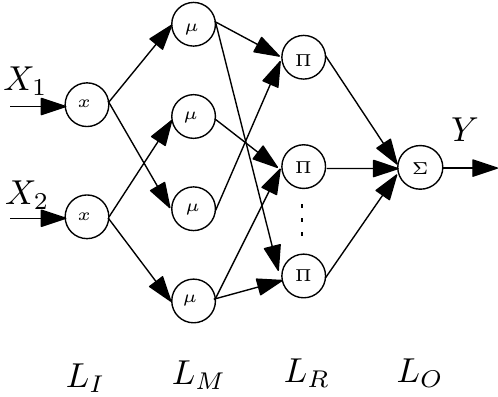}
		\label{fig_nfs_arch_imro}
	}
	
	\subfigure[]
	{
		\includegraphics[width=0.45\columnwidth]{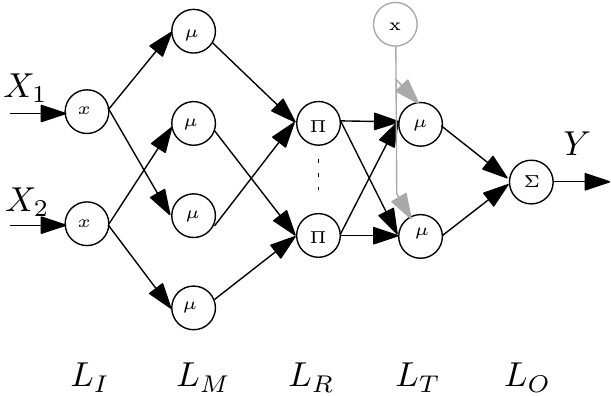}
		\label{fig_nfs_arch_imrto}
	}
	\subfigure[]
	{
		\includegraphics[width=0.45\columnwidth]{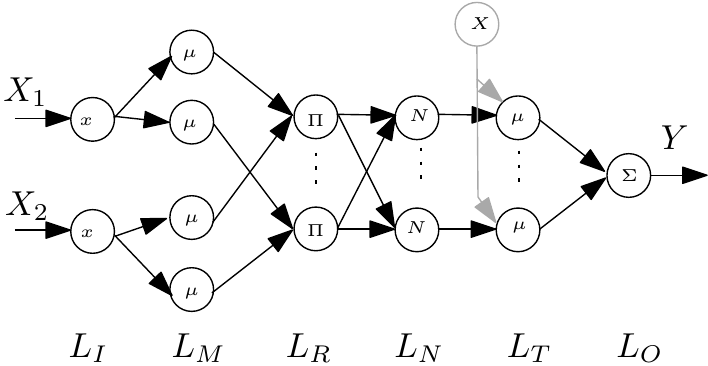}
		\label{fig_nfs_arch_imnrto}
	}
	\caption[Feedforward NFS architectures]{Feedforward NFS architectures.}
	\label{fig_nfs_ff}
\end{figure}

\cite{buckley1995neural} showed a design of three-layered IRO NFS architecture and implemented IRO NFS for \textit{discrete fuzzy systems} and \textit{non-discrete fuzzy systems} (Fig.~\ref{fig_nfs_ff}a). Being a three-layered architecture, their discrete IRO NFS architecture implemented fuzzy rules as the links between layers $ L_I $ and $ L_R $, and the layer $ L_O $ processed incoming signals from the transfer functions (nodes at layer $ L_R $) using some aggregation function $ \theta(.) $. The rules in the discrete IRO NFS is, therefore, can run in parallel. However, for a large input, the rules can grow to huge unmanageable size for a low discrete factor~\citep{buckley1995neural}. On the other hand, in a non-discrete IRO fuzzy system, the hidden layer $ L_R $ nodes represent the rules and the links between $ L_I $ and $ L_R $ are set to 1. The nodes at the output layer $ L_O $ represent an aggregation of {\color{black}the signals} from $ L_R $.

In~\citep{buckley1995neural} IRO NFS, the fuzzy rules are implemented as a whole {\color{black}either for the links} between $ L_I $ and $ L_R $ or for the nodes at $ L_R $. Whereas, \cite{nauck1997neuro} proposed a three-layered IRO NFS architecture with the link between layers $ L_I $ and $ L_R $ and between layers $ L_R $ and $ L_O $ representing MFs also called fuzzy weights. In other words, the links between $ L_I $ and $ L_R $ fuzzify the inputs before feeding them to nodes at $ L_R $ and defuzzify them before feeding them to nodes at $ L_O $. 

IRO NFS architecture shown in Fig.~\ref{fig_nfs_ff}b was proposed for specific problems like classification and approximation bearing abbreviations NEFCLASS~\citep{nauck1997neuro} and NEFPROX~\citep{nauck1999neuro} respectively.
{\color{black}NFSs} are shown in Fig.~\ref{fig_nfs_arch_iro1} implement the links as the fuzzy weights that improve the NFS \textit{interpretability} since it avoids more than one MFs to be assigned to {\color{black}similar} terms~\citep{nauck1997neuro}.

\paragraph*{IMRO architecture:} 
NFS design IMRO: input, membership, rule, and output architecture  (Fig.~\ref{fig_nfs_ff}c) directly computes the output $ a^{(6)} $ of FIS by assigning weight to the links between layer $ L_R $ and $ L_O $~\citep{lin2001self,wu2001fast}. %model SCFNN\citep{lin2001self} GD-FNN\citep{wu2001fast}
The IMRO NFS architecture by~\cite{wang1999self} is a four-layered configuration, where layers $ L_I $ and $ L_M $ fuzzify the inputs. The layer $ L_R $ consists of two nodes: $ a^{(6)}_1 $  and $ a^{(6)}_2 $. The first node $ a^{(6)}_1 $ computes a weighted sum  $ a^{(6)}_1 = a^{(3)}_1 = \sum_{i=1}^{m} w_i a^{(2)}; 1 \le i \le m$ of the incoming inputs from $ L_M $, where $ m $ {\color{black}is the} number of nodes at layer $ L_M $, and $ w_i $ is the links' weights between  $ L_M $ from $ L_R $. The weight $ w_i $ represent consequent part's FS's center. The second node $ a^{(6)}_2 $ computes sum $ a^{(6)}_2 = a^{(3)}_2 = \sum_{i=1}^{m} a^{(2)}$ of incoming inputs from $ L_M $, where the link's weight between layers $ L_M $ and $ L_R $ are set to 1. The output layer $ L_O $ node, therefore, realizes $ a^{(7)} = a^{(6)}_1/a^{(6)}_2$. 

\paragraph*{IMRNO/IMRTO architecture:} 
The five-layer NFS architecture (Fig.~\ref{fig_nfs_ff}d) adds a layer $ L_N $ or $ L_T $ between the layers $ L_R $ and $ L_O $ to perform fuzzy quantification via rule normalization or via a fuzzy term nodes~\citep{kasabov1997funn,kim1999hyfis}. Example of an IMRNO NFS architecture with a normalization layer $ L_N $ between $ L_R $ and $ L_O $ is in~\citep{kasabov1997funn}. %model D-FNN~\citep{kasabov1997funn}
Whereas, an IMRTO NFS architectures with a term layer $ L_T $ is the common practice. The nodes at the layer $ L_T $ compute fuzzy outputs, and the links between $ L_R $ and $ L_T $ represent firing strength (\textit{confidence factor}) of the rules at $ L_R $~\citep{kasabov1997funn,kim1999hyfis,kasabov2001line,kasabov2002denfis}.  %FuNN/2~\citep{kasabov1997funn}, HyFIS~\citep{kim1999hyfis}, EFuNN~\citep{kasabov2001line}, and DENFIS~\cite{kasabov2002denfis}

Contrary to IMRNO and IMRTO architectures, the five-layered NFS presented by \cite{leng2006design} is an \textbf{IRNTO} architecture that has layers $ L_I $, $ L_R $, $ L_N $, $ L_T $, and $ L_O $. In IRNTO model, nodes at layer $ L_R $ combine both MF layer $ L_M $ and rule layer $ L_R $, and the term layer $ L_T $ between $ L_N $ and $ L_O $ perform a TSK-type consequent operation for the rule.

In general, five-layer NFS architecture implements $ L_I $, $ L_M$, and $ L_R $ as its rule's antecedent, where nodes at $ L_R $ implements rule's $ \prod (.) $ or AND function. The layer $ L_T $ and $ L_O $ implements the rule's consequent part and perform defuzzification. However, apart from $ \prod (.) $ and defuzzyification at layers $ L_R $ and $ L_T $ example of $ \min(.) $ operator at $ L_R $ and $ \max(.) $ operator at $ L_T $ is available in~\citep{shann1995fuzzy}.

\paragraph*{IMRNTO architecture:} IMRNTO NFS architecture is the most popular NFS architecture, which is attributed to the efficiency and explicit presence of FIS's components in the architecture~\citep{jang1991fuzzy,horikawa1992fuzzy}. ANFIS being the most popular implementation of IMRNTO NFS~\citep{jang1993anfis}. 
%..ANFIS \citep{jang1993anfis} GNN-FISs: \cite{jang1991fuzzy}
IMRNTO NFS are six-layered architecture with layers $ L_I $, $ L_M $, $ L_R $, $ L_N $, $ L_T $ and $ L_O $. The functioning of the nodes are described in Sec.~\ref{sec_nfs_layer}.

\paragraph*{IMRNTXO architecture:} 
Beyond IMRNTO NFS architecture, IMRNTXO NFS architecture includes an additional layer that performs certain computation receiving inputs from layer $ L_T $ and fed the computed output $ a^{(6)} $ to the node(s) at layer $ L_O $. The model:  \textit{modified fuzzy polynomial neural network}~\citep{park2002fuzzy} is in an example of such seven-layered architecture, where a polynomial NN that implements a polynomial function (like bilinear and biquadratic), which resembles consequent part of TSK type.

{\color{black}
Of general NFS architecture in Fig.~\ref{fig_nfs_gen_layer}, five variation in NFS architectures formation is shown in Fig.~\ref{fig_nfs_ff} are IRO (three layers), IMRO (four layers), IMRNO/IMRTO/IRNTO (five layers), IMRNTO (six layers), and IMRNTXO (seven layers). The choice of a particular variation in NFS formation has its advantages and disadvantages. For example, IRO architecture limits itself to three layers, and that restricts it to compute entire FIS operations on a few nodes. IRO architecture computes input fuzzification at input layer node that limits it to mix with multiple FSs and when input fuzzification takes place at the links between input and rule layer an input mix with all available FSs for a fully connected network, that limits a proper fuzzy partitioning. However, IRO architectures are easy to implement and they can be translated to fuzzy rules easier than more complex architecture. 

The four-layer IMRO architecture solves the fuzzy partition issues that may appear in the layer IRO architecture since it adds a membership layer between input and rule layer. In IMRO architecture, the weight optimization of between the input and membership layer may lead to direct optimization of the FS shapes in addition to a comparatively more variation in rule design (Fig.~\ref{fig_nfs_ff}c) than IRO architecture.

The five-layer and six-layer architectures IMRNO/IMRTO/IRNTO and IMRNTO add FIS components more explicitly than the three-layer and four-layers architectures. Thus, they offer more efficient ways to design of NFS as a FIS system. In five-layer architecture forth layer is chosen as a normalization layer or term layer, whereas the six-layer architecture uses both normalization and term layers to its architecture. Moreover, seven layer architecture IMRNTXO adds an extra layer for a special purpose such as a polynomial network operation as an extra layer.  

The difference among the various architectures is apparent regards to the increasingly explicit presence of the FIS components into the architectures with a higher number of layers than the architectures with a lower number of layers. The explicit presence also offers efficiency and opportunity to optimization NFS architecture to individual FIS component.}

\subsubsection{Feedback/Recurrent designs}
\label{sec_nfs_arch_fb}
\begin{figure}
    \centering
    \subfigure[]
    {
        \includegraphics[width=0.4\columnwidth]{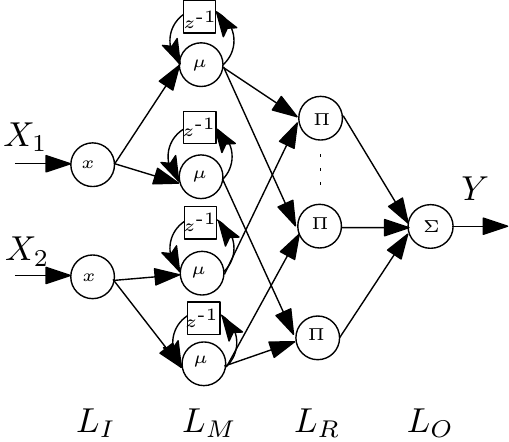}
        \label{fig_nfs_arch_imro_rev_m}%	
    }
    \subfigure[]
    {
        \includegraphics[width=0.4\columnwidth]{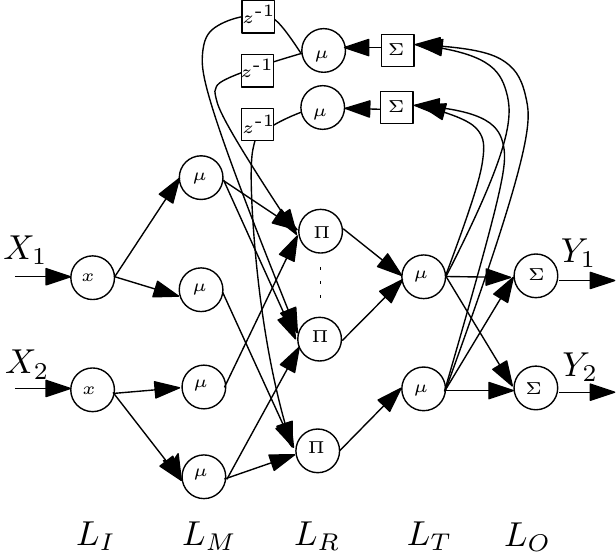}
        \label{fig_nfs_arch_imro_rev_r}
    }
    \caption[Feedback NFS architectures]{Feedback NFS architectures.}
    \label{fig_nfs_fb}
\end{figure}
Unlike feedforward architecture that models static system and can adapt to a dynamic system  through a prepared training set and incremental learning methods, the feedback/recurrent design accommodates dynamic system directly into its structure (model's learning) either via an external feedback mechanism or via an internal mechanism~\citep{mastorocostas2002recurrent}. {\color{black} The recurrent/feedback NFS (RNFS) helps in {\color{black}the} implementation of the systems that require its output $ Y_t $ at time step $ t $ is to be fed as the input $ Y_{t+1} $ to the network at next step $ t+1 $. The external feedback RNFS is the most straightforward implementation of RNFS architecture where rules receive network output directly as its input in the next time step. Whereas, the internal feedback NFS design fits when a system require memory elements to be implemented as an FIS component to define the temporal relation of a dynamic system. That is, in the next step, RNFS's particular layer (e.g., membership, rule, or term layer) receives input $ y_{t+1} $ (the output $ y_{t} $ ot the previous time step).} The example of both recurrent NFS (RNFS) categories are as follows:     

\paragraph*{External feedback RNFS}
Let denote external RNFS architecture design by~\textbf{$\overleftarrow{\textbf{I \dots R \ldots O}}$}, which indicates that the NFS architecture may remain the same as a basic feedforward NFS,  but the system incorporates the feedback through one or multiple sources. Such a feedback adaptation can be incorporated through the learning algorithm like temporal backpropagation, e.g., recurrence in ANFIS~\citep{jang1992self}. 
\paragraph*{Internal feedback RNFS}
Internal feedback NFS design \textbf{I$\overleftarrow{\textbf{M}}$RO} ~\citep{lee2000identification} takes inputs to its MF node as per $ a^{(1)}(k+1) = a^{(1)}(k) + a^{(2)}(k-1) $. That is, the recurrence occurs at the MF nodes which enabled the membership layer node to operate as a memory unit that extends the NFS ability for the temporal problems (Fig.~\ref{fig_nfs_fb}a). Unlike \textbf{I$\overleftarrow{\textbf{M}}$RO} design, the memory element in the design \textbf{IM$\overleftarrow{\textbf{RT}}$O} are added at rule layer, and the nodes are called \textit{context element} (Fig.~\ref{fig_nfs_fb}b) that accommodates both spatial firing from MF nodes and feedback (temporal) firing from term nodes~\citep{juang1999recurrent}.
\textbf{IMR$\overleftarrow{\textbf{T}}$O} is the third type of internal feedback design implements, where term nodes  act as the memory element~\citep{mastorocostas2002recurrent}.

\subsubsection{Graph and network based architectures:}
Apart from the two class of architecture, a general graphical model for information flow was proposed as the \textit{Fuzzy Peri nets}~\citep{looney1988fuzzy}. Fuzzy Peri net is directed graph with nodes (neurons) and transition bars (links) that are enabled or disabled when neurons fire. The NFS feedforward and feedback architecture, therefore, can be {\color{black}thought of} as the special case of graphical representation. Additionally, examples of FISs combined with \textit{adaptive resonance theory} (ART) to create fuzzy ART architecture is available in~\citep{carpenter1991fuzzy}. Similarly, {\color{black}FISs} were also fused with the \textit{min-max network} to create a fuzzy min-max network architecture \citep{simpson1992fuzzy} and fused with \textit{radial basis function} (RBF) network to created fuzzy RBF architecture~\citep{cho1996radial}.

%\textbf{Network node design}
%\cite{kwan1994fuzzy}, \citep{lin1995neural}
%\textbf{Problem specific NFS design}
%
%\textbf{NFS for linguistic variables}
%Contribution of NFS concept for linguistic variables:  \citep{yager1999implementing}
%Network Linguistic process: \citep{bortolan1998architecture}
%--------
%Counterproagation Network: whether it is somewhat different than backpropagation:\citep{chang2001counterpropagation}

\section{Hierarchical fuzzy systems}
\label{sec_hfs}
GFS is a process of empowering FISs for automatic optimization and learning, which focuses {\color{black}on} designing FIS's components. NFS is NN inspired and it enables the arrangement of FIS's components into a network-like structure. Whereas, the hierarchical fuzzy systems (HFS) is a hierarchical arrangement of two or more small standard FISs, (say fuzzy logic units - FLU denoted as $ N_i $ in Fig.~\ref{fig_hfs_fuzzy}) into a hierarchical structure. Hence, HFS invites the following questions:
\begin{enumerate}[\hspace{2em}(1)]
    \setlength{\itemsep}{0pt}
    \setlength{\parskip}{0pt}
    \item What are the basic advantages of arranging small FLUs?
    \item What are the possible ways to arrange FLUs?
\end{enumerate} 

\subsection{Properties of hierarchical fuzzy systems}
\label{sec_hfs_property}
Let's take Fig.~\ref{fig_rule_table} example, a standard practice of rule set formation for FISs. Now assume the rule table in Fig.~\ref{fig_rule_table} has $ P = 2 $ inputs, and each input takes $ k = 3$ FSs. Hence, the number of rules will be $ k^P = 3^2 $, which means that the number of rules grow exponential at the rate of $ k^P $, and subsequently, the number of parameters to be optimized grow exponentially. This phenomenon is known as the \textit{rule explosion} and the \textit{curse of dimensionality}. The rule explosion reduces the basic FIS's property: \textit{interpretation}, i.e., the reasoning as to how the output was obtained for the inputs become unknown. It also led to infeasible \textit{computation} in both space (rule storage space) and time~\citep{torra2002review}. Additional, in both GFS and NFS, the input-space partitioning play a crucial role in the FIS's construction and both GFS and NFS have to employ an external method like clustering to reduce the input space dimensionality. \cite{hoffmann2001genetic} illustrated a GP-based binary-tree like input-space partition that hierarchically partition inputs space, but they form a standalone FIS. 

\begin{figure}
    \centering
    \subfigure[General HFS]
    {
        \includegraphics[scale= 0.6]{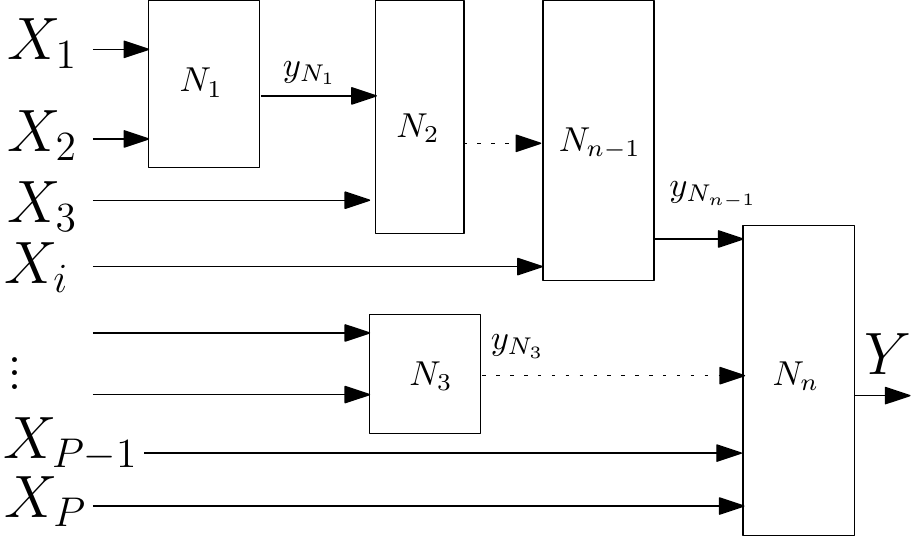}
    }\quad
    \subfigure[Cascaded HFS]
    { 
        \includegraphics[scale= 0.6]{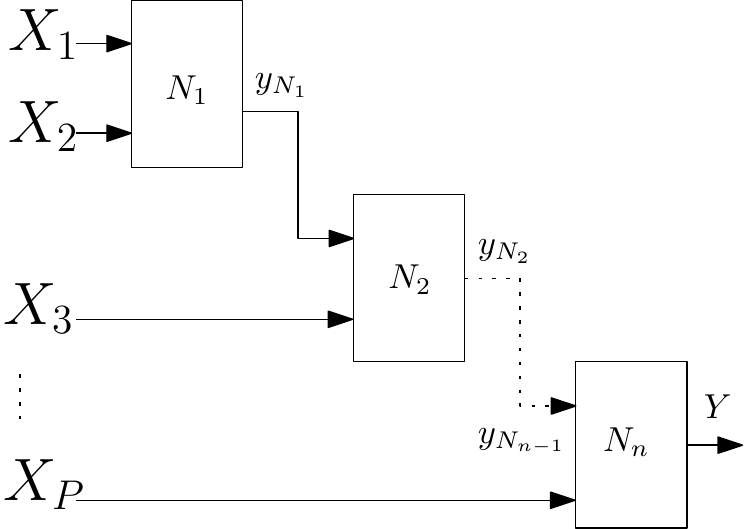}
    }\quad
    \subfigure[Tree-like HFS]
    { 
        \includegraphics[scale= 0.5]{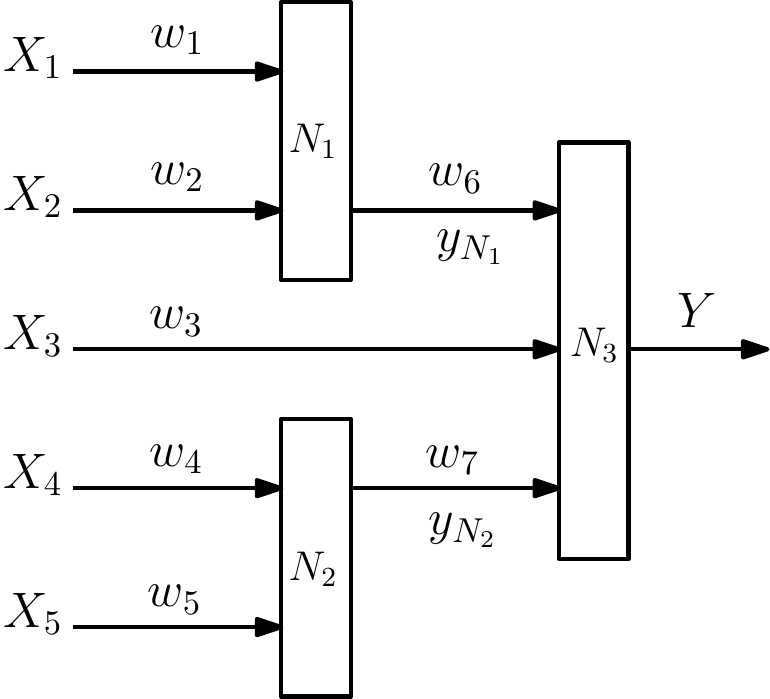}
    }
    \caption{Typical hierarchical fuzzy systems (HFS): (a) A general combination of low-dimensional fuzzy systems called fuzzy logic units (FLUs) $ N_i $ in multiple stages and (b) an incremental combination of low-dimensional FLUs~\citep{chung2000multistage}. The inputs $ X_1 $, $ X_2 $, $ \ldots, X_P $ in (a) and (b) are traditionally selected by applying an expert's knowledge~\citep{raju1991hierarchical}. (c) Tree-like HFS (also called aggregated output HFS) with three FLUs $ N_1 $, $ N_2 $, and $ N_3 $ takes inputs $ X_1 $, $ X_2 $, $ X_3 $, $ X_4 $, and $ X_5 $~\citep{ojha2018multiobjective}.}
    \label{fig_hfs_fuzzy}
\end{figure}

\cite{raju1991hierarchical} initiated the design of hierarchical FIS (HFS) that was composed of low-dimensional fuzzy subsystems, called fuzzy logic unit (FLU). One of the arguments for HFS was to overcome the \textit{curse of dimensionality}~\citep{brown1995high} and stop the \textit{rule explosion} by combining several sub-fuzzy systems receiving only a few inputs from the whole set of inputs (Fig.~\ref{fig_hfs_fuzzy}) This allows the reduction of fuzzy rules, total system's parameters, and the computation time. Also, the hierarchical design of fuzzy subsystem  found to have a universal approximation ability~\citep{wang1999analysis,zeng2005approximation,wang1998universal}.   
Moreover, HFS offers intelligent control over the system for a dynamically changing domain environment~\citep{karr2000synergistic}. Such a control may be implemented by allowing one of the FLU in HFS to act as performance checker and optimize entire HFS with its feedback.       

Torra et al.~\cite{torra2002review} reviewed HFS that presents the following  observations for the defining HFS architecture: If some functions are not  decomposable then HFS design may not be possible, but for some functions, HFS is proved to be a universal approximator~\citep{wang1998universal}. If the system's non-linearities are independent, then separate FLUs can be constructed. If no preference is given to the order (importance) of variables, then a general HFS is trivial to design, else preferred variables should go at beginning stages of hierarchy. If MF for a variable is sharp, then more MFs should be defined for that variable~\citep{wang1999analysis}. Finally, the interpretability of HFS might become unknown while reasoning (defuzzification) are repeated at multiple stages~\citep{maeda1996investigation}. 

\cite{wang2006survey} summarized literature work to investigate the reasoning transparency for the intermediate variable generated by defuzzification at the FLUs at different stages, and concluded that a little work had been done to understand intermediate variables fully. However, in this view, the HFS's interpretability can be improved, provided sufficient monotonicity of FLUs concerning the inputs~\cite{magdalena2018hierarchical}. \cite{kouikoglou2009monotonicity} concluded that under certain conditions~\citep{won2002parameter}, the single-stage HFS's output is monotonic. Hence, it is sufficient for the monotonicity of multi-stage HFS design. 

\subsection{Implementations of hierarchical fuzzy systems}
\label{sec_hfs_impl}
The classification of HFS types is intuitive since the HFS design is a {\color{black}modular} arrangement. Thus, HFS have variety in design and modeling~\cite{lee2003modeling}. A general HFS design is any combination of FLUs in stages (Fig.~\ref{fig_hfs_fuzzy}). Special cases of general arrangement can be a cascaded (incremental) design of FLUs~\citep{chung2000multistage} and chain wise FLUs arrangement~\citep{domingo1997knowledge}. 

\paragraph{Converting standard FISs to HFS} 
Standard FISs can be transformed to HFS. \cite{joo2003method} transformed standard FIS which has $ k^P $ rules for $ P $ inputs and $ k $ FS. And the FIS  was3-D matrix (cube) with each 2-D slice (rule table as per Fig.~\ref{fig_rule_table}). Each 2D slice was then transformed to a FLU. Similarly, a type-2 HFS having multiple levels of FLUs implementation was proposed by~\cite{hagras2004hierarchical} for automatic mobile robot navigation behaviors control. It divided layers (stages) for managing navigation behaviors and as the navigation behaviors in multiple levels. First level accommodated low level behaviors and highest level act as the coordination. An HFS with combining layer wise rule in a hierarchical manner was presented by \cite{fernandez2009hierarchical} where rules were arranged in two layers and for each layer, KBs were generated by linguistics rule generation method and the rules were selected by GA.

\paragraph{NNs inspired HFS}

{\color{black}\cite{joo2005class} presented a feedforward NN like HFS design} that takes the previous layer FLUs input to the THEN part of the rule in current layers. \cite{yu2007system} implemented a hierarchical fuzzy NNs approach and trained hierarchical fuzzy NNs using the backpropagation-like algorithm. Unlike HFS by~\cite{joo2005class} where FLUs are proposed to arranged in a network-like structure, in HFS by~\cite{yu2007system}, each FLU is an NFS. \cite{mohammadzadeh2016modified} proposed self-structuring hierarchical type-2 NFS (SHT2FNN). Similar to \cite{yu2007system} approach, in SHT2FNN, each FLU is a self-structuring NFS and that the arrangement of FLUs was a cascaded design.  

 %\cite{hwang2014adaptive} implemented an adaptive fuzzy hierarchical sliding-mode control method where each layer receives a sliding surface from its previous layer. In this arrangement, top layer (final stage FLU) accommodates all the subsystems outputs. Similarly, formulated a  \cite{zhao2017adaptive} hierarchical sliding-mode control for a , multiinput multioutput system.

\paragraph{Automatic HFS formation}
A majority HFS takes a manual design; whereas, \cite{chen2007automatic} explained the structural optimization of the HFS where hierarchical arrangements of low-dimensional TSK-type FISs were optimized using probabilistic incremental program evolution~\cite{salustowicz1997probabilistic}.  \cite{ojha2018multiobjective} proposed a hierarchical fuzzy inference tree approach (HFIT$^M$) that has an automatic arrangement of FLUs using GP for type-1 and type-2 TSK FISs. HFIT$^M$ offered automatic selection of the input variables for each FLUs and that the order of input variables are automatically determined along with the HFS structure's automatic determination.

\section{Evolving fuzzy systems}
\label{sec_efs}
Standard FIS, GFS, NFS, and HFS are the concepts of creating a system for modeling and learning from data. Often data are dynamic, i.e., domain environment changes in time. Therefore, any systems relied on the data needs to be updated. Hence, GFS, NFS, and HFS system that embraces and adapt itself to the dynamic nature of data is evolving fuzzy systems (EFS). EFS systems embed provisions for dynamic (online) training of systems for streaming (real-time) data~\cite{angelov2009evolving,kasabov1998evolving}. In EFS, a system incrementally evolves fuzzy rules incoming new data~\citep{lughofer2011evolving}. Hence, EFS answers the following questions:
\begin{enumerate}[\hspace{2em}(1)]
    \setlength{\itemsep}{0pt}
    \setlength{\parskip}{0pt}
    \item How a fuzzy system adapt to the incoming data stream?
    \item Which components of a fuzzy system are made flexible to evolve?  
\end{enumerate}

\subsection{Incremental learning of fuzzy systems}
\label{sec_efs_learn}
Incoming data stream are processed to train and test a system incrementally, i.e., \textit{incremental learning}, \textit{dynamic learning} and \textit{online learning}~\citep{losing2018incremental}. It is a strategy for data-driven training and testing of a system incrementally for unseen data without re-training the system entirely from scratch. Incremental learning, therefore, should take care of \textit{noise} and \textit{concept drift} in the data stream~\citep{schlimmer1986incremental}. Noise and concept drift in data may be introduced over time compared to initial training data, i.e., input feature does not describe the output class~\citep{gama2014survey}. Often for the non-stationary environment, the relation between input features and output class change over time~\citep{elwell2011incremental}. Fuzzy rules are capable of evolving to accommodate noise and concept drift for the data stream~\citep{baruah2011evolving}. 

EFS approaches manages the concept drift by first detecting the drift (data shift) and then reacting to the drift. \cite{lughofer2011handling} describes detection of drift includes tracking of fuzzy rules \textit{age} and evolving the rules' antecedent part using evolving-clustering~\citep{angelov2004approach,angelov2010evolving} and evolving-vector quantization~\citep{lughofer2008extensions,lughofer2007evolving} and evolving the rules' consequent part. Moreover, the gradual concept drift that is hard to detect can be managed by incremental rule splitting~\citep{lughofer2018incremental}.

\subsection{Implementations of evolving fuzzy systems}
\label{sec_efs_impl}
\begin{figure}
    \centering
    \includegraphics[width=\textwidth]{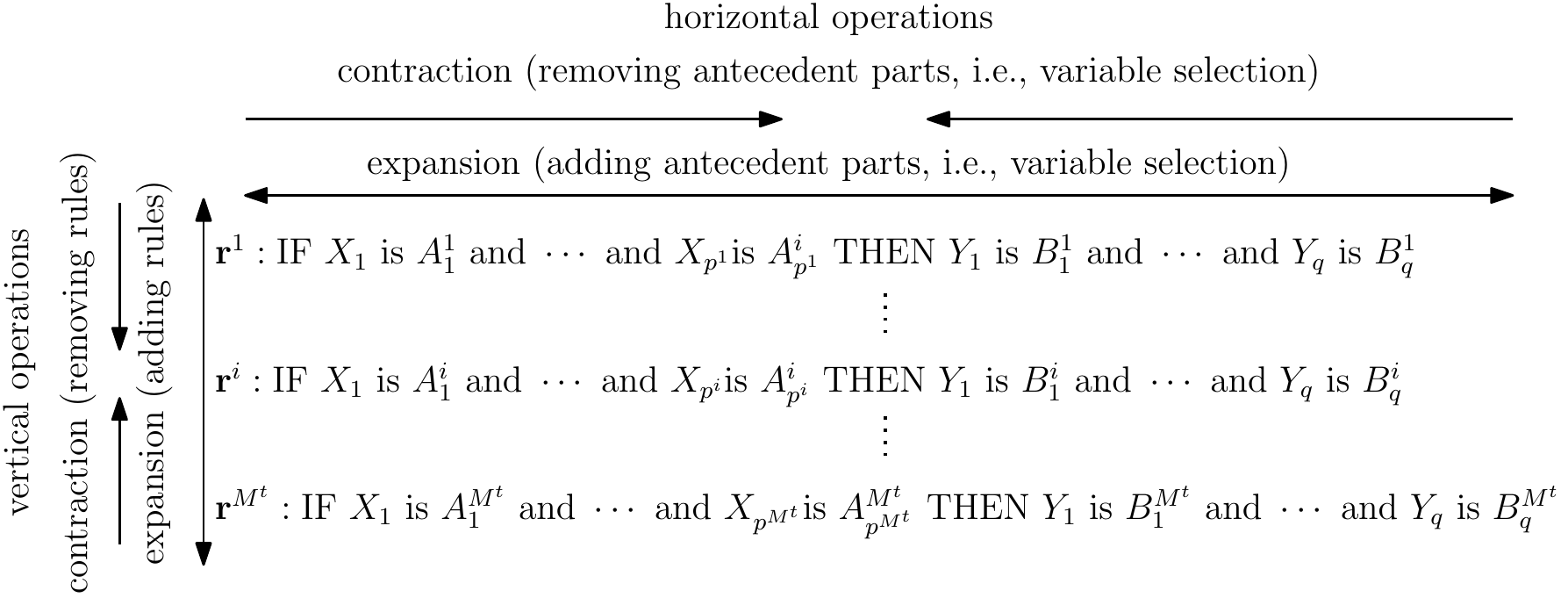}
    \caption[Evolving fuzzy system: typical dynamic RB learning]{Evolving fuzzy system: typical dynamic RB learning. The symbols are as follows: $ p^i $ and $ q^i $ indicates {\color{black}the} number of inputs $ X $ and outputs $ Y $ variable to a rule $ i $, respectively; $ M^t $ is {\color{black}the total rules} in the RB at time $ t $; $ A^i $ is the fuzzy set at the antecedent part of a rule; and $ B^i $ is the function of the consequent part of a rule~\citep{angelov2004approach}.}
    \label{fig_efs_rules}
\end{figure}

A fuzzy system, irrespective of its being a standard FIS, a GFS, an NFS, or an HFS when implements incremental learning capability should evolve (alternately we may say modify or refine) itself internally for it is incrementally fed unseen data stream~\citep{angelov2004approach}. 
There are two broad categories have been investigated to incorporate EFS concepts in FISs: standard FISs to EFS~\citep{angelov2009evolving} and NFS to EFS~\citep{kasabov2001evolving}: 

\paragraph{FISs $ \rightarrow $ EFS}
In standard FISs, incremental learning is offered by adding or removing rules in an RB (vertical direction manipulation), or by adding or removing antecedent part of the rules in an RB (horizontal direction manipulation) as shown in Fig.~\ref{fig_efs_rules}. In Fig.~\ref{fig_efs_rules}, each rule may acquire $ p^i $ variables using the evolving clustering methods, i.e., the number of variables are determined automatically; whereas, in traditional clustering methods, the number of clusters has to be predetermined. Moreover, the antecedent part of the rules may expand and contract based on incoming data. Similarly, the number of rules may also be reduced or increased by adding or deleting rules from the RB. Hence, {\color{black}the total} rules $ M^t $ in the RB are time dependent~\citep{angelov2004approach}.   

Evolving Takagi-Sugeno fuzzy system (eTS)~\citep{angelov2004approach,angelov2010evolving} is an example of EFS that modify and update its RB on arrival of every piece of {\color{black}the} new data point. It employs online clustering that checks the influence of new data point on the input space partitioning and then it modifies the cluster centers and add new rules in RB. Accordingly, it modifies {\color{black}the} rule's consequent parameter. Similarly, flexible fuzzy inference systems (FLEXFIS)~\citep{lughofer2008flexfis} rely on the incremental update of cluster centers for the arrivals of new data and accordingly adapt its antecedent and consequent parameters. 

The principle of applying incremental clustering to verify new data point and its influence leads to several EFS designs like evolving participatory Kernel recursive least squares model~\citep{lima2010evolving} that uses participatory learning~\citep{lima2006participatory}. Both eTS and participatory learning concepts were used for evolving a rule having multivariate Gaussian functions at its antecedent part that preserve information between input variable interactions~\citep{lemos2011multivariable}. \cite{zhou2007autonomous} offers an \textit{evolving self-organizing map} for clustering that replaced the online clustering method in eTS for constructing an evolving EFS classifier.

%~\cite{lughofer2015generalized}Generalized smart evolving fuzzy systems
\cite{lughofer2013line} investigated interpretability aspects such as distinguishability, simplicity, consistency, coverage and completeness, feature importance levels, rule importance levels and interpretation of consequent. They concluded that a very few EFS approach takes care of complexity reduction such as the elimination of redundancies~\citep{lughofer2011line} to improve interpretability conclusion.

\paragraph{NFS $ \rightarrow $ EFS}
EFS concept applies to NFS paradigms. NFS is evolved dynamically based on every incoming new data~\citep{kasabov2001evolving,kasabov2002denfis}. Such systems are also called \textit{self-evolving} NFS or \textit{adaptive} NFS~\citep{angelov2004flexible}. In self-evolving NFS, the network design has two main parts: {\color{black}antecedent and consequent}. For every incoming data, the antecedent part learns new information by using unsupervised means of learning through cluster evolving method, and accordingly, the consequent part weights are updated to accommodate the new information contained in the incoming data.

Incremental learning in NFS is a similar concept as incremental learning in NNs where growing and pruning network architecture may refer to rule addition and deletion in NFS architecture, and learning of weights at the output layer may refer to learning NFS architecture consequent layer parameters~\citep{wang1992smart,feng2009error}. As of topological level refinement, evolving NFS architecture may refer to augmented topological concepts~\citep{stanley2002evolving}. Evolving NFS and evolving standard FISs has similar incremental mechanism when it comes to input space partitioning. Both have a major dependency on evolving clustering method (ECM) of inputs space~\citep{kasabov2002denfis}. %Fuzzy ARTMAP~\cite{carpenter1992fuzzy} is a kind of representation of incremental learning that generate new decision clusters for new data that are different from previously seen data. 

Dynamic evolving neuro-fuzzy inference systems, DENFIS~\citep{kasabov2001evolving,kasabov2002denfis,kasabov2001evolving}  relay on ECM and refine its rule layer ($ L_R $) in its IMRNO/IMRTO architecture (see~\ref{sec_nfs_arch_ff}) by operations like: creating new rule nodes, deleting existing rule nodes, updating existing rules, aggregating two or more rule nodes. Like DENFIS, self-organizing fuzzy neural network, SONFIN~\citep{juang1998online} employ clustering methods for input space partitioning and methods of parameter optimization of rules consequent parts. However, it starts with no rule in its structure by examining center of first incoming input data and the first rule and subsequently perform a check on every a new piece of data for the aggregated firing strength of existing rules in the structure an if the aggregated firing strength is found week (i.e., lower than a pre-defined threshold) new rule are added to the structure. Structure adaptation on inputs space clustering are: sequential adaptive FISs~\citep{tung2011safin}, generalized dynamic NFS~\citep{wu2001fast}, self-evolving interval type-2 NFS~\citep{juang2008self}, self-organizing NFS~\citep{wang1999self}, recurrent self-evolving NFS with local feedbacks~\citep{juang2010recurrent}, and mutually recurrent interval type-2 NFS~\citep{lin2013mutually}.

%\paragraph{HFS $ \rightarrow $ EFS}
In summary, the EFS needs the following steps:
\begin{enumerate}[\hspace{2em} Step 1: ]
    \setlength{\itemsep}{0pt}
    \setlength{\parskip}{0pt}
    \item Construct an initial fuzzy system in \textit{batch mode} or construct EFS in \textit{online mode} from scratch with no rule in RB initial, and add rules as per step 2.
    \item Apply \textit{incremental clustering} or an \textit{incremental inputs-space partitioning} mechanism to check on incoming data. 
    \item Evolve (add, delete, modify) existing EFS rules or rule structure as per step 2.
    \item Continue step 2 and step 3 for every new piece of data. 
\end{enumerate} 

\section{Multiobjective fuzzy systems}
\label{sec_mfs}
Multiobjective fuzzy system (MFS) enable a fuzzy system to manage multiple objectives associated with the system, and that system may have been modeled using any of these concepts: standard FIS, GFS, NFS, HFS or EFS. That is, an MFS empowers a FIS to manage multiple objectives which may come from two directions: one, from the problem domain, and two, from the system's own trade-off. This review discusses the objectives inherent {\color{black}in FIS} itself. 

A data-driven modeling system owns a single objective: \textit{cost function}. The  cost function can be the approximation error minimization or the classification accuracy maximization. The minimization or maximization of the cost function is subjected system's parameter optimization. For a FISs, the cost function is subjected to rules and rule's parameters optimization. The primary goal of a FIS is to draw reasoning from the system, i.e., FIS should have \textit{interpretability} property. 

Additionally, FIS often gains \textit{complexity} when having numerous rules. For NFS complexity can be the nodes interconnections. The complexity reduction and interpretability improvement are often FIS's objectives. An MFS deals with multiple objectives that are conflicting {\color{black}with} each other. Hence, MFS answers the following questions:
\begin{enumerate}[\hspace{2em} (1 )]
	\setlength{\itemsep}{0pt}
	\setlength{\parskip}{0pt}
	\item What are the multiple objectives that are conflicting associated with the system?
	\item Which two or more objectives associated with the system should be optimized?
	\item How to define the selected two or more objectives functions?  
	\item How to manage {\color{black}conflicting} objectives?
\end{enumerate} 

\subsection{Multiobjective trade-offs}
\label{sec_mfs_trade_off}
Two basic approaches manage the trade-offs of multiple objectives: (1) by aggregating multiple objectives $ c_{f_1}, c_{f_2}, \ldots, c_{f_m} $ into a single scalrized objective function $ c_f $, e.g.{\color{black},} sum $ c_f = \sum_{i=1}^m c_{f_i} $, product $ c_f = \prod_{i=1}^m c_{f_i} $, or weighted sum $ c_f = \sum_{i=1}^m c_{f_i} w_i $, etc.~\citep{ishibuchi2007multiobjective}; and (2) by optimizing multiple objectives $ c_{f_1}, c_{f_2}, \ldots, c_{f_m}$ simultaneousy~\citep{deb2002fast}. These two approaches may respectively be called \textbf{non-Pareto approach} and \textbf{Pareto approach}~\citep{coello2007evolutionary}. The Pareto-based approach, since optimize function simultaneously, offers a \textit{nonominated} solution where no one objective is dominant than the other, whereas, in non-Pareto approach, one objective may dominate the other. Therefore, a Pareto-based approach is a formidable option to obtain a generalized solution (a FIS)~\citep{zitzler1999multiobjective}. 

\begin{figure}
    \centering
    \subfigure[]
    {
        \includegraphics[width=0.4\columnwidth]{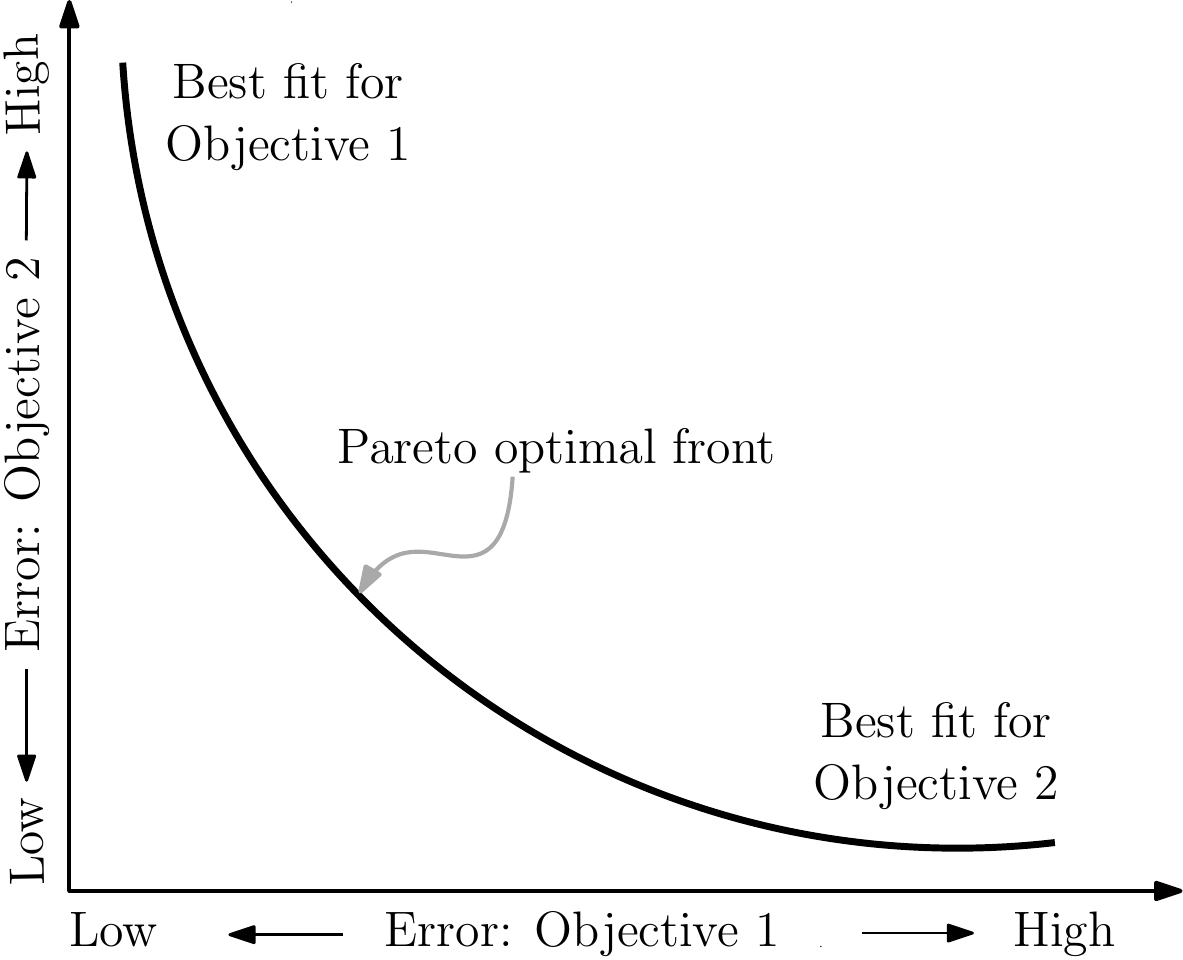}
    }\quad
    \subfigure[]
    { 
        \includegraphics[width=0.4\columnwidth]{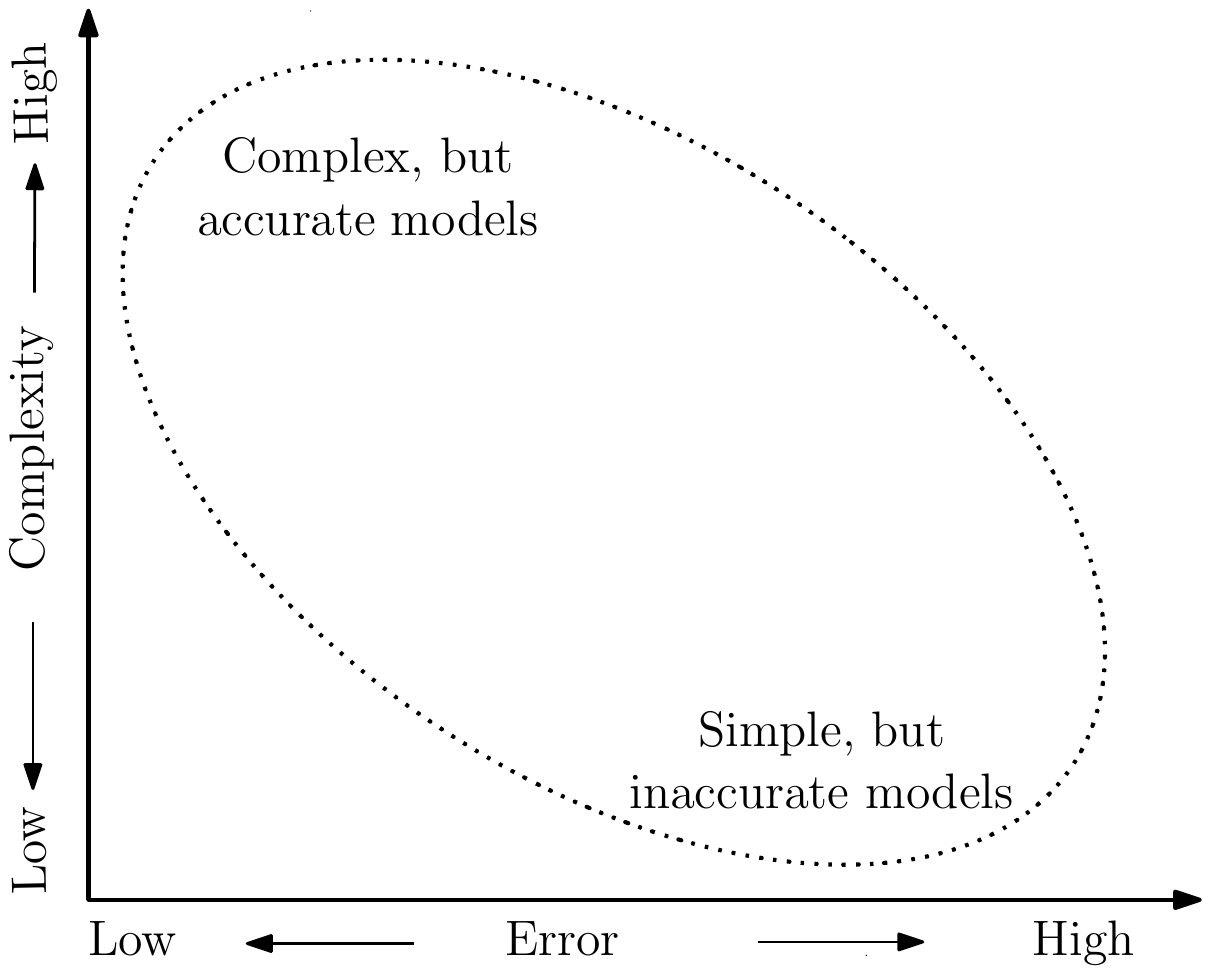}
    }
 	\caption{Multiobjective trade-offs: (a) Solutions on Pareto optimal front in a two-objective solution space. (b) Fuzzy inference system solution space where objective 1 is the error (1-accuracy) of the system and objective 2 is the complexity (interpretability) of the system.}
 	\label{fig_mfs_dom}
 \end{figure}

Fig.~\ref{fig_mfs_dom} shows two-objectives solution space where solutions lying on Pareto optimal front are feasible solution (Fig.~\ref{fig_mfs_dom}a), and the solutions within the boundary of may vary in their objective, i.e., a solution (a FIS, $ R $) may be complex but accurate and another solution may be simple but inaccurate (Fig.~\ref{fig_mfs_dom}b). 
Hence, no single solution exists that may satisfy both objectives. Therefore, multiobjective optimization takes the form: $ \min \{ c_{f_1}, c_{f_2}, \ldots, c_{f_m}\}$ or $ \max \{ c_{f_1}, c_{f_2}, \ldots, c_{f_m} \} $, i.e., a multiobjective optimization needs to be performed as:
\begin{flushleft}
    $\mbox{minimize (or maximize) } \{c_{f_1}(R), c_{f_2}(R), \ldots, c_{f_m}(R)\}$\\
    $\mbox{subject to } (R) \in S,  \nonumber $\\
\end{flushleft}
\noindent where $m \ge 2$ is the number of objective functions $c_{f_i}: \mathbb{R}^m \to  \mathbb{R}_{\ge 0}$. The vector 
of objective functions is denoted by $ \mathbf{c}_f = \langle c_{f_1}(R), c_{f_2}(R), \ldots, c_{f_m}(R) \rangle $. The solution $R = \{ \mathbf{r}_1, \mathbf{r}_2,\ldots, \mathbf{r}_M \}$ is a set of $ M $ fuzzy rules
belonging to the set of solution space $S$. The set of rules $ R $ indicate a solution of GFS, NFS, HFS, or EFS. The word ``minimize'' or ``maximize'' indicates the minimization (or maximization) all the objective functions simultaneously. 

A nondominated solution is one in which no one objective function can be improved without a simultaneous detriment to at least one of the other objectives of the solution. The nondominated solution is also known as a Pareto-optimal solution.

\begin{mydef}
    \label{def_Pareto_dom}
    Pareto-dominance - A solution  $ R_1 $ is said to dominate a solution $ R_2 $ if $ \forall i = 1,2,\ldots,m $, $ c_{f_i}(R_1) \le c_{f_i}(R_2) $, and there exists $ j \in \left\lbrace 1,2,\ldots,m\right\rbrace  $ such that $ c_{f_j}(R_1) < c_{f_j}(R_2) $.
\end{mydef}

\begin{mydef}
    Pareto-optimal - A solution $ R_1 $ is called \textit{Pareto-optimal} if there does not exist any other solution that dominates it. Pareto-optimal front is a set of \textit{Pareto-optimal} solutions.
\end{mydef} 

\subsection{Implementations of  multiobjective fuzzy systems}
\label{sec_mfs_impl}
For FISs together with accuracy $ c_{accuracy} $ (performance improvement) of systems the interpretability $ c_{interpretability}  $ (transparency and reasoning improvement) is always a desirable objective. Additionally, complexity $ c_{complexity} $ of system play another important role in improving computational time, as well as, it may play a role in {\color{black}a} system's interpretability improvement~\citep{jin2000fuzzy}.

{\color{black}\cite{guillaume2001designing} summarized the three necessary condition for the interpretability: (1) fuzzy partition must be readable, (2) rule set must be small, (3) rule set must be incomplete. These three necessary conditions were described by~\cite{guillaume2001designing} to be met by two ways: a good rule induction method and FIS's structure and parameter optimization.

Defining a system's accuracy is subjected to the domain of problem, whereas interpretability definition is a challenging task~\citep{guillaume2001designing,casillas2013interpretability}. Especially when condition (1) set by~\cite{guillaume2001designing}. However, }the definition of interpretability and complexity may be straightforward like reduction of {\color{black}rules and  parameters} in some case, but in some cases, it can be challenging when interpretability and complexity may mean the interaction of rule and interconnections of {\color{black}the} node~\citep{ishibuchi2007multiobjective}. {\color{black}Hence, based on the definition of a rule vector in Eq.~\ref{eq_michigan_vector}, the objectives interpretability (or complexity) can be formulated as:
\begin{equation}
\label{eq_mo_bjectoives}
c_{interpretability}  = \left\lbrace
\begin{array}{l}
count_{elements}(\mathbf{r}_i)\\
count_{rules}(\mathcal{S})
\end{array}\right.
\end{equation}
where $ \mathbf{r}_i $ a rule vector in Eq.~\eqref{eq_michigan_vector} and $ \mathcal{S} $ is a set (population) of rules. 

A question ``how the best parameter and best rules are to be selected'' arises from the use of $ count(\cdot) $ as an interpretability objective function is answered by employing the following methods: (1) variable selection: regularity criteria, geometric criteria, individual discriminant power, and entropy variable index; and (2) rule base optimization: incremental rule generation, rule merging, and statisc based rule evaluation~\citep{guillaume2001designing}. Moreover, multiobjective optimization in conjuncture with these methods controls both accuracy and interpretability.}

An evolutionary multiobjective algorithm like nondominated sorting GA (NSGA-II)~\citep{deb2002fast} or strength Pareto EA (SPEA) ~\citep{zitzler1999multiobjective} can be applied for optimizing FIS's multiple {\color{black}objectives} simultaneously. The \textbf{interpretability against accuracy} and \textbf{complexity against accuracy} are typical evolutionary multiobjective optimization scenarios~\citep{ishibuchi2007analysis}. 

\cite{ishibuchi1997single} formulated two objectives as the maximization of accuracy and minimization of a number of rules while applying a multiobjective GA for obtaining a set on nondominated solutions. Similarly, ~\cite{alcala2007multi} considered the number of rule minimization as the interpretability measures and mean squared error minimization as the accuracy measure while applying SPEA for optimizing these two objectives simultaneously to conclude that the multiobjective led to the removal of the rules having little importance. Moreover, Pareto-based multiobjective optimization algorithms were used to optimize accuracy-complexity trade-off as a number of rule reduction and the accuracy maximization~\citep{ishibuchi2006evolutionary,gacto2009adaptation,ducange2010multi}. 

Similarly, in~\citep{cordon2003multiobjective,wang2005multi,munoz2008automatic,alcala2009multiobjective,antonelli2011learning}, simultaneous learning of knowledge-base was proposed, which included feature selection, rule complexity minimization together with approximation error minimization, etc. In~\citep{antonelli2012genetic}, a co-evolutionary approach that aims towards combining multiobjective approach with single objective approach was presented. In a co-evolutionary approach, first, a multiobjective GA determined a Pareto optimal solution by finding a trade-off between accuracy and rule complexity. Then, a single objective GA was applied to reduce training instances. ~\cite{fazzolari2013review} summarized research works focused on multiobjective optimization.

\cite{pulkkinen2010dynamically} defined transparency of fuzzy partitions as the interpretability indicator where transparency was described as the MFs number reduction, MF's diversity, MF's normality assurance, and MF's shape symmetry assurance. The transparency-accuracy trade-off was then optimized as multiobjective optimization. Similarly, \cite{rey2017multi} took a detailed description of interpretability to put interpretability-accuracy to test. In fact, they took three objectives: maximizing accuracy, maximizing interpretability, and maximizing rule relevance. {\color{black}\cite{rey2017multi} {\color{black}measured} the accuracy as a means squared error minimization and defined the interpretability in terms of the reduction of (1) number of rules, (2) number of MFs, (3) incoherence among rules (increase rule's coherency), (4) irreverent rules. A details study on rule coherence and rule relevance are offered by~\cite{dubois1997checking} and \cite{yen1999simplifying}, respectively.}

For NFS, HFS, and the FISs that have structural representation, the structure simplification is one of the objectives which may indeed indicate to a number of rule reduction, parameter reduction, and rule interaction simplification like {\color{black}the number of MFs} reduction. \cite{ojha2018multiobjective} employed multiobjective genetic programming (MOGA) for the simplification of the model structure while improving accuracy and improving diversity in the rules. These three objectives are conflicting {\color{black}with} each other. Therefore, the Pareto optimal set of nondominated solutions offer to chose a solution as desirable in the problem's context.       

\section{Challenges and opportunities}
\label{sec_fuzzy_rev_challenges}
With the success of FIS and research in FIS's multiple directions like GFS, NFS, HFS, EFS, and MFS comes multiple challenges and multiple opportunities: %These challenges and opportunities may come from the following:  
  
\paragraph{Nature of fuzzy systems}
The basic FIS's property is its ability to model with an explanation as to how  for an input the model achieves its objectives. This FIS's property is referred to {\color{black}as}  \textit{transparency and interpretability}~\cite{casillas2013interpretability}. Although a lot of work has been done to define and preserve transparency and interpretability of a FISs~\cite{guillaume2001designing,gacto2011interpretability,cpalka2014new}, often with {\color{black}the} growing number of fuzzy rules while solving complex problems and with the complex interactions of rules within models (e.g., NFS and HFS), FISs {\color{black}lose} their reasoning {\color{black}ability}. Hence, preserving transparency and interpretability remains a challenging issue in FIS's modeling~\citep{buckley1994fuzzy,andrews1995survey,herrera2008genetic,fazzolari2013review}.    

Additionally, the basic unit of FISs is MF. The designs and the assignments of MFs to inputs variables have to be a careful art since it influences the FIS's reasoning. For the most modeling methods, expert knowledge is required, in both cases: when the input-space partitioning is performed manually, and when the input-space partitioned performed using clustering, number of the clusters has to be predetermined. Therefore, efficient automatic input-space partitioning can play a crucial role in FISs modeling~\citep{jain2010data}. Moreover, research on incorporating FSs like hesitant FSs~\citep{torra2010hesitant}, intuitionistic FSs~\citep{atanassov1986intuitionistic} and their type-2 versions~\citep{mendel2002type} to data-driven FIS's modeling present further opportunities.

\paragraph{Nature of data}
The quality of the data-driven model relies on the quality of data that is sufficient and balanced~\citep{bellman2013dynamic}. Usually, FISs are good at managing noisy and imprecise data, but most data source offer \textit{unstructured} data~\citep{feldman2007text}, and experimental research often produce \textit{heterogeneous} data~\citep{castano2001global}. Additionally, {\color{black}for pattern recognition problems,} the training data supply for generalized extrapolation and modeling are sometimes \textit{insufficient}~\citep{jackson1972interpretation,pradlwarter2008use} and sometimes are \textit{imbalanced}~\citep{chawla2002smote,alshomrani2015proposal}. {\color{black} These issues are dealt with a method like synthetic minority over-sampling technique (SMOTE)~\citep{chawla2002smote}. Here, rather than generating synthetic samples, training method may be modified for the rule induction sensitive to imbalanced datasets such as the cost-sensitive rule-based system~\citep{lopez2015cost} and the metacognitive learning scheme~\citep{das2015evolving}.}

Other crucial issues are \textit{high dimensionality} and \textit{abundance}. Some fuzzy systems like HFS offer a solution to curse of dimensionality to some extent. However, the volume of data is a challenge for FISs to maintain its interpretability-accuracy trade-offs~\citep{ishibuchi2007analysis}. In addition to high dimensionality, some training data are \textit{non-stationary} that show drift in concept when data are fed at a regular interval, i.e., data are fed in the stream. The EFS manage using a refinement of the system for evolving FISs over time through incremental learning~\citep{kasabov1998evolving,angelov2009evolving}, and iterative rule learning algorithm like multi-stage genetic fuzzy system~\citep{gonzalez1997multi} are potential EFS and may use population-based  incremental learning~\citep{baluja1994population}.

\paragraph{Nature of algorithms}
The optimization algorithms such as the EAs~\citep{cordon2004ten}, MHs~\citep{castillo2012optimization,valdez2014survey},  least squares method~\citep{wang1992fuzzy}, gradient descent algorithm~\citep{jang1993anfis} are  
exhaustively used for optimization FISs. However, their efficiency is subjective to formulation (encoding) of FISs. A variety of encoding mechanism has been proposed in the past (Section~\ref{sec_gfs}) which indicate that FISs being the integration of various decomposable components offer a reach possibility of encoding and their optimization.

{\color{black}Mostly present approached under MFS treat two objective interpretability (complexity) and accuracy (error) for the simultaneous optimization while applying evolutionary multiobjective. Popular evolutionary multiobjective like NSGA-II~\citep{deb2002fast} is good at optimizing two or three objectives, but their performance decreases as the number of objectives increases~\citep{purshouse2003evolutionary}. Hence, challenges to simultaneously optimize multiple objectives related to FIS such as FIS's interpretability,  consistency, coherence, complexity, and accuracy by applying evolutionary multiobjective that deals with multiple objectives like NSGA-III~\citep{deb2014evolutionary}.}

\paragraph{Nature of network}
The FISs like  NFS, HFS, and EFS offer a connectionist model that beers \textit{network structure}. Algorithms built the network structure based on input-space partitioning, and the intuition for input-space partitioning arrive from the problem domain. While doing so, for high dimensional and complex problems, the network structure may grow big enough to lose interpretability. Therefore, multiobjective optimization embedding growing and pruning  mechanism may check interpretability-accuracy trade-off. Additionally, connectionist models optimization using EAs needs more attention~\citep{seng1999tuning}. 

Rules extracted from connectionist models beers complex interactions, therefore, an algorithm capable of explaining the complex interaction of rules will help to solve complex problems having abundance data and unstructured data without losing the basic FIS's properties. {\color{black}In this view, future research direction \textit{deep fuzzy systems} (DFS) can be defined in two ways: 
    
First straightforward definition, specifically in the context of pattern classification tasks, would be a system whose input data  go through a \textit{convolution} process which is coupled with GFS, NFS, HFS, or EFS. This definition is similar (and inspired by) to the deep convolutional neural network~\citep{lecun2015deep}. Thus it may be termed explicitly as the \textit{deep convolution fuzzy systems}. 

Second, a system will be a \textit{deep fuzzy system} if it relies on multiple layers of network architecture as described in~\citep{hinton2012deep}. A multilayer NFS architecture, e.g., by~\cite{deng2017hierarchical} and a multi-stage HFS architecture, e.g., by~\cite{ojha2018multiobjective} may fit this definition. The DFS dimension of FIS research is still an open problem to explore and innovate.}

\section{Conclusions}
\label{sec_fuzzy_rev_con}
\begin{figure}
    \centering
    \includegraphics[width=0.8\textwidth]{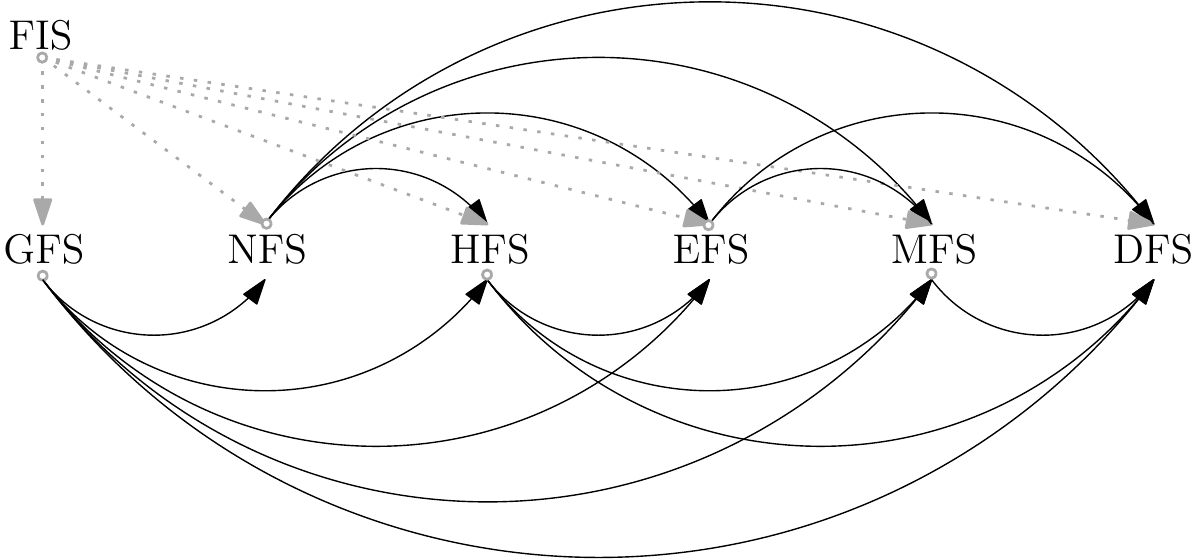}
    \caption{Fuzzy system complexities and concept entailment.}
    \label{fig_complexity}
\end{figure}
This paper reviewed five dimensions of fuzzy systems (FIS): genetic fuzzy systems (GFS), neuro-fuzzy systems (NFS), hierarchical fuzzy systems (HFS), evolving fuzzy systems (EFS), and multiobjective fuzzy systems (MFS). The review linked these dimensions since their are concepts transcend to another dimension. For example, standard FISs, when encoded (formulated) as an optimization problem, GFS offers methods and operators to yield optimal rule structure. FISs also directly be formulated into other dimensions: NFS, HFS, EFS, and MFS. As well as the NFS, HFS, EFS, and MFS when optimized using evolutionary algorithms and metaheuristics are in some sense entails GFS concept. Similarly, NFS forward its concept to HFS, EFS, MFS respectively when hierarchical arrangements of NFS are made, evolving NFS are made, and multiobjective optimization of NFS are done. GFS, NFS, and HFS also offer deep fuzzy systems (DFS) developments directions. When DFS have both evolving and multiobjective viewpoints, it inherits EFS and MFS concepts. Fig.~\ref{fig_complexity} is a summary of the links between the multiple dimensions of FISs and complexity of concept entailment.  The summary in Fig.~\ref{fig_complexity} indicate the challenges and opportunities lie ahead in FISs research: in rules extraction as the number of rules grows with the sophistication of the methods; in constructing network structure for rules, in making hybrid optimizations approach like evolutionary algorithm and particle swarm optimization;  in combining one FIS dimension's concept with another; and in trend towards development of DFS. Moreover, challenges and opportunities in the treatment of FISs for non-stationary data and multiobjective optimization of interpretability-accuracy  trade-off. 

\section*{Acknowledgment}
Authors would like to thank the all the anonymous reviewers for the
technical comments, which enhanced the contents of the preliminary
version of this paper.

%\nocite{*}
\small
\footnotesize 
%\setstretch{1} saves two pages
\bibliographystyle{agsm} % for alphabatical order and author year formate. Amaging!
%\bibliographystyle{IEEEtran}
% argument is your BibTeX string definitions and bibliography database(s)
\bibliography{fuzzy_review_ref_final}

@book{jain2017fuzzy,
    title={Fuzzy and neuro-fuzzy systems in medicine},
    author={Jain, Lakhmi C and Kandel, Abraham and Teodorescu, Horia-Nicolai L},
    year={2017},
    publisher={CRC Press}
}

@article{mouzouris1997nonsingleton,
    title={Nonsingleton fuzzy logic systems: theory and application},
    author={Mouzouris, George C and Mendel, Jerry M},
    journal={IEEE Trans. Fuzzy Syst.},
    volume={5},
    number={1},
    pages={56--71},
    year={1997},
    publisher={IEEE}
}

@article{sahab2011adaptive,
    title={Adaptive non-singleton type-2 fuzzy logic systems: A way forward for handling numerical uncertainties in real world applications},
    author={Sahab, Nazanin and Hagras, Hani},
    journal={Int. J. Comput. Commun. Control},
    volume={6},
    number={3},
    pages={503--529},
    year={2011}
}

@Article{komiyama2017chemistry,
  author    = {Komiyama, Makoto and Yoshimoto, Keitaro and Sisido, Masahiko and Ariga, Katsuhiko},
  title     = {Chemistry can make strict and fuzzy controls for bio-systems: {DNA} nanoarchitectonics and cell-macromolecular nanoarchitectonics},
  journal   = {Bull. Chem. Soc. Jpn.},
  year      = {2017},
  volume    = {90},
  number    = {9},
  pages     = {967--1004},
  publisher = {The Chemical Society of Japan},
}

@article{elhag2015combination,
    title={On the combination of genetic fuzzy systems and pairwise learning for improving detection rates on intrusion detection systems},
    author={Elhag, Salma and Fern{\'a}ndez, Alberto and Bawakid, Abdullah and Alshomrani, Saleh and Herrera, Francisco},
    journal={Expert Syst. Appl.},
    volume={42},
    number={1},
    pages={193--202},
    year={2015},
    publisher={Elsevier}
}

@inproceedings{mamdani1974application,
    author       = {Mamdani, Ebrahim H},
    title        = {Application of fuzzy algorithms for control of simple dynamic plant},
    booktitle    = {Proc. Inst. Electr. Eng.},
    year         = {1974},
    volume       = {121},
    issue       = {12},
    pages        = {1585--1588},
    organization = {IET},
}

@article{takagi1985fuzzy,
    title={Fuzzy identification of systems and its applications to modeling and control},
    author={Takagi, Tomohiro and Sugeno, Michio},
    journal={IEEE Trans. Syst. Man Cybern.},
    volume = {15},
    number={1},
    pages={116--132},
    year={1985},
    publisher={IEEE}
}

@article{mendel2013km,
    title={On {KM} algorithms for solving type-2 fuzzy set problems},
    author={Mendel, Jerry M},
    journal={IEEE Trans. Fuzzy Syst.},
    volume={21},
    number={3},
    pages={426--446},
    year={2013},
    publisher={IEEE}
}

@article{karnik1999type,
    title={Type-2 fuzzy logic systems},
    author={Karnik, Nilesh N and Mendel, Jerry M and Liang, Qilian},
    journal={IEEE Trans. Fuzzy Syst.},
    volume={7},
    number={6},
    pages={643--658},
    year={1999},
    publisher={IEEE}
}

@article{wu2009enhanced,
    title={Enhanced {K}arnik--{M}endel algorithms},
    author={Wu, Dongrui and Mendel, Jerry M},
    journal={IEEE Trans. Fuzzy Syst.},
    volume={17},
    number={4},
    pages={923--934},
    year={2009},
    publisher={IEEE}
}

@article{herrera2008genetic,
    title={Genetic fuzzy systems: taxonomy, current research trends and prospects},
    author={Herrera, Francisco},
    journal={Evol. Intell.},
    volume={1},
    number={1},
    pages={27--46},
    year={2008},
    publisher={Springer}
}

@Article{ojha2017metaheuristic,
  author    = {Ojha, Varun Kumar and Abraham, Ajith and Sn{\'a}{\v{s}}el, V{\'a}clav},
  title     = {Metaheuristic design of feedforward neural networks: A review of two decades of research},
  journal   = {Eng. Appl. Artif. Intell.},
  year      = {2017},
  volume    = {60},
  pages     = {97--116},
  publisher = {Elsevier},
}

@InProceedings{caruana2004data,
  author       = {Caruana, Rich and Niculescu-Mizil, Alexandru},
  title        = {Data mining in metric space: An empirical analysis of supervised learning performance criteria},
  booktitle    = {Proc. 10 ACM SIGKDD, Int. Conf. Knowl. Discovery Data Min.},
  year         = {2004},
  pages        = {69--78},
  organization = {ACM},
}

@article{zadeh1999fuzzy,
    title={Fuzzy sets as a basis for a theory of possibility},
    author={Zadeh, Lotfi A},
    journal={Fuzzy Sets Syst.},
    volume={100},
    number={1},
    pages={9--34},
    year={1999},
    publisher={Elsevier Science}
}

@article{jin2000fuzzy,
    title={Fuzzy modeling of high-dimensional systems: complexity reduction and interpretability improvement},
    author={Jin, Yaochu},
    journal={IEEE Trans. Fuzzy Syst.},
    volume={8},
    number={2},
    pages={212--221},
    year={2000},
    publisher={IEEE}
}

@Article{jang1993anfis,
  author    = {Jang, Jyh-Shing Roger},
  title     = {{ANFIS}: Adaptive-network-based fuzzy inference system},
  journal   = {IEEE Trans. Syst. Man Cybern.},
  year      = {1993},
  volume    = {23},
  number    = {3},
  pages     = {665--685},
  note      = {10471},
  publisher = {IEEE},
}

@article{juang1998online,
    author    = {Juang, Chia-Feng and Lin, Chin-Teng},
    title     = {An online self-constructing neural fuzzy inference network and its applications},
    journal   = {IEEE Trans. Fuzzy Syst.},
    year      = {1998},
    volume    = {6},
    number    = {1},
    pages     = {12--32},
    publisher = {IEEE},
}

@article{kasabov2002denfis,
    author    = {Kasabov, Nikola K and Song, Qun},
    title     = {{DENFIS}: dynamic evolving neural-fuzzy inference system and its application for time-series prediction},
    journal   = {IEEE Trans. Fuzzy Syst.},
    year      = {2002},
    volume    = {10},
    number    = {2},
    pages     = {144--154},
    publisher = {IEEE},
}

@book{cord2001genetic,
    title={Genetic fuzzy systems: evolutionary tuning and learning of fuzzy knowledge bases},
    author={Cord, Oscar and Herrera, Francisco and Hoffmann, Frank and Magdalena, Luis},
    volume={19},
    year={2001},
    publisher={World Scientific}
}

@article{mao2005adaptive,
    title={Adaptive-tree-structure-based fuzzy inference system},
    author={Mao, Jianqin and Zhang, Jiangang and Yue, Yufang and Ding, Haishan},
    journal={IEEE Trans. Fuzzy Syst.},
    volume={13},
    number={1},
    pages={1--12},
    year={2005},
    publisher={IEEE}
}

@article{chien2002learning,
    title={Learning discriminant functions with fuzzy attributes for classification using genetic programming},
    author={Chien, Been-Chian and Lin, Jung Yi and Hong, Tzung-Pei},
    journal={Expert Syst. Appl.},
    volume={23},
    number={1},
    pages={31--37},
    year={2002},
    publisher={Elsevier}
}

@article{funabashi1995fuzzy,
    author  = {Funabashi, Motohisa and Maeda, Akira and Morooka, Yasuo and Mori, Kiyomi},
    title   = {Fuzzy and Neural Hybrid Expert Systems: Synergetic {AI}},
    journal = {IEEE Expert},
    year    = {1995},
    volume  = {10},
    number  = {4},
    pages   = {32--40},
}

@book{back1996evolutionary,
    title={Evolutionary algorithms in theory and practice: evolution strategies, evolutionary programming, genetic algorithms},
    author={Back, Thomas},
    year={1996},
    publisher={Oxford University Press}
}

@book{talbi2009metaheuristics,
    title={Metaheuristics: from design to implementation},
    author={Talbi, El-Ghazali},
    volume={74},
    year={2009},
    publisher={John Wiley \& Sons}
}

@article{cordon2004ten,
    title={Ten years of genetic fuzzy systems: current framework and new trends},
    author={Cord{\'o}n, O and Gomide, F and Herrera, F and Hoffmann, F and Magdalena, L},
    journal={Fuzzy Sets Syst.},
    volume={1},
    number={141},
    pages={5--31},
    year={2004}
}

@Article{sahin2012hybrid,
  author    = {Sahin, Seda and Tolun, Mehmet R and Hassanpour, Reza},
  title     = {Hybrid expert systems: A survey of current approaches and applications},
  journal   = {Expert Syst. Appl.},
  year      = {2012},
  volume    = {39},
  number    = {4},
  pages     = {4609--4617},
  publisher = {Elsevier},
}

@article{herrera1995tuning,
    title={Tuning fuzzy logic controllers by genetic algorithms},
    author={Herrera, Francisco and Lozano, Manuel and Verdegay, Jose L},
    journal={Int. J. Approximate Reasoning},
    volume={12},
    number={3-4},
    pages={299--315},
    year={1995},
    publisher={Elsevier}
}

@Article{ishibuchi1997comparison,
  author    = {Ishibuchi, Hisao and Nakashima, Tomoharu and Murata, Tadahiko},
  title     = {Comparison of the {M}ichigan and {P}ittsburgh approaches to the design of fuzzy classification systems},
  journal   = {Electron. Commun. Jpn. Part III Fundam. Electron. Sci.},
  year      = {1997},
  volume    = {80},
  number    = {12},
  pages     = {10--19},
  publisher = {Wiley Online Library},
}

@inproceedings{ojha2016metaheuristic,
    title={Metaheuristic tuning of type-{II} fuzzy inference systems for data mining},
    author={Ojha, Varun Kumar and Abraham, Ajith and Sn{\'a}{\v{s}}el, V{\'a}clav},
    booktitle={IEEE Int. Conf. Fuzzy Syst.},
    pages={610--617},
    year={2016},
    organization={IEEE}
}

@inproceedings{thrift1991fuzzy,
    author    = {Thrift, Philip R},
    title     = {Fuzzy Logic Synthesis with Genetic Algorithms},
    booktitle = {Proc. 4th Int. Conf. Genetic Algorithms},
    year      = {1991},
    pages     = {509--513},
    place     = {San Diego}
}

@article{kim1995designing,
    title={Designing fuzzy net controllers using genetic algorithms},
    author={Kim, Jinwoo and Moon, Yoonkeon and Zeigler, Bernard P},
    journal={IEEE Control Syst.},
    volume={15},
    number={3},
    pages={66--72},
    year={1995},
    publisher={IEEE}
}

@article{hadavandi2010integration,
    title={Integration of genetic fuzzy systems and artificial neural networks for stock price forecasting},
    author={Hadavandi, Esmaeil and Shavandi, Hassan and Ghanbari, Arash},
    journal={Knowledge-Based Syst.},
    volume={23},
    number={8},
    pages={800--808},
    year={2010},
    publisher={Elsevier},
    note={152}
}

@phdthesis{michigan1982,
    author={Booker, Lashon Bernard},
    title={Intelligent Behavior As an Adaptation to the Task Environment},
    school={University of Michigan},
    year= {1982},
    address={Ann Arbor, MI, USA},
}

@phdthesis{pittsburgh1980,
    author =    {Smith, Stephen Frederick},
    title =     {A Learning System Based on Genetic Adaptive Algorithms},
    school = {University of Pittsburgh},
    year =      {1980},
    address =   {Pittsburgh, PA, USA},
}

@inproceedings{lee1993integrating,
    title={Integrating design stage of fuzzy systems using genetic algorithms},
    author={Lee, Michael A and Takagi, Hideyuki},
    booktitle={2nd IEEE Int. Conf.  Fuzzy Syst.},
    pages={612--617},
    year={1993},
    organization={IEEE}
}

@Article{papadakis2002ga,
  author    = {Papadakis, Stelios E and Theocharis, JB},
  title     = {A {GA}-based fuzzy modeling approach for generating {TSK} models},
  journal   = {Fuzzy Sets Syst.},
  year      = {2002},
  volume    = {131},
  number    = {2},
  pages     = {121--152},
  publisher = {Elsevier},
}

@article{wu2006genetic,
    title={Genetic learning and performance evaluation of interval type-2 fuzzy logic controllers},
    author={Wu, Dongrui and Tan, Woei Wan},
    journal={Eng. Appl. Artif. Intell.},
    volume={19},
    number={8},
    pages={829--841},
    year={2006},
    publisher={Elsevier}
}

@article{ishibuchi1995selecting,
    title={Selecting fuzzy if-then rules for classification problems using genetic algorithms},
    author={Ishibuchi, Hisao and Nozaki, Ken and Yamamoto, Naohisa and Tanaka, Hideo},
    journal={IEEE Trans. Fuzzy Syst.},
    volume={3},
    number={3},
    pages={260--270},
    year={1995},
    publisher={IEEE}
}

@article{hoffmann1997evolutionary,
    title={Evolutionary design of a fuzzy knowledge base for a mobile robot},
    author={Hoffmann, Frank and Pfister, Gerd},
    journal={Int. J. Approximate Reasoning},
    volume={17},
    number={4},
    pages={447--469},
    year={1997},
    publisher={Elsevier}
}

@article{melin2012genetic,
    title={Genetic optimization of modular neural networks with fuzzy response integration for human recognition},
    author={Melin, Patricia and S{\'a}nchez, Daniela and Castillo, Oscar},
    journal={Inf. Sci.},
    volume={197},
    pages={1--19},
    year={2012},
    publisher={Elsevier}
}

@article{carse1996evolving,
    title={Evolving fuzzy rule based controllers using genetic algorithms},
    author={Carse, Brian and Fogarty, Terence C and Munro, Alistair},
    journal={Fuzzy Sets Syst.},
    volume={80},
    number={3},
    pages={273--293},
    year={1996},
    publisher={Elsevier}
}

@article{tsang2007genetic,
    title={Genetic-fuzzy rule mining approach and evaluation of feature selection techniques for anomaly intrusion detection},
    author={Tsang, Chi-Ho and Kwong, Sam and Wang, Hanli},
    journal={Pattern Recognit.},
    volume={40},
    number={9},
    pages={2373--2391},
    year={2007},
    publisher={Elsevier}
}

@article{goldberg1988genetic,
    title={Genetic algorithms and machine learning},
    author={Goldberg, David E and Holland, John H},
    journal={Mach. Learn.},
    volume={3},
    number={2},
    pages={95--99},
    year={1988},
    publisher={Springer}
}

@inproceedings{kennedy1997discrete,
    title={A discrete binary version of the particle swarm algorithm},
    author={Kennedy, James and Eberhart, Russell C},
    booktitle={IEEE Int. Conf. Syst. Man Cybern},
    volume={5},
    pages={4104--4108},
    year={1997},
    organization={IEEE}
}

@article{dorigo1999ant,
    title={Ant algorithms for discrete optimization},
    author={Dorigo, Marco and Caro, Gianni Di and Gambardella, Luca M},
    journal={Artif. Life},
    volume={5},
    number={2},
    pages={137--172},
    year={1999},
    publisher={MIT Press}
}

@incollection{wright1991genetic,
    title={Genetic algorithms for real parameter optimization},
    author={Wright, Alden H},
    booktitle={Foundations of Genetic Algorithms},
    volume={1},
    pages={205--218},
    year={1991},
    publisher={Elsevier}
}

@incollection{kennedy2011particle,
    title={Particle swarm optimization},
    author={Kennedy, James},
    booktitle={Encyclopedia of Machine Learning},
    pages={760--766},
    year={2011},
    publisher={Springer}
}

@article{socha2008ant,
    title={Ant colony optimization for continuous domains},
    author={Socha, Krzysztof and Dorigo, Marco},
    journal={Eur. J. Oper. Res.},
    volume={185},
    number={3},
    pages={1155--1173},
    year={2008},
    publisher={Elsevier}
}

@book{yang2010nature,
    title={Nature-inspired metaheuristic algorithms},
    author={Yang, Xin-She},
    year={2010},
    publisher={Luniver Press}
}

@article{goldberg1991real,
    title={Real-coded genetic algorithms, virtual alphabets, and blocking},
    author={Goldberg, David E},
    journal={Complex Syst.},
    volume={5},
    number={2},
    pages={139--167},
    year={1991}
}

@incollection{eshelman1993real,
    title={Real-coded genetic algorithms and interval-schemata},
    author={Eshelman, Larry J and Schaffer, J David},
    booktitle={Foundations of genetic algorithms},
    volume={2},
    pages={187--202},
    year={1993},
    publisher={Elsevier}
}

@inproceedings{ishibuchi1999hybrid,
    author    = {Ishibuchi, Hisao and Nakashima, Tomoharu and Kuroda, Tetsuya},
    title     = {A hybrid fuzzy genetics-based machine learning algorithm: hybridization of {M}ichigan approach and {P}ittsburgh approach},
    booktitle = {Int. Conf. Syst. Man Cybern},
    year      = {1999},
    volume    = {1},
    pages     = {296--301},
}

@article{cordon1997three,
    title={A three-stage evolutionary process for learning descriptive and approximate fuzzy-logic-controller knowledge bases from examples},
    author={Cord{\'o}n, Oscar and Herrera, Francisco},
    journal={Int. J. Approximate Reasoning},
    volume={17},
    number={4},
    pages={369--407},
    year={1997},
    publisher={Elsevier}
}

@inproceedings{martinez2010fuzzy,
    title={Fuzzy logic controllers optimization using genetic algorithms and particle swarm optimization},
    author={Martinez-Soto, Ricardo and Castillo, Oscar and Aguilar, Luis T and Melin, Patricia},
    booktitle={Mexican Int. Conf. Artif. Intell.},
    pages={475--486},
    year={2010},
    organization={Springer}
}

@inproceedings{shahzad2009hybrid,
    title={A hybrid {GA-PSO} fuzzy system for user identification on smart phones},
    author={Shahzad, Muhammad and Zahid, Saira and Farooq, Muddassar},
    booktitle={Proc. 11th Annu. Conf. Genetic and Evol. Comput.},
    pages={1617--1624},
    year={2009},
    organization={ACM}
}

@Article{martinez2015hybrid,
  author    = {Mart{\'\i}nez-Soto, Ricardo and Castillo, Oscar and Aguilar, Luis T and Rodriguez, Antonio},
  title     = {A hybrid optimization method with {PSO} and {GA} to automatically design Type-1 and Type-2 fuzzy logic controllers},
  journal   = {Int. J. Mach. Learn. Cybern.},
  year      = {2015},
  volume    = {6},
  number    = {2},
  pages     = {175--196},
  publisher = {Springer},
}

@article{das2015evolving,
    title={An evolving interval type-2 neurofuzzy inference system and its metacognitive sequential learning algorithm},
    author={Das, Ankit Kumar and Subramanian, Kartick and Sundaram, Suresh},
    journal={IEEE Trans. Fuzzy Syst.},
    volume={23},
    number={6},
    pages={2080--2093},
    year={2015},
    publisher={IEEE}
}

@article{valdez2011improved,
    title={An improved evolutionary method with fuzzy logic for combining particle swarm optimization and genetic algorithms},
    author={Valdez, Fevrier and Melin, Patricia and Castillo, Oscar},
    journal={Appl. Soft Comput.},
    volume={11},
    number={2},
    pages={2625--2632},
    year={2011},
    publisher={Elsevier}
}

@article{castillo2012comparative,
    title={Comparative study of bio-inspired algorithms applied to the optimization of type-1 and type-2 fuzzy controllers for an autonomous mobile robot},
    author={Castillo, Oscar and Mart{\'\i}nez-Marroqu{\'\i}n, Ricardo and Melin, Patricia and Valdez, Fevrier and Soria, Jos{\'e}},
    journal={Inf. Sci.},
    volume={192},
    pages={19--38},
    year={2012},
    publisher={Elsevier}
}

@Article{melin2013optimal,
  author    = {Melin, Patricia and Olivas, Frumen and Castillo, Oscar and Valdez, Fevrier and Soria, Jose and Valdez, Mario},
  title     = {Optimal design of fuzzy classification systems using {PSO} with dynamic parameter adaptation through fuzzy logic},
  journal   = {Expert Syst. Appl.},
  year      = {2013},
  volume    = {40},
  number    = {8},
  pages     = {3196--3206},
  publisher = {Elsevier},
}

@article{pandiarajan2016fuzzy,
    title={Fuzzy harmony search algorithm based optimal power flow for power system security enhancement},
    author={Pandiarajan, K and Babulal, CK},
    journal={Int. J. Electr. Power Energy Syst.},
    volume={78},
    pages={72--79},
    year={2016},
    publisher={Elsevier}
}

@article{habbi2015self,
    title={Self-generated fuzzy systems design using artificial bee colony optimization},
    author={Habbi, Hacene and Boudouaoui, Yassine and Karaboga, Dervis and Ozturk, Celal},
    journal={Inf. Sci.},
    volume={295},
    pages={145--159},
    year={2015},
    publisher={Elsevier}
}

@article{verma2017optimal,
    title={An optimal fuzzy system for edge detection in color images using bacterial foraging algorithm},
    author={Verma, Om Prakash and Parihar, Anil Singh},
    journal={IEEE Trans. Fuzzy Syst.},
    volume={25},
    number={1},
    pages={114--127},
    year={2017},
    publisher={IEEE}
}

@article{juang2000genetic,
    title={Genetic reinforcement learning through symbiotic evolution for fuzzy controller design},
    author={Juang, Chia-Feng and Lin, Jiann-Yow and Lin, Chin-Teng},
    journal={IEEE Trans. Syst. Man Cybern. Part B Cybern.},
    volume={30},
    number={2},
    pages={290--302},
    year={2000},
    publisher={IEEE}
}

@Article{juang2009reinforcement,
  author    = {Juang, Chia-Feng and Hsu, Chia-Hung},
  title     = {Reinforcement interval type-2 fuzzy controller design by online rule generation and {Q}-value-aided ant colony optimization},
  journal   = {IEEE Trans. Syst. Man Cybern. Part B Cybern.},
  year      = {2009},
  volume    = {39},
  number    = {6},
  pages     = {1528--1542},
  publisher = {IEEE},
}

@InProceedings{venturini1993sia,
  author       = {Venturini, Gilles},
  title        = {{SIA}: A supervised inductive algorithm with genetic search for learning attributes based concepts},
  booktitle    = {Eur. Conf. Machine Learning},
  year         = {1993},
  pages        = {280--296},
  organization = {Springer},
}

@incollection{gonzalez1997multi,
    title={Multi-stage genetic fuzzy systems based on the iterative rule learning approach},
    author={Gonz{\'a}lez, Mu{\~n}oz, Antonio and Herrera, Francisco},
    booktitle={Mathware \& Soft Comput.},
    volume= {4},
    issue={3},
    year={1997},
    publisher={Polytechnic University of Catalonia}
}

@article{ahn2007iterative,
    title={Iterative learning control: Brief survey and categorization},
    author={Ahn, Hyo-Sung and Chen, YangQuan and Moore, Kevin L},
    journal={IEEE Trans. Syst. Man Cybern. Part C Appl. Rev.},
    volume={37},
    number={6},
    pages={1099--1121},
    year={2007},
    publisher={IEEE}
}

@article{greene1993competition,
    title={Competition-based induction of decision models from examples},
    author={Greene, David Perry and Smith, Stephen F},
    journal={Mach. Learn.},
    volume={13},
    number={2-3},
    pages={229--257},
    year={1993},
    publisher={Springer}
}

@article{whitehead1996cooperative,
    title={Cooperative-competitive genetic evolution of radial basis function centers and widths for time series prediction},
    author={Whitehead, Bruce A and Choate, Timothy D},
    journal={IEEE Trans. Neural Netw.},
    volume={7},
    number={4},
    pages={869--880},
    year={1996},
    publisher={IEEE}
}

@Article{delgado2004coevolutionary,
  author    = {Delgado, Myriam Regattieri and Von Zuben, Fernando and Gomide, Fernando},
  title     = {Coevolutionary genetic fuzzy systems: A hierarchical collaborative approach},
  journal   = {Fuzzy Sets Syst.},
  year      = {2004},
  volume    = {141},
  number    = {1},
  pages     = {89--106},
  publisher = {Elsevier},
}

@Book{koza1994genetic,
  title     = {Genetic programming {II}: Automatic discovery of reusable programs},
  publisher = {MIT Press Cambridge},
  year      = {1994},
  author    = {Koza, John R and Rice, James P},
  volume    = {40},
}

@Article{sanchez2001combining,
  author    = {S{\'a}nchez, Luciano and Couso, In{\'e}s and Corrales, Jose A.},
  title     = {Combining {GP} operators with {SA} search to evolve fuzzy rule based classifiers},
  journal   = {Inf. Sci.},
  year      = {2001},
  volume    = {136},
  number    = {1-4},
  pages     = {175--191},
  publisher = {Elsevier},
}

@article{cordon2002new,
    title={A new evolutionary algorithm combining simulated annealing and genetic programming for relevance feedback in fuzzy information retrieval systems},
    author={Cord{\'o}n, Oscar and Moya, F and Zarco, Carmen},
    journal={Soft Comput.},
    volume={6},
    number={5},
    pages={308--319},
    year={2002},
    publisher={Springer}
}

@book{aarts1988simulated,
    title={Simulated annealing and Boltzmann machines},
    author={Aarts, Emile and Korst, Jan},
    year={1988},
    publisher={John Wiley \& Sons}
}

@inproceedings{jang1991fuzzy,
    title={Fuzzy Modeling Using Generalized Neural Networks and Kalman Filter Algorithm},
    author={Jang, Jyh-Shing Roger},
    booktitle={AAAI},
    volume={91},
    pages={762--767},
    year={1991}
}

@Article{buckley1994fuzzy,
  author    = {Buckley, James J and Hayashi, Yoichi},
  title     = {Fuzzy Neural Networks: A survey},
  journal   = {Fuzzy Sets Syst.},
  year      = {1994},
  volume    = {66},
  number    = {1},
  pages     = {1--13},
  publisher = {Elsevier},
}

@Article{karaboga2018adaptive,
  author    = {Karaboga, Dervis and Kaya, Ebubekir},
  title     = {Adaptive network based fuzzy inference system ({ANFIS}) training approaches: A comprehensive survey},
  journal   = {Artif. Intell. Rev.},
  year      = {2018},
  pages     = {1--31},
  publisher = {Springer},
}

@article{andrews1995survey,
    title={Survey and critique of techniques for extracting rules from trained artificial neural networks},
    author={Andrews, Robert and Diederich, Joachim and Tickle, Alan B},
    journal={Knowledge-Based Syst.},
    volume={8},
    number={6},
    pages={373--389},
    year={1995},
    publisher={Elsevier}
}

@article{feuring1999stability,
    author    = {Feuring, Thomas and Buckley, James J and Lippe, Wolfram-M and Tenhagen, Andreas},
    title     = {Stability analysis of neural net controllers using fuzzy neural networks},
    journal   = {Fuzzy Sets Syst.},
    year      = {1999},
    volume    = {101},
    number    = {2},
    pages     = {303--313},
    publisher = {Elsevier},
}

@article{ishibuchi2001numerical,
    author    = {Ishibuchi, Hisao and Nii, Manabu},
    title     = {Numerical analysis of the learning of fuzzified neural networks from fuzzy if--then rules},
    journal   = {Fuzzy Sets Syst.},
    year      = {2001},
    volume    = {120},
    number    = {2},
    pages     = {281--307},
    publisher = {Elsevier},
}

@article{buckley1999equivalence,
    title={On the equivalence of neural nets and fuzzy expert systems},
    author={Buckley, James J and Hayashi, Yoichi and Czoga{\l}a, Ernest},
    journal={Fuzzy Sets Syst.},
    volume={100},
    pages={145--150},
    year={1999},
    publisher={Elsevier}
}

@article{li2000equivalence,
    author    = {Li, Hong-Xing and Chen, CL Philip},
    title     = {The equivalence between fuzzy logic systems and feedforward neural networks},
    journal   = {IEEE Trans. Neural Netw.},
    year      = {2000},
    volume    = {11},
    number    = {2},
    pages     = {356--365},
    publisher = {IEEE},
}

@article{lin1994reinforcement,
    author    = {Lin, Chin-Teng and Lee, CS George},
    title     = {Reinforcement structure/parameter learning for neural-network-based fuzzy logic control systems},
    journal   = {IEEE Trans. Fuzzy Syst.},
    year      = {1994},
    volume    = {2},
    number    = {1},
    pages     = {46--63},
    publisher = {IEEE},
}

@article{moriarty1996efficient,
    title={Efficient reinforcement learning through symbiotic evolution},
    author={Moriarty, David E and Mikkulainen, Risto},
    journal={Mach. Learn.},
    volume={22},
    number={1-3},
    pages={11--32},
    year={1996},
    publisher={Springer}
}

@article{berenji1992learning,
    author    = {Berenji, Hamid R and Khedkar, Pratap},
    title     = {Learning and tuning fuzzy logic controllers through reinforcements},
    journal   = {IEEE Trans. Neural Netw.},
    year      = {1992},
    volume    = {3},
    number    = {5},
    pages     = {724--740},
    publisher = {IEEE},
}

@inproceedings{nauck1993fuzzy,
    author       = {Nauck, Detlef and Kruse, Rudolf},
    title        = {A fuzzy neural network learning fuzzy control rules and membership functions by fuzzy error backpropagation},
    booktitle    = {IEEE Int. Conf. Neural Netw.},
    year         = {1993},
    pages        = {1022--1027},
    organization = {IEEE},
}

@article{wang2002self,
    author    = {Wang, Jeen-Shing and Lee, CS George},
    title     = {Self-adaptive neuro-fuzzy inference systems for classification applications},
    journal   = {IEEE Trans. Fuzzy Syst.},
    year      = {2002},
    volume    = {10},
    number    = {6},
    pages     = {790--802},
    publisher = {IEEE},
}

@article{wang1999self,
    title={A self-organizing neural-network-based fuzzy system},
    author={Wang, Yin and Rong, Gang},
    journal={Fuzzy Sets Syst.},
    volume={103},
    number={1},
    pages={1--11},
    year={1999},
    publisher={Elsevier}
}

@article{lin2001self,
    author    = {Lin, Faa-Jeng and Lin, Chih-Hong and Shen, Po-Hung},
    title     = {Self-constructing fuzzy neural network speed controller for permanent-magnet synchronous motor drive},
    journal   = {IEEE Trans. Fuzzy Syst.},
    year      = {2001},
    volume    = {9},
    number    = {5},
    pages     = {751--759},
    publisher = {IEEE},
}

@article{lin1995neural,
    author    = {Lin, Chin-Teng},
    title     = {A neural fuzzy control system with structure and parameter learning},
    journal   = {Fuzzy Sets Syst.},
    year      = {1995},
    volume    = {70},
    number    = {2},
    pages     = {183--212},
    note      = {235},
    publisher = {Elsevier},
}

@article{wu2000dynamic,
    author    = {Wu, Shiqian and Er, Meng Joo},
    title     = {Dynamic fuzzy neural networks-a novel approach to function approximation},
    journal   = {IEEE Trans. Syst. Man Cybern. Part B Cybern.},
    year      = {2000},
    volume    = {30},
    number    = {2},
    pages     = {358--364},
    publisher = {IEEE},
}

@article{kasabov2001line,
    title={On-line learning, reasoning, rule extraction and aggregation in locally optimized evolving fuzzy neural networks},
    author={Kasabov, Nikola K},
    journal={Neurocomputing},
    volume={41},
    number={1},
    pages={25--45},
    year={2001},
    publisher={Elsevier}
}

@article{jang1992self,
    author    = {Jang, Jyh-Shing R},
    title     = {Self-learning fuzzy controllers based on temporal backpropagation},
    journal   = {IEEE Trans. Neural Netw.},
    year      = {1992},
    volume    = {3},
    number    = {5},
    pages     = {714--723},
    publisher = {IEEE},
}

@article{juang1999recurrent,
    author    = {Juang, Chia-Feng and Lin, Chin-Teng},
    title     = {A recurrent self-organizing neural fuzzy inference network},
    journal   = {IEEE Trans. Neural Netw.},
    year      = {1999},
    volume    = {10},
    number    = {4},
    pages     = {828--845},
    publisher = {IEEE},
}

@article{mastorocostas2002recurrent,
    author    = {Mastorocostas, Paris A and Theocharis, John B},
    title     = {A recurrent fuzzy-neural model for dynamic system identification},
    journal   = {IEEE Trans. Syst. Man Cybern. Part B Cybern.},
    year      = {2002},
    volume    = {32},
    number    = {2},
    pages     = {176--190},
    publisher = {IEEE},
}

@InProceedings{brown1995high,
    author    = {Brown, M and Bossley, KM and Mills, DJ and Harris, CJ},
    title     = {High dimensional neurofuzzy systems: overcoming the curse of dimensionality},
    booktitle = {Proc. 1995 IEEE Int. Fuzzy Syst.},
    year      = {1995},
    volume    = {4},
    pages     = {2139--2146},
}

@article{kasabov2001evolving,
    title={Evolving fuzzy Neural Networks for supervised/unsupervised online knowledge-based learning},
    author={Kasabov, Nikola},
    journal={IEEE Trans. Syst. Man Cybern. Part B Cybern.},
    volume={31},
    number={6},
    pages={902--918},
    year={2001},
    publisher={IEEE}
}

@article{deng2003line,
    title={On-line pattern analysis by evolving self-organizing maps},
    author={Deng, Da and Kasabov, Nikola},
    journal={Neurocomputing},
    volume={51},
    pages={87--103},
    year={2003},
    publisher={Elsevier}
}

@article{deb2002fast,
    title={A fast and elitist multiobjective genetic algorithm: {NSGA-II}},
    author={Deb, Kalyanmoy and Pratap, Amrit and Agarwal, Sameer and Meyarivan, TAMT},
    journal={IEEE Trans. Evol. Comput.},
    volume={6},
    number={2},
    pages={182--197},
    year={2002},
    publisher={IEEE}
}

@book{coello2007evolutionary,
    title={Evolutionary algorithms for solving multi-objective problems},
    author={Coello, Carlos A Coello and Lamont, Gary B and Van Veldhuizen, David A},
    volume={5},
    year={2007},
    publisher={Springer}
}

@Article{zitzler1999multiobjective,
  author    = {Zitzler, Eckart and Thiele, Lothar},
  title     = {Multiobjective evolutionary algorithms: A comparative case study and the strength {P}areto approach},
  journal   = {IEEE Trans. Evol. Comput.},
  year      = {1999},
  volume    = {3},
  number    = {4},
  pages     = {257--271},
  publisher = {IEEE},
}

@Article{tung2011safin,
  author    = {Tung, Sau Wai and Quek, Chai and Guan, Cuntai},
  title     = {{SaFIN}: A self-adaptive fuzzy inference network},
  journal   = {IEEE Trans. Neural Netw.},
  year      = {2011},
  volume    = {22},
  number    = {12},
  pages     = {1928--1940},
  publisher = {IEEE},
}

@article{wu2001fast,
    title={A fast approach for automatic generation of fuzzy rules by generalized dynamic fuzzy neural networks},
    author={Wu, Shiqian and Er, Meng Joo and Gao, Yang},
    journal={IEEE Trans. Fuzzy Syst.},
    volume={9},
    number={4},
    pages={578--594},
    year={2001},
    publisher={IEEE}
}

@article{lin2013mutually,
    title={A mutually recurrent interval type-2 neural fuzzy system ({MRIT2NFS}) with self-evolving structure and parameters},
    author={Lin, Yang-Yin and Chang, Jyh-Yeong and Pal, Nikhil R and Lin, Chin-Teng},
    journal={IEEE Trans. Fuzzy Syst.},
    volume={21},
    number={3},
    pages={492--509},
    year={2013},
    publisher={IEEE}
}

@article{juang2008self,
    title={A self-evolving interval type-2 fuzzy neural network with online structure and parameter learning},
    author={Juang, Chia-Feng and Tsao, Yu-Wei},
    journal={IEEE Trans. Fuzzy Syst.},
    volume={16},
    number={6},
    pages={1411--1424},
    year={2008},
    publisher={IEEE}
}

@article{horikawa1992fuzzy,
    author  = {Horikawa, Shin-ichi and Furuhashi, Takeshi and Uchikawa, Yoshiki},
    title   = {On fuzzy modeling using fuzzy neural networks with the back-propagation algorithm},
    journal = {IEEE Trans. Neural Netw.},
    year    = {1992},
    volume  = {3},
    number  = {5},
    pages   = {801--806},
}

@article{park2002fuzzy,
    author    = {Park, Byoung-Jun and Pedrycz, Witold and Oh, Sung-Kwun},
    title     = {Fuzzy polynomial neural networks: hybrid architectures of fuzzy modeling},
    journal   = {IEEE Trans. Fuzzy Syst.},
    year      = {2002},
    volume    = {10},
    number    = {5},
    pages     = {607--621},
    publisher = {IEEE},
}

@inproceedings{masuoka1990neurofuzzy,
    title={Neurofuzzy system-fuzzy inference using a structured neural network},
    author={Masuoka, Ryusuke and Watanabe, Nobuo and Kawamura, Akira and Owada, Yuri and Asakawa, Kazuo},
    booktitle={Proc. Int. Conf.  Fuzzy Logic \& Neural Netw.},
    pages={173--177},
    year={1990}
}

@article{buckley1995neural,
    title={Neural nets for fuzzy systems},
    author={Buckley, James J and Yoichi, Hayashi},
    journal={Fuzzy Sets Syst.},
    volume={71},
    number={3},
    pages={265--276},
    year={1995},
    publisher={Elsevier}
}

@article{nauck1997neuro,
    author    = {Nauck, Detlef and Kruse, Rudolf},
    title     = {A neuro-fuzzy method to learn fuzzy classification rules from data},
    journal   = {Fuzzy Sets Syst.},
    year      = {1997},
    volume    = {89},
    number    = {3},
    pages     = {277--288},
    publisher = {Elsevier},
}

@article{nauck1999neuro,
    author    = {Nauck, Detlef and Kruse, Rudolf},
    title     = {Neuro-fuzzy systems for function approximation},
    journal   = {Fuzzy Sets Syst.},
    year      = {1999},
    volume    = {101},
    number    = {2},
    pages     = {261--271},
    note      = {262},
    publisher = {Elsevier},
}

@article{kasabov1997funn,
    title={{FuNN}-2-a fuzzy neural network architecture for adaptive learning and knowledge acquisition},
    author={Kasabov, Nikola K and Kim, Jaesoo and Watts, Michael J and Gray, Andrew R},
    journal={Inf. Sci.},
    volume={101},
    number={3},
    pages={155--175},
    year={1997},
    publisher={Elsevier}
}

@Article{kim1999hyfis,
  author    = {Kim, Jaesoo and Kasabov, N},
  title     = {{HyFIS}: Adaptive neuro-fuzzy inference systems and their application to nonlinear dynamical systems},
  journal   = {Neural Netw.},
  year      = {1999},
  volume    = {12},
  number    = {9},
  pages     = {1301--1319},
  publisher = {Elsevier},
}

@article{leng2006design,
    author    = {Leng, Gang and McGinnity, Thomas Martin and Prasad, Girijesh},
    title     = {Design for self-organizing fuzzy neural networks based on genetic algorithms},
    journal   = {IEEE Trans. Fuzzy Syst.},
    year      = {2006},
    volume    = {14},
    number    = {6},
    pages     = {755--766},
    publisher = {IEEE},
}

@article{shann1995fuzzy,
    title={A fuzzy neural network for rule acquiring on fuzzy control systems},
    author={Shann, JJ and Fu, HC},
    journal={Fuzzy Sets Syst.},
    volume={71},
    number={3},
    pages={345--357},
    year={1995},
    publisher={Elsevier}
}

@article{looney1988fuzzy,
    author    = {Looney, Carl G},
    title     = {Fuzzy Petri nets for rule-based decisionmaking},
    journal   = {IEEE Trans. Syst. Man Cybern.},
    year      = {1988},
    volume    = {18},
    number    = {1},
    pages     = {178--183},
    publisher = {IEEE},
}

@article{carpenter1991fuzzy,
    title={Fuzzy {ART}: Fast stable learning and categorization of analog patterns by an adaptive resonance system},
    author={Carpenter, Gail A and Grossberg, Stephen and Rosen, David B},
    journal={Neural Netw.},
    volume={4},
    number={6},
    pages={759--771},
    year={1991},
    publisher={Elsevier}
}

@article{simpson1992fuzzy,
    author    = {Simpson, Patrick K},
    title     = {Fuzzy min-max neural networks--Part 1: Classification},
    journal   = {IEEE Trans. Neural Netw.},
    year      = {1992},
    volume    = {3},
    number    = {5},
    pages     = {776--786},
    publisher = {IEEE},
}

@article{cho1996radial,
    author    = {Cho, Kwang Bo and Wang, Bo Hyeun},
    title     = {Radial basis function based adaptive fuzzy systems and their applications to system identification and prediction},
    journal   = {Fuzzy Sets Syst.},
    year      = {1996},
    volume    = {83},
    number    = {3},
    pages     = {325--339},
    publisher = {Elsevier},
}

@article{lee2000identification,
    author    = {Lee, Ching-Hung and Teng, Ching-Cheng},
    title     = {Identification and control of dynamic systems using recurrent fuzzy neural networks},
    journal   = {IEEE Trans. Fuzzy Syst.},
    year      = {2000},
    volume    = {8},
    number    = {4},
    pages     = {349--366},
    publisher = {IEEE},
}

@article{torra2002review,
    author =    {Torra, Vicen{\c{c}}},
    title =     {A review of the construction of hierarchical fuzzy systems},
    journal = {Int. J. Intell. Syst.},
    year =      {2002},
    volume =    {17},
    number =    {5},
    pages =     {531--543},
    publisher = {Wiley Online Library}
}

@Article{hoffmann2001genetic,
  author    = {Hoffmann, Frank and Nelles, Oliver},
  title     = {Genetic programming for model selection of {TSK}-fuzzy systems},
  journal   = {Inf. Sci.},
  year      = {2001},
  volume    = {136},
  number    = {1-4},
  pages     = {7--28},
  publisher = {Elsevier},
}

@article{raju1991hierarchical,
    title={Hierarchical fuzzy control},
    author={Raju, GVS and Zhou, Jun and Kisner, Roger A},
    journal={Int. J. Control},
    volume={54},
    number={5},
    pages={1201--1216},
    year={1991},
    publisher={Taylor \& Francis}
}

@article{wang1999analysis,
    title={Analysis and design of hierarchical fuzzy systems},
    author={Wang, Li-Xin},
    journal={IEEE Trans. Fuzzy Syst.},
    volume={7},
    number={5},
    pages={617--624},
    year={1999},
    publisher={IEEE}
}

@article{zeng2005approximation,
    title={Approximation capabilities of hierarchical fuzzy systems},
    author={Zeng, X-J and Keane, John A},
    journal={IEEE Trans. Fuzzy Syst.},
    volume={13},
    number={5},
    pages={659--672},
    year={2005},
    publisher={IEEE}
}

@article{wang1998universal,
    title={Universal approximation by hierarchical fuzzy systems},
    author={Wang, Li-Xin},
    journal={Fuzzy Sets Syst.},
    volume={93},
    number={2},
    pages={223--230},
    year={1998},
    publisher={Elsevier}
}

@article{chung2000multistage,
    title={On multistage fuzzy neural network modeling},
    author={Chung, Fu-Lai and Duan, Ji-Cheng},
    journal={IEEE Trans. Fuzzy Syst.},
    volume={8},
    number={2},
    pages={125--142},
    year={2000},
    publisher={IEEE}
}

@article{ojha2018multiobjective,
    title={Multiobjective programming for type-2 hierarchical fuzzy inference trees},
    author={Ojha, Varun Kumar and Sn{\'a}{\v{s}}el, V{\'a}clav and Abraham, Ajith},
    journal={IEEE Trans. Fuzzy Syst.},
    volume={26},
    number={2},
    pages={915--936},
    year={2018},
    publisher={IEEE}
}

@inproceedings{karr2000synergistic,
    title={A synergistic architecture for adaptive, intelligent control system development},
    author={Karr, Charles L},
    booktitle={26th Annu. Conf. IEEE Ind. Electron. Soc. (IECON)},
    volume={4},
    pages={2986--2991},
    year={2000},
    organization={IEEE}
}

@Article{maeda1996investigation,
  author    = {Maeda, Hiroshi},
  title     = {An investigation on the spread of fuzziness in multi-fold multi-stage approximate reasoning by pictorial representation---under sup-min composition and triangular type membership function},
  journal   = {Fuzzy Sets Syst.},
  year      = {1996},
  volume    = {80},
  number    = {2},
  pages     = {133--148},
  publisher = {Elsevier},
}

@article{wang2006survey,
    title={A survey of hierarchical fuzzy systems},
    author={Wang, Di and Zeng, Xiao-jun and Keane, J},
    journal={Int. J. Comput. Cognition},
    volume={4},
    number={1},
    pages={18--29},
    year={2006},
    publisher={Citeseer}
}

@inproceedings{magdalena2018hierarchical,
    title={Do Hierarchical Fuzzy Systems Really Improve Interpretability?},
    author={Magdalena, Luis},
    booktitle={Int. Conf. Inf. Process. Manage. Uncertainty in Knowledge-Based Syst.},
    pages={16--26},
    year={2018},
    organization={Springer}
}

@article{kouikoglou2009monotonicity,
    title={On the monotonicity of hierarchical sum--product fuzzy systems},
    author={Kouikoglou, Vassilis S and Phillis, Yannis A},
    journal={Fuzzy Sets Syst.},
    volume={160},
    number={24},
    pages={3530--3538},
    year={2009},
    publisher={Elsevier}
}

@article{won2002parameter,
    title={Parameter conditions for monotonic {T}akagi--{S}ugeno--{K}ang fuzzy system},
    author={Won, Jin M and Park, Sang Y and Lee, Jin S},
    journal={Fuzzy Sets Syst.},
    volume={132},
    number={2},
    pages={135--146},
    year={2002},
    publisher={Elsevier}
}

@article{lee2003modeling,
    title={Modeling of hierarchical fuzzy systems},
    author={Lee, Ming-Ling and Chung, Hung-Yuan and Yu, Fang-Ming},
    journal={Fuzzy Sets Syst.},
    volume={138},
    number={2},
    pages={343--361},
    year={2003},
    publisher={Elsevier}
}

@article{domingo1997knowledge,
    title={A knowledge level analysis of taxonomic domains},
    author={Domingo, Marta and Sierra, Carles},
    journal={Int. J. Intell. Syst.},
    volume={12},
    number={2},
    pages={105--135},
    year={1997},
    publisher={Wiley Online Library}
}

@inproceedings{joo2003method,
    title={A method of converting conventional fuzzy logic system to 2 layered hierarchical fuzzy system},
    author={Joo, Moo G},
    booktitle={IEEE Int. Conf. Fuzzy Syst.},
    volume={2},
    pages={1357--1362},
    year={2003},
    organization={IEEE}
}

@article{hagras2004hierarchical,
    title={A hierarchical type-2 fuzzy logic control architecture for autonomous mobile robots},
    author={Hagras, Hani A},
    journal={IEEE Trans. Fuzzy Syst.},
    volume={12},
    number={4},
    pages={524--539},
    year={2004},
    publisher={IEEE}
}

@article{fernandez2009hierarchical,
    title={Hierarchical fuzzy rule based classification systems with genetic rule selection for imbalanced data-sets},
    author={Fern{\'a}ndez, Alberto and del Jesus, Mar{\'\i}a Jos{\'e} and Herrera, Francisco},
    journal={Int. J. Approximate Reasoning},
    volume={50},
    number={3},
    pages={561--577},
    year={2009},
    publisher={Elsevier}
}

@article{joo2005class,
    title={A class of hierarchical fuzzy systems with constraints on the fuzzy rules},
    author={Joo, Moon G and Lee, Jin S},
    journal={IEEE Trans. Fuzzy Syst.},
    volume={13},
    number={2},
    pages={194--203},
    year={2005},
    publisher={IEEE}
}

@article{yu2007system,
    title={System identification using hierarchical fuzzy neural networks with stable learning algorithm},
    author={Yu, Wen and Moreno-Armendariz, Marco A and Rodriguez, Floriberto Ortiz},
    journal={J. Intell. Fuzzy Syst.},
    volume={18},
    number={2},
    pages={171--183},
    year={2007},
    publisher={IOS Press}
}

@article{mohammadzadeh2016modified,
    title={A modified sliding mode approach for synchronization of fractional-order chaotic/hyperchaotic systems by using new self-structuring hierarchical type-2 fuzzy neural network},
    author={Mohammadzadeh, Ardashir and Ghaemi, Sehraneh},
    journal={Neurocomputing},
    volume={191},
    pages={200--213},
    year={2016},
    publisher={Elsevier}
}

@article{chen2007automatic,
    author    = {Chen, Yuehui and Yang, Bo and Abraham, Ajith and Peng, Lizhi},
    title     = {Automatic design of hierarchical {T}akagi--{S}ugeno type fuzzy systems using evolutionary algorithms},
    journal   = {IEEE Trans. Fuzzy Syst.},
    year      = {2007},
    volume    = {15},
    number    = {3},
    pages     = {385--397},
    note      = {109},
    publisher = {IEEE},
}

@article{salustowicz1997probabilistic,
    title={Probabilistic incremental program evolution},
    author={Salustowicz, Rafal and Schmidhuber, J{\"u}rgen},
    journal={Evol. Comput.},
    volume={5},
    number={2},
    pages={123--141},
    year={1997},
    publisher={MIT Press}
}

@article{angelov2009evolving,
    title={Evolving fuzzy systems},
    author={Angelov, Plamen},
    journal={Encycl. Complexity Syst. Sci.},
    pages={3242--3255},
    year={2009},
    publisher={Springer}
}

@Article{losing2018incremental,
  author    = {Losing, Viktor and Hammer, Barbara and Wersing, Heiko},
  title     = {Incremental on-line learning: A review and comparison of state of the art algorithms},
  journal   = {Neurocomputing},
  year      = {2018},
  volume    = {275},
  pages     = {1261--1274},
  publisher = {Elsevier},
}

@techreport{baluja1994population,
    title={Population-based incremental learning. a method for integrating genetic search based function optimization and competitive learning},
    author={Baluja, Shumeet},
    year={1994},
    number={CMU-CS-94-163},
    institution={Carnegie Mellon University}
}

@Article{gama2014survey,
  author    = {Gama, Jo{\~a}o and {\v{Z}}liobait{\.e}, Indr{\.e} and Bifet, Albert and Pechenizkiy, Mykola and Bouchachia, Abdelhamid},
  title     = {A survey on concept drift adaptation},
  journal   = {ACM Comput. Surv.},
  year      = {2014},
  volume    = {46},
  number    = {4},
  pages     = {44},
  publisher = {ACM},
}

@article{elwell2011incremental,
    title={Incremental learning of concept drift in nonstationary environments},
    author={Elwell, Ryan and Polikar, Robi},
    journal={IEEE Trans. Neural Netw.},
    volume={22},
    number={10},
    pages={1517--1531},
    year={2011},
    publisher={IEEE}
}

@article{lughofer2011handling,
    title={Handling drifts and shifts in on-line data streams with evolving fuzzy systems},
    author={Lughofer, Edwin and Angelov, Plamen},
    journal={Appl. Soft Comput.},
    volume={11},
    number={2},
    pages={2057--2068},
    year={2011},
    publisher={Elsevier}
}

@article{angelov2004approach,
    title={An approach to online identification of {T}akagi-{S}ugeno fuzzy models},
    author={Angelov, Plamen P and Filev, Dimitar P},
    journal={IEEE Trans. Syst. Man Cybern. Part B Cybern.},
    volume={34},
    number={1},
    pages={484--498},
    year={2004},
    publisher={IEEE}
}

@article{angelov2010evolving,
    title={Evolving {T}akagi-{S}ugeno Fuzzy Systems from Streaming Data ({eTS+})},
    author={Angelov, Plamen},
    journal={Evol. Intell. Syst. Method. Appl.},
    pages={21--50},
    year={2010},
    publisher={Wiley Online Library}
}

@article{lughofer2008extensions,
    title={Extensions of vector quantization for incremental clustering},
    author={Lughofer, Edwin},
    journal={Pattern Recognit.},
    volume={41},
    number={3},
    pages={995--1011},
    year={2008},
    publisher={Elsevier}
}

@inproceedings{lughofer2007evolving,
    title={Evolving single-and multi-model fuzzy classifiers with {FLEXFIS}-Class},
    author={Lughofer, Edwin and Angelov, Plamen and Zhou, Xiaowei},
    booktitle={IEEE Int. Conf. Fuzzy Syst.},
    pages={1--6},
    year={2007},
    organization={IEEE}
}

@article{lughofer2018incremental,
    title={Incremental rule splitting in generalized evolving fuzzy systems for autonomous drift compensation},
    author={Lughofer, Edwin and Pratama, Mahardhika and Skrjanc, Igor},
    journal={IEEE Trans. Fuzzy Syst.},
    volume={26},
    number={4},
    pages={1854--1865},
    year={2018},
    publisher={IEEE}
}

@Article{baruah2011evolving,
  author    = {Baruah, Rashmi Dutta and Angelov, Plamen},
  title     = {Evolving fuzzy systems for data streams: A survey},
  journal   = {Wiley Interdiscip. Rev.: Data Min. Knowl. Discovery},
  year      = {2011},
  volume    = {1},
  number    = {6},
  pages     = {461--476},
  publisher = {Wiley Online Library},
}

@book{lughofer2011evolving,
    title={Evolving fuzzy systems-methodologies, advanced concepts and applications},
    author={Lughofer, Edwin},
    volume={53},
    year={2011},
    publisher={Springer}
}

@Article{lughofer2008flexfis,
  author    = {Lughofer, Edwin David},
  title     = {{FLEXFIS}: A robust incremental learning approach for evolving {T}akagi--{S}ugeno fuzzy models},
  journal   = {IEEE Trans. Fuzzy Syst.},
  year      = {2008},
  volume    = {16},
  number    = {6},
  pages     = {1393--1410},
  publisher = {IEEE},
}

@article{angelov2008evolving,
    title={Evolving fuzzy-rule-based classifiers from data streams},
    author={Angelov, Plamen P and Zhou, Xiaowei},
    journal={IEEE Trans. Fuzzy Syst.},
    volume={16},
    number={6},
    pages={1462--1475},
    year={2008},
    publisher={IEEE}
}

@article{lima2010evolving,
    title={Evolving fuzzy modeling using participatory learning},
    author={Lima, Elton and Hell, M and Ballini, R and Gomide, Fernando},
    journal={Evol. Intell. Syst. Method. Appl.},
    pages={67--86},
    year={2010},
    publisher={Wiley Online Library}
}

@article{lemos2011multivariable,
    title={Multivariable gaussian evolving fuzzy modeling system},
    author={Lemos, Andre and Caminhas, Walmir and Gomide, Fernando},
    journal={IEEE Trans. Fuzzy Syst.},
    volume={19},
    number={1},
    pages={91--104},
    year={2011},
    publisher={IEEE}
}

@inproceedings{zhou2007autonomous,
    title={Autonomous visual self-localization in completely unknown environment using evolving fuzzy rule-based classifier},
    author={Zhou, Xiaowei and Angelov, Plamen},
    booktitle={IEEE Symp. Comput.  Intell. in Secur. and Defense Appl.},
    pages={131--138},
    year={2007},
    organization={IEEE}
}

@inproceedings{lima2006participatory,
    title={Participatory evolving fuzzy modeling},
    author={Lima, Elton and Gomide, Fernando and Ballini, Rosangela},
    booktitle={Int. Symp. on Evol. Fuzzy Syst.},
    pages={36--41},
    year={2006},
    organization={IEEE}
}

@article{lughofer2013line,
    title={On-line assurance of interpretability criteria in evolving fuzzy systems--achievements, new concepts and open issues},
    author={Lughofer, Edwin},
    journal={Inf. Sci.},
    volume={251},
    pages={22--46},
    year={2013},
    publisher={Elsevier}
}

@article{lughofer2011line,
    title={On-line elimination of local redundancies in evolving fuzzy systems},
    author={Lughofer, Edwin and Bouchot, Jean-Luc and Shaker, Ammar},
    journal={Evol. Syst.},
    volume={2},
    number={3},
    pages={165--187},
    year={2011},
    publisher={Springer}
}

@inproceedings{gomez2002evolving,
    title={Evolving fuzzy classifiers for intrusion detection},
    author={Gomez, Jonatan and Dasgupta, Dipankar},
    booktitle={Proc. 2002 IEEE Workshop on Inf. Assur.},
    volume={6},
    issue={3},
    pages={321--323},
    year={2002},
    organization={New York: IEEE Computer Press}
}

@article{lemos2013adaptive,
    title={Adaptive fault detection and diagnosis using an evolving fuzzy classifier},
    author={Lemos, Andre and Caminhas, Walmir and Gomide, Fernando},
    journal={Inf. Sci.},
    volume={220},
    pages={64--85},
    year={2013},
    publisher={Elsevier}
}

@article{stanley2002evolving,
    title={Evolving neural networks through augmenting topologies},
    author={Stanley, Kenneth O and Miikkulainen, Risto},
    journal={Evol. Comput.},
    volume={10},
    number={2},
    pages={99--127},
    year={2002},
    publisher={MIT Press}
}

@article{schlimmer1986incremental,
    title={Incremental learning from noisy data},
    author={Schlimmer, Jeffrey C and Granger, Richard H},
    journal={Mach. Learn.},
    volume={1},
    number={3},
    pages={317--354},
    year={1986},
    publisher={Springer}
}

@inproceedings{wang1992smart,
    title={A smart algorithm for incremental learning},
    author={Wang, EH-C and Kuh, Anthony},
    booktitle={Int. Jt. Conf. Neural Netw.},
    volume={3},
    pages={121--126},
    year={1992},
    organization={IEEE}
}

@article{feng2009error,
    title={Error minimized extreme learning machine with growth of hidden nodes and incremental learning},
    author={Feng, Guorui and Huang, Guang-Bin and Lin, Qingping and Gay, Robert Kheng Leng},
    journal={IEEE Trans. Neural Netw.},
    volume={20},
    number={8},
    pages={1352--1357},
    year={2009}
}

@article{angelov2004flexible,
    title={Flexible models with evolving structure},
    author={Angelov, Plamen P and Filev, Dimitar P},
    journal={Int. J. Intell. Syst.},
    volume={19},
    number={4},
    pages={327--340},
    year={2004},
    publisher={Wiley Online Library}
}

@article{juang2010recurrent,
    title={A recurrent self-evolving fuzzy neural network with local feedbacks and its application to dynamic system processing},
    author={Juang, Chia-Feng and Lin, Yang-Yin and Tu, Chiu-Chuan},
    journal={Fuzzy Sets Syst.},
    volume={161},
    number={19},
    pages={2552--2568},
    year={2010},
    publisher={Elsevier}
}

@inproceedings{kasabov1998evolving,
    title={Evolving fuzzy neural networks-algorithms, applications and biological motivation},
    author={Kasabov, Nikola},
    booktitle={Proc. 4th Int. Conf. Soft Comput.},
    volume={1},
    pages={271--274},
    year={1998},
    publisher={World Scientific}
}

@book{casillas2013interpretability,
    title={Interpretability issues in fuzzy modeling},
    author={Casillas, Jorge and Cord{\'o}n, Oscar and Triguero, Francisco Herrera and Magdalena, Luis},
    volume={128},
    year={2013},
    publisher={Springer}
}

@article{pulkkinen2010dynamically,
    title={A dynamically constrained multiobjective genetic fuzzy system for regression problems},
    author={Pulkkinen, Pietari and Koivisto, Hannu},
    journal={IEEE Trans. Fuzzy Syst.},
    volume={18},
    number={1},
    pages={161--177},
    year={2010},
    publisher={IEEE}
}

@InProceedings{ishibuchi2007multiobjective,
    author    = {Ishibuchi, Hisao},
    title     = {Multiobjective genetic fuzzy systems: review and future research directions},
    booktitle = {IEEE Int. Fuzzy Syst.},
    year      = {2007},
    pages     = {1--6},
}

@article{ishibuchi2007analysis,
    title={Analysis of interpretability-accuracy tradeoff of fuzzy systems by multiobjective fuzzy genetics-based machine learning},
    author={Ishibuchi, Hisao and Nojima, Yusuke},
    journal={Int. J. Approximate Reasoning},
    volume={44},
    number={1},
    pages={4--31},
    year={2007},
    publisher={Elsevier}
}

@article{ishibuchi1997single,
    author =    {Ishibuchi, Hisao and Murata, Tadahiko and T{\"u}rk{\c{s}}en, IB},
    title =     {Single-objective and two-objective genetic algorithms for selecting linguistic rules for pattern classification problems},
    journal={Fuzzy Sets Syst.},
    year =      {1997},
    volume =    {89},
    number =    {2},
    pages =     {135--150},
    publisher = {Elsevier}
}

@article{alcala2007multi,
    author =    {Alcal{\'a}, Rafael and Gacto, Mar{\'\i}a Jos{\'e} and Herrera, Francisco and Alcal{\'a}-Fdez, Jes{\'u}s},
    title =     {A multi-objective genetic algorithm for tuning and rule selection to obtain accurate and compact linguistic fuzzy rule-based systems},
    journal={Int. J. Uncertainty Fuzziness Knowledge-Based Syst.},
    year =      {2007},
    volume =    {15},
    number =    {05},
    pages =     {539--557},
    publisher = {World Scientific}
}

@Article{guillaume2001designing,
  author    = {Guillaume, Serge},
  title     = {Designing fuzzy inference systems from data: An interpretability-oriented review},
  journal   = {IEEE Trans. Fuzzy Syst.},
  year      = {2001},
  volume    = {9},
  number    = {3},
  pages     = {426--443},
  publisher = {IEEE},
}

@article{ishibuchi2006evolutionary,
    author =    {Ishibuchi, Hisao and Nojima, Yusuke},
    title =     {Evolutionary multiobjective optimization for the design of fuzzy rule-based ensemble classifiers},
    journal={Int. J. Hybrid Intell. Syst.},
    year =      {2006},
    volume =    {3},
    number =    {3},
    pages =     {129--145},
    publisher = {IOS Press}
}

@article{gacto2009adaptation,
    author    = {Gacto, Mar{\'\i}a Jos{\'e} and Alcal{\'a}, Rafael and Herrera, Francisco},
    title     = {Adaptation and application of multi-objective evolutionary algorithms for rule reduction and parameter tuning of fuzzy rule-based systems},
    journal   = {Soft Comput.},
    year      = {2009},
    volume    = {13},
    number    = {5},
    pages     = {419--436},
    publisher = {Springer},
}

@article{ducange2010multi,
    author    = {Ducange, Pietro and Lazzerini, Beatrice and Marcelloni, Francesco},
    title     = {Multi-objective genetic fuzzy classifiers for imbalanced and cost-sensitive datasets},
    journal   = {Soft Comput.},
    year      = {2010},
    volume    = {14},
    number    = {7},
    pages     = {713--728},
    publisher = {Springer},
}

@InCollection{cordon2003multiobjective,
    author    = {Cord{\'o}n, Oscar and Del Jesus, Mar{\'\i}a Jos{\'e} and Herrera, Francisco and Magdalena, Luis and Villar, Pedro},
    title     = {A multiobjective genetic learning process for joint feature selection and granularity and contexts learning in fuzzy rule-based classification systems},
    booktitle = {Interpretability Issues in Fuzzy Modeling},
    publisher = {Springer},
    year      = {2003},
    pages     = {79--99},
}

@article{wang2005multi,
    author =    {Wang, Hanli and Kwong, Sam and Jin, Yaochu and Wei, Wei and Man, Kim-Fung},
    title =     {Multi-objective hierarchical genetic algorithm for interpretable fuzzy rule-based knowledge extraction},
    journal={Fuzzy Sets Syst.},
    year =      {2005},
    volume =    {149},
    number =    {1},
    pages =     {149--186},
    publisher = {Elsevier}
}

@article{munoz2008automatic,
    author = {Munoz-Salinas, Rafael and Aguirre, Eugenio and Cord{\'o}n, Oscar and Garc{\'\i}a-Silvente, Miguel},
    title = {Automatic tuning of a fuzzy visual system using evolutionary algorithms: single-objective versus multiobjective approaches},
    journal={IEEE Trans. Fuzzy Syst.},
    year = {2008},
    volume = {16},
    number = {2},
    pages = {485--501},
    publisher = {IEEE}
}

@article{alcala2009multiobjective,
    title={A multiobjective evolutionary approach to concurrently learn rule and data bases of linguistic fuzzy-rule-based systems},
    author={Alcal{\'a}, Rafael and Ducange, Pietro and Herrera, Francisco and Lazzerini, Beatrice and Marcelloni, Francesco},
    journal={IEEE Trans. Fuzzy Syst.},
    volume={17},
    number={5},
    pages={1106--1122},
    year={2009},
    publisher={IEEE}
}

@article{antonelli2011learning,
    title={Learning knowledge bases of multi-objective evolutionary fuzzy systems by simultaneously optimizing accuracy, complexity and partition integrity},
    author={Antonelli, Michela and Ducange, Pietro and Lazzerini, Beatrice and Marcelloni, Francesco},
    journal={Soft Comput.},
    volume={15},
    number={12},
    pages={2335--2354},
    year={2011},
    publisher={Springer}
}

@Article{antonelli2012genetic,
  author    = {Antonelli, Michela and Ducange, Pietro and Marcelloni, Francesco},
  title     = {Genetic training instance selection in multiobjective evolutionary fuzzy systems: A coevolutionary approach},
  journal   = {IEEE Trans. Fuzzy Syst.},
  year      = {2012},
  volume    = {20},
  number    = {2},
  pages     = {276--290},
  publisher = {IEEE},
}

@article{fazzolari2013review,
    author =    {Fazzolari, Michela and Alcala, Rafael and Nojima, Yusuke and Ishibuchi, Hisao and Herrera, Francisco},
    title =     {A review of the application of multiobjective evolutionary fuzzy systems: current status and further directions},
    journal={IEEE Trans. Fuzzy Syst.},
    year =      {2013},
    volume =    {21},
    number =    {1},
    pages =     {45--65},
}

@article{deng2017hierarchical,
    title={A hierarchical fused fuzzy deep neural network for data classification},
    author={Deng, Yue and Ren, Zhiquan and Kong, Youyong and Bao, Feng and Dai, Qionghai},
    journal={IEEE Trans. Fuzzy Syst.},
    volume={25},
    number={4},
    pages={1006--1012},
    year={2017},
    publisher={IEEE}
}

@article{alshomrani2015proposal,
    title={A proposal for evolutionary fuzzy systems using feature weighting: dealing with overlapping in imbalanced datasets},
    author={Alshomrani, Saleh and Bawakid, Abdullah and Shim, Seong-O and Fern{\'a}ndez, Alberto and Herrera, Francisco},
    journal={Knowledge-Based Syst.},
    volume={73},
    pages={1--17},
    year={2015},
    publisher={Elsevier}
}

@Article{rey2017multi,
  author    = {Rey, M Isabel and Galende, Marta and Fuente, Maria J and Sainz-Palmero, GI},
  title     = {Multi-objective based Fuzzy Rule Based Systems ({FRBSs}) for trade-off improvement in accuracy and interpretability: A rule relevance point of view},
  journal   = {Knowledge-Based Syst.},
  year      = {2017},
  volume    = {127},
  pages     = {67--84},
  publisher = {Elsevier},
}

@article{yen1999simplifying,
    title={Simplifying fuzzy rule-based models using orthogonal transformation methods},
    author={Yen, John and Wang, Liang},
    journal={IEEE Trans. Syst. Man Cybern. Part B Cybern.},
    volume={29},
    number={1},
    pages={13--24},
    year={1999},
    publisher={IEEE}
}

@article{dubois1997checking,
    title={Checking the coherence and redundancy of fuzzy knowledge bases},
    author={Dubois, Didier and Prade, Henri and Ughetto, Laurent},
    journal={IEEE Trans. Fuzzy Syst.},
    volume={5},
    number={3},
    pages={398--417},
    year={1997},
    publisher={IEEE}
}

@Article{gacto2011interpretability,
  author    = {Gacto, Maria Jose and Alcal{\'a}, Rafael and Herrera, Francisco},
  title     = {Interpretability of linguistic fuzzy rule-based systems: An overview of interpretability measures},
  journal   = {Inf. Sci.},
  year      = {2011},
  volume    = {181},
  number    = {20},
  pages     = {4340--4360},
  publisher = {Elsevier},
}

@article{cpalka2014new,
    title={A new method for designing neuro-fuzzy systems for nonlinear modelling with interpretability aspects},
    author={Cpa{\l}ka, K and {\L}apa, K and Przyby{\l}, A and Zalasi{\'n}ski, M},
    journal={Neurocomputing},
    volume={135},
    pages={203--217},
    year={2014},
    publisher={Elsevier}
}

@article{torra2010hesitant,
    title={Hesitant fuzzy sets},
    author={Torra, Vicen{\c{c}}},
    journal={Int. J. Intell. Syst.},
    volume={25},
    number={6},
    pages={529--539},
    year={2010},
    publisher={Wiley Online Library}
}

@article{atanassov1986intuitionistic,
    title={Intuitionistic fuzzy sets.},
    author={Atanassov, Krassimir T},
    journal={Fuzzy sets and Systems},
    volume={20},
    number={1},
    pages={87--96},
    year={1986}
}

@article{mendel2002type,
    title={Type-2 fuzzy sets made simple},
    author={Mendel, Jerry M and John, RI Bob},
    journal={IEEE Trans. Fuzzy Syst.},
    volume={10},
    number={2},
    pages={117--127},
    year={2002},
    publisher={IEEE}
}

@Article{jain2010data,
  author    = {Jain, Anil K},
  title     = {Data clustering: 50 years beyond {K}-means},
  journal   = {Pattern Recognit. Lett.},
  year      = {2010},
  volume    = {31},
  number    = {8},
  pages     = {651--666},
  publisher = {Elsevier},
}

@book{bellman2013dynamic,
    title={Dynamic programming},
    author={Bellman, Richard},
    year={2013},
    publisher={Courier Corporation}
}

@article{castano2001global,
    title={Global viewing of heterogeneous data sources},
    author={Castano, Silvana and De Antonellis, Valeria},
    journal={IEEE Trans. Knowl. Data Eng.},
    volume={13},
    number={2},
    pages={277--297},
    year={2001},
    publisher={IEEE}
}

@Book{feldman2007text,
  title     = {The text mining handbook: Advanced approaches in analyzing unstructured data},
  publisher = {Cambridge University Press},
  year      = {2007},
  author    = {Feldman, Ronen and Sanger, James},
}

@article{jackson1972interpretation,
    title={Interpretation of inaccurate, insufficient and inconsistent data},
    author={Jackson, David D},
    journal={Geophys. J. R. Astron. Soc.},
    volume={28},
    number={2},
    pages={97--109},
    year={1972},
    publisher={Wiley Online Library}
}

@article{pradlwarter2008use,
    title={The use of kernel densities and confidence intervals to cope with insufficient data in validation experiments},
    author={Pradlwarter, HJ and Schu{\"e}ller, GI},
    journal={Comput. Methods Appl. Mech. Eng.},
    volume={197},
    number={29-32},
    pages={2550--2560},
    year={2008},
    publisher={Elsevier}
}

@Article{chawla2002smote,
  author  = {Chawla, Nitesh V and Bowyer, Kevin W and Hall, Lawrence O and Kegelmeyer, W Philip},
  title   = {{SMOTE}: synthetic minority over-sampling technique},
  journal = {J. Artif. Intell. Res.},
  year    = {2002},
  volume  = {16},
  pages   = {321--357},
}

@article{lopez2015cost,
    title={Cost-sensitive linguistic fuzzy rule based classification systems under the MapReduce framework for imbalanced big data},
    author={L{\'o}pez, Victoria and Del R{\'\i}o, Sara and Ben{\'\i}tez, Jos{\'e} Manuel and Herrera, Francisco},
    journal={Fuzzy Sets Syst.},
    volume={258},
    pages={5--38},
    year={2015},
    publisher={Elsevier}
}

@article{wang1992fuzzy,
    title={Fuzzy basis functions, universal approximation, and orthogonal least-squares learning},
    author={Wang, L-X and Mendel, Jerry M},
    journal={IEEE Trans. Neural Netw.},
    volume={3},
    number={5},
    pages={807--814},
    year={1992},
    publisher={IEEE}
}

@Article{castillo2012optimization,
  author    = {Castillo, Oscar and Melin, Patricia},
  title     = {Optimization of type-2 fuzzy systems based on bio-inspired methods: A concise review},
  journal   = {Inf. Sci.},
  year      = {2012},
  volume    = {205},
  pages     = {1--19},
  publisher = {Elsevier},
}

@InProceedings{purshouse2003evolutionary,
  author       = {Purshouse, Robin C and Fleming, Peter J},
  title        = {Evolutionary many-objective optimisation: An exploratory analysis},
  booktitle    = {Evolutionary Computation, 2003. CEC'03. The 2003 Congress on},
  year         = {2003},
  volume       = {3},
  pages        = {2066--2073},
  organization = {IEEE},
}

@article{deb2014evolutionary,
    title={An evolutionary many-objective optimization algorithm using reference-point-based nondominated sorting approach, part {I}: Solving problems with box constraints},
    author={Deb, Kalyanmoy and Jain, Himanshu},
    journal={IEEE Trans. Evol.Comput.},
    volume={18},
    number={4},
    pages={577--601},
    year={2014},
    publisher={IEEE}
}

@article{lecun2015deep,
    title={Deep learning},
    author={LeCun, Yann and Bengio, Yoshua and Hinton, Geoffrey},
    journal={Nature},
    volume={521},
    number={7553},
    pages={436},
    year={2015},
    publisher={Nature Publishing Group}
}

@article{hinton2012deep,
    title={Deep neural networks for acoustic modeling in speech recognition},
    author={Hinton, Geoffrey and Deng, Li and Yu, Dong and Dahl, George and Mohamed, Abdel-rahman and Jaitly, Navdeep and Senior, Andrew and Vanhoucke, Vincent and Nguyen, Patrick and Kingsbury, Brian and others},
    journal={IEEE Signal Process Mag.},
    volume={29},
    year={2012}
}

@article{seng1999tuning,
    title={Tuning of a neuro-fuzzy controller by genetic algorithm},
    author={Seng, Teo Lian and Khalid, M Bin and Yusof, Rubiyah},
    journal={IEEE Trans. Syst. Man Cybern. Part B Cybern.},
    volume={29},
    number={2},
    pages={226--236},
    year={1999},
    publisher={IEEE}
}

@Article{lee1990fuzzy,
  author    = {Lee, Chuen-Chien},
  title     = {Fuzzy logic in control systems: fuzzy logic controller. {I}},
  journal   = {IEEE Trans. Syst. Man Cybern.},
  year      = {1990},
  volume    = {20},
  number    = {2},
  pages     = {404--418},
  publisher = {IEEE},
}

@article{wang1996approach,
    title={An approach to fuzzy control of nonlinear systems: Stability and design issues},
    author={Wang, Hua O and Tanaka, Kazuo and Griffin, Michael F},
    journal={IEEE Trans. Fuzzy Syst.},
    volume={4},
    number={1},
    pages={14--23},
    year={1996},
    publisher={IEEE}
}

@book{bojadziev2007fuzzy,
    title={Fuzzy logic for business, finance, and management},
    author={Bojadziev, George},
    volume={23},
    year={2007},
    publisher={World Scientific}
}

@book{jin2008fuzzy,
    title={Fuzzy systems in bioinformatics and computational biology},
    author={Jin, Yaochu and Wang, Lipo},
    volume={242},
    year={2008},
    publisher={Springer}
}

@article{precup2011survey,
    title={A survey on industrial applications of fuzzy control},
    author={Precup, Radu-Emil and Hellendoorn, Hans},
    journal={Comput. Ind.},
    volume={62},
    number={3},
    pages={213--226},
    year={2011},
    publisher={Elsevier}
}

@Article{liao2005expert,
  author    = {Liao, Shu-Hsien},
  title     = {Expert system methodologies and applications---a decade review from 1995 to 2004},
  journal   = {Expert Systems with applications},
  year      = {2005},
  volume    = {28},
  number    = {1},
  pages     = {93--103},
  publisher = {Elsevier},
}

@article{zadeh1988fuzzy,
    title={Fuzzy logic},
    author={Zadeh, Lotfi Asker},
    journal={Computer},
    volume={21},
    number={4},
    pages={83--93},
    year={1988},
    publisher={IEEE}
}

@book{zadeh1992fuzzy,
    title={Fuzzy logic for the management of uncertainty},
    author={Zadeh, Lotfi A and Kacprzyk, Janusz},
    year={1992},
    publisher={John Wiley \& Sons}
}

@article{melin2011face,
    title={Face recognition with an improved interval type-2 fuzzy logic sugeno integral and modular neural networks},
    author={Melin, Patricia and Mendoza, Olivia and Castillo, Oscar},
    journal={IEEE Trans. Syst. Man Cybern. Part A Syst. Humans},
    volume={41},
    number={5},
    pages={1001--1012},
    year={2011},
    publisher={IEEE}
}

@article{castillo2014review,
    title={A review on interval type-2 fuzzy logic applications in intelligent control},
    author={Castillo, Oscar and Melin, Patricia},
    journal={Inf. Sci.},
    volume={279},
    pages={615--631},
    year={2014},
    publisher={Elsevier}
}

@Article{valdez2014survey,
  author    = {Valdez, Fevrier and Melin, Patricia and Castillo, Oscar},
  title     = {A survey on nature-inspired optimization algorithms with fuzzy logic for dynamic parameter adaptation},
  journal   = {Expert Syst. Appl.},
  year      = {2014},
  volume    = {41},
  number    = {14},
  pages     = {6459--6466},
  publisher = {Elsevier},
}
\end{document}